\pdfoutput=1
\documentclass[11pt]{article}

\usepackage{acl}

\usepackage{times}
\usepackage{latexsym}

\usepackage[T1]{fontenc}
\usepackage[utf8]{inputenc}

\usepackage{microtype}

\usepackage{inconsolata}

\usepackage{amsmath,amsfonts,bm}

\def\eqref#1{equation~\ref{#1}}
\def\1{\bm{1}}

\DeclareMathAlphabet{\mathsfit}{\encodingdefault}{\sfdefault}{m}{sl}
\SetMathAlphabet{\mathsfit}{bold}{\encodingdefault}{\sfdefault}{bx}{n}

\usepackage{hyperref}
\usepackage{url}
\usepackage{multirow}
\usepackage{booktabs}
\usepackage{floatrow}
\usepackage{graphicx} % For adding images
\usepackage{smile}
\usepackage{enumitem}
\usepackage{colortbl}
\usepackage{xspace}
\usepackage{xcolor,colortbl}
\usepackage{subfigure}
\usepackage{caption}
\usepackage[normalem]{ulem} % Load the ulem package
\usepackage{listings}
\usepackage{float}

\newcommand{\ours}{\textsc{ClinGen}\xspace}

\lstdefinestyle{mystyle}{
  basicstyle=\ttfamily,
  frame=single,
  breaklines=true,
  breakindent=0pt,
  backgroundcolor=\color{gray!10}, % Change the background color
}

\newcommand{\blue}[1]{{\color{black}{#1}}}

\title{Knowledge-Infused Prompting: Assessing and Advancing Clinical Text Data Generation with Large Language Models}

\author{Ran Xu$^{\heartsuit}$, Hejie Cui$^{\heartsuit}$, Yue Yu{$^\spadesuit$}, Xuan Kan$^{\heartsuit}$, Wenqi Shi{$^\spadesuit$}, Yuchen Zhuang{$^\spadesuit$}, \\ \bf May D. Wang{$^\spadesuit$}, Wei Jin$^{\heartsuit}$, Joyce C. Ho$^{\heartsuit}$, Carl Yang$^{\heartsuit}$ \\
${\heartsuit}$ Emory University \quad {$^\spadesuit$} Georgia Institute of Technology \\
\texttt{\{ran.xu,hejie.cui,xuan.kan,wei.jin,joyce.c.ho,j.carlyang\}@emory.edu} \\
\texttt{\{yueyu,wshi83,yczhuang\}@gatech.edu}
}

\begin{document}
\maketitle
\begin{abstract}
Clinical natural language processing faces challenges like complex medical terminology and clinical contexts.
% Clinical natural language processing requires methods that can address domain-specific challenges, such as complex medical terminology and clinical contexts. 
Recently, large language models (LLMs) have shown promise in this domain. Yet, their direct deployment can lead to privacy issues and are constrained by resources. 
To address this challenge, we delve into synthetic clinical text generation with LLMs for clinical NLP tasks. 
% Our approach, {\ours}, integrates clinical knowledge extraction and context-informed LLM prompting, drawing from domain-specific knowledge graphs and LLMs to infuse clinical knowledge for data generation. 
We propose an innovative, resource-efficient approach, {\ours}, which infuses knowledge into the process. Our model involves clinical knowledge extraction and context-informed LLM prompting. 
Both clinical topics and writing styles are drawn from external domain-specific knowledge graphs and LLMs to guide data generation. 
Our extensive empirical study across 8 clinical NLP tasks and 18 datasets reveals that {\ours} consistently enhances performance across various tasks by 7.7\%-8.7\% on average, effectively aligning the distribution of real datasets and  enriching the diversity of generated training instances. 
Our code is available at \url{https://github.com/ritaranx/ClinGen}.
\end{abstract}

\vspace{-1ex}
\section{Introduction}
\label{sec:intro}
Clinical Natural Language Processing (NLP) emerges as a distinct subfield including the extraction, analysis, and interpretation of unstructured clinical text~\citep{wornow2023shaky}.  
Despite its significance, unique challenges exist for methodology development in clinical NLP. For example, clinical texts are often dense with abbreviations and specialized medical terminologies can be perplexing to standard NLP models~\citep{lee2023ai}. 
Fortunately, recent advances in Large Language Models (LLMs)~\citep{brown2020language,chung2022scaling,ouyang2022training,chatgpt,gpt4} provide a promising way to resolve these issues, as they contain billions of parameters and have been pretrained on massive corpora, thus inherently capture a significant amount of clinical knowledge~\citep{agrawal2022large,singhal2022large}.  
% However, direct application of LLMs in clinical settings presents substantial challenges. The deployment and execution of these mammoth models can lead to considerable computational and financial overhead~\citep{lee2023ai}. Furthermore, the sensitive nature of medical data elevates privacy concerns and demands rigorous adherence to regulations~\citep{moor2023foundation}. 
% These unique challenges within the clinical domain necessitate tailored solutions that balance complexity with privacy and feasibility~\cite {tu2023towards,liu2023utility}.
% However, directly applying these general LLMs to the medical domain is inadequate. Currently, a domain adaption step like clinical data fine-tuning is needed to address these unique challenges like terminology complexities and boost model performance~\citep{tu2023towards,liu2023utility}.
These progresses inspire the need for designing specialized approaches for adapting LLMs to clinical settings, which both address the terminology complexities and improve models through clinical data finetuning~\citep{tu2023towards,liu2023utility}.

% The sensitive information within clinical data also necessitates privacy concerns~\citep{moor2023foundation}. 
% Such challenges highlight the need for specialized approaches for clinical settings, 
% which both address the terminology complexities and improve models through clinical data finetuning~\citep{tu2023towards,liu2023utility}.

% The potential that LLMs hold for the clinical domain is significant,  and this promise is anticipated to increase as more specialized clinical data becomes accessible~\citep{fries2022bigbio,tu2023towards,fleming2023medalign,bisercic2023interpretable}. \wei{can we have some transition from this paragraph to the next one? e.g., mentioning clinical synthetic data generation or inference}

% On a parallel note, large language models (LLMs) have demonstrated remarkable performance on a wide spectrum of NLP tasks~\citep{brown2020language,chung2022scaling,ouyang2022training,chatgpt,gpt4}. These models, characterized by the vast number of parameters, have the capacity to understand and generate human-like text~\citep{wei2022chain}. In recent studies, several works have ventured into applying LLMs specifically to the clinical domain and suggest that LLMs inherently capture a significant amount of clinical knowledge~\citep{agrawal2022large,nori2023capabilities,pmlr-v209-eric23a,wong2023scaling,singhal2022large,singhal2023towards,luo2022biogpt,liu2023radiology}. 

% \paragraph{Why study clinical synthetic data generation?} 
Despite the strong capacity of general LLMs, 
directly applying them to infer over clinical text data is often undesired in practice. 
Firstly, these LLMs often have billions of parameters that translate to significant computational resources even for inference, leading to \emph{increased infrastructure costs} and \emph{long inference time}. 
%Such limitations can be particularly constraining in real-time clinical settings where swift decision-making is paramount.
Furthermore, the sensitive patient information in the clinical text naturally raises \emph{privacy and regulatory compliance concerns}~\citep{mesko2023imperative}. 
To combat these challenges, generating synthetic training data using LLMs serves as a promising solution, as it leverages the capability of LLMs in a resource-efficient and privacy-centric way. When trained with synthetic data mimicking real-world clinical data, models can achieve high performance while obeying data protection regulations. %This paves the way for the broader and safer integration of advanced NLP techniques into the healthcare domain.

% \cite{regen,patron,yu2023large}
% Baseline \cite{} 

% DINO~\cite{schick2021generating} 
% ZeroGen~\cite{ye2022zerogen}, SuperGen~\cite{meng2022generating}
% FewGen~\cite{meng2023tuning}
% GPT3Mix~\cite{gpt3mix}

% \paragraph{Limitations of existing work and our contributions.} 
Synthetic data generation with LLMs is a popular research area in NLP~\citep{meng2022generating,ye2022zerogen,ye2022progen,wang2023lets}, with a focus on gemeral-domain data. 
% However, when considering producing high-quality data that conforms to the distribution of the original dataset, simply adapting LLMs trained on general texts to generate clinical data presents unique challenges. 
However, adapting LLMs trained on general texts for generating high-quality clinical data poses distinct challenges. 
To assess the quality of data generated by existing methods, we carry out an  evaluation centered on distribution and diversity, detailed in Section~\ref{sec:preliminary}, which
% . Insights from the t-SNE embeddings visualization and the Central Moment Discrepancy (CMD) score 
indicate a noteworthy data distribution shift. We further examine the clinically-related entity quantities and frequencies in synthetic data, where a notable decline is observed when contrasting synthetic data with ground truth data.  
While some research has delved into clinical data generation with language models, many of these efforts are tailored to specific tasks. 
Examples include medical dialogues~\citep{chintagunta2021medically}, clinical notes~\citep{giorgi2023clinical}, 
% medical QA~\citep{guo2023dr}, 
and electronic health records~\citep{ive2020generation}. 
These studies often directly adopt language models for text generation, and sometimes on excessive training data.
Till now, a unified principle to better adapt LLMs for generating synthetic text for facilitating clinical downstream applications is still missing.
% This limitation is especially salient in real-world contexts when usually only minimal training data is available. 

Motivated by the above analysis, we propose \ours, a \emph{clinical knowledge-infused}  framework for high-quality clinical text generation in few-shot scenarios. 
% Our ultimate goal is to narrow the gap between synthetic and ground-truth data and encourage the topic diversity of the generated text. 
Our ultimate goal is to bridge the gap between synthetic and real data while enhancing topic diversity. 
Towards this end, we propose to utilize clinical knowledge extraction to contextualize the prompts. \blue{This includes generating clinical topics on entity and relation information from both KGs and LLMs and deriving writing style suggestions from LLMs.}
By doing this, {\ours} integrates both \emph{non-parametric insights} from external clinical knowledge graphs with the \emph{intrinsic parametric knowledge} encoded in LLMs and \emph{enjoys higher diversity} via dynamically composing different topics and writing styles together during the data generation process.
It is worth noting that, \ours only relies on minimal additional human efforts, and can be readily applied to a wide array of core tasks in clinical NLP.
% We conduct an exhaustive empirical evaluation of synthetic clinical data generation \textbf{across 7 clinical NLP tasks and 16 datasets}. Empirical findings demonstrate that the proposed knowledge-infused framework not only aligns more closely with the distribution of the original data and amplifies the diversity of the generated training samples but also consistently enhances performance across various tasks.
% \wei{can we use some numbers to highlight the improvement? For example, XXXX performs the best in most cases with improvements up to XX\% over the best baselines. }

% \ran{release code}

Our contributions can be summarized as follows:

$\bullet$ We propose {\ours}, a generic clinical knowledge-infused framework for clinical text data generation in few-shot settings. It can be readily applied to a wide range of tasks in clinical NLP.

$\bullet$ We present an analysis of the pitfall of existing data generation approaches for clinical text data, and propose a 
simple yet effective strategy to extract clinical knowledge and customize the prompts toward target clinical NLP tasks. This includes generating clinical topics from both KGs and LLMs and deriving writing style suggestions from LLMs. 

$\bullet$ We conduct an exhaustive evaluation of synthetic clinical data generation \textbf{across 8 clinical NLP tasks and 18 datasets}. Empirical findings demonstrate that {\ours} not only aligns more closely with the distribution of the original data but also amplifies the diversity of the generated training samples. The empirical performance gains are consistent across various tasks with different LLMs and classifiers (8.7\% for PubMedBERT$_{\texttt{Base}}$ and 7.7\% for PubMedBERT$_{\texttt{Large}}$).

\section{Related Work}
% \vspace{-1ex}
Generating additional training data enables a more precise analysis of medical text, and has gained more attention in the past years. 
Earlier research has employed data augmentation techniques to generate similar samples to existing instances with word substitution~\citep{kang2021umls}, back translation~\citep{uda}, pretrained transformers~\citep{xu2023weakly,melm}. But they often yield rigid transformations and the quality of the augmented text cannot be always guaranteed. 
% Another line of work focuses on leveraging external knowledge to create weak  abels~\citep{ratner2017snorkel,fries2017swellshark,wang2019clinical}. 
% \footnote{Some works also name it as `distant supervision'~\citep{mintz2009distant,min2013distant,liang2020bond}.}. 
% These methods typically require domain expertise and additional task-specific corpora, which can be resource-intensive to obtain for low-resource clinical tasks. 
% Moreover, designing effective rules can be challenging due to the reliance on domain-specific knowledge. 
% \citep{zhang2021wrench}.

The emergence of LLMs has presented new possibilities for synthetic data generation~\citep{meng2022generating,meng2023tuning,ye2022zerogen,li-etal-2023-synthetic}. However, these methods often use generic and simple prompts that may not fully capture domain-specific knowledge, thus potentially limiting the quality of the generated data. 
\citet{liu2022wanli,chung-etal-2023-increasing,yu2023large} employ interactive learning to generate instances, at the cost of additional human efforts.
Several recent studies explore LLM-based synthetic data generation for clinical NLP.  
\citet{tang2023does} rely on a \emph{much larger training set} to generate candidate entities, which {disregards the practical low-resource setting}~\citep{perez2021true}. 
% Furthermore, these studies often focus on specific target tasks, lacking generality in terms of exploring a diverse set of clinical NLP applications.
Moreover, these studies often concentrate on specific target tasks, thus lacking generality for diverse clinical NLP scenarios.

On the other hand, several works aimed at optimizing prompts using LLMs~\citep{zhou2023large,wang2023promptagent} or knowledge graphs~\citep{cui2023a,liu-etal-2022-generated,chen2022knowprompt}, yet they mainly focus on refining prompts to obtain the answer for the given input, and the prompt template often remains unchanged. 
Instead, we focus on the different task of generating training instances. By composing different topics and styles together, we can generate diverse templates for prompting LLMs to improve the quality of the synthetic data.

% \vspace{-3ex}
 \begin{figure*}[!t]
	\centering
	\vspace{-2.5ex}
	\subfigure[CMD]{
		\includegraphics[width=0.34\linewidth]{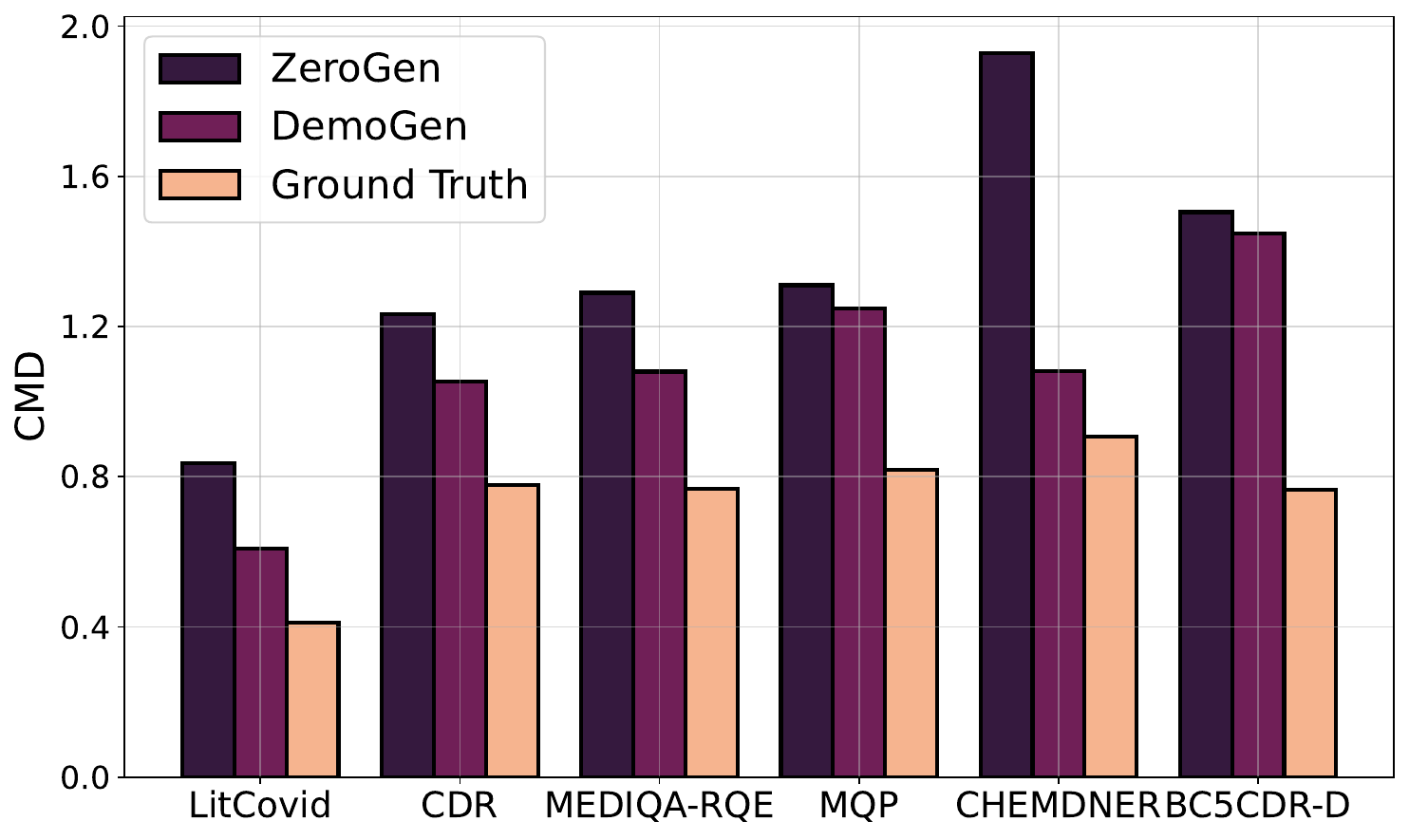}
		\label{fig:cmd-baseline}
	} %\hfill
         \hspace{-1.5ex}
	% \subfigure[MEDIQA-RQE]{
	% 	\includegraphics[width=0.25\linewidth]{figures/mediqa_rqe_sentencebert_bsl.pdf}
	% 	\label{fig:mediqa_rqe_sentencebert_bsl}
	% }\hspace{-1.5ex}
     \subfigure[Entity Coverage]{
		\includegraphics[width=0.34\linewidth]{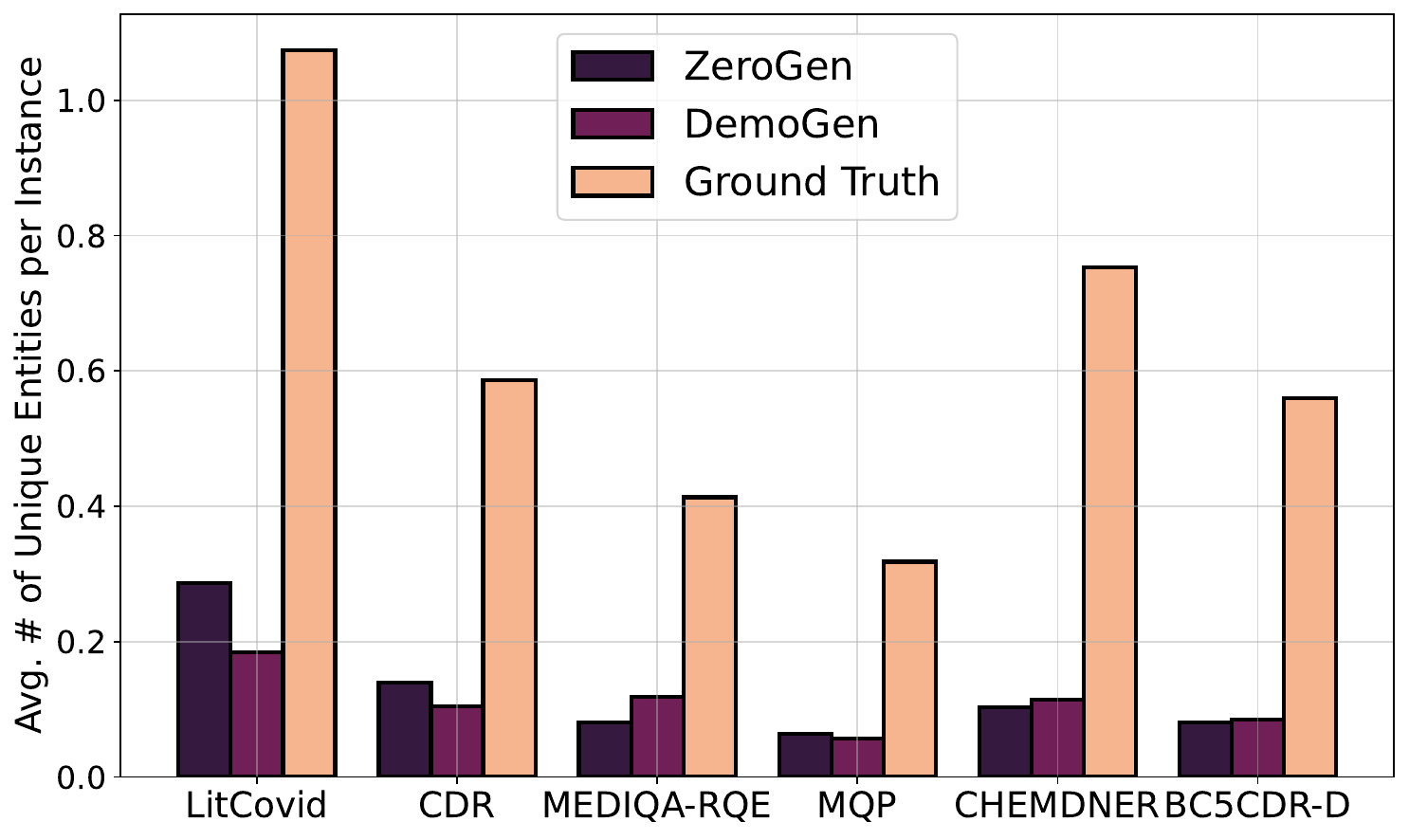}
		\label{fig:avg-entity-baseline}
	}
 \hspace{-1.5ex}
      \subfigure[Entity Frequency]{
		\includegraphics[width=0.27\linewidth]{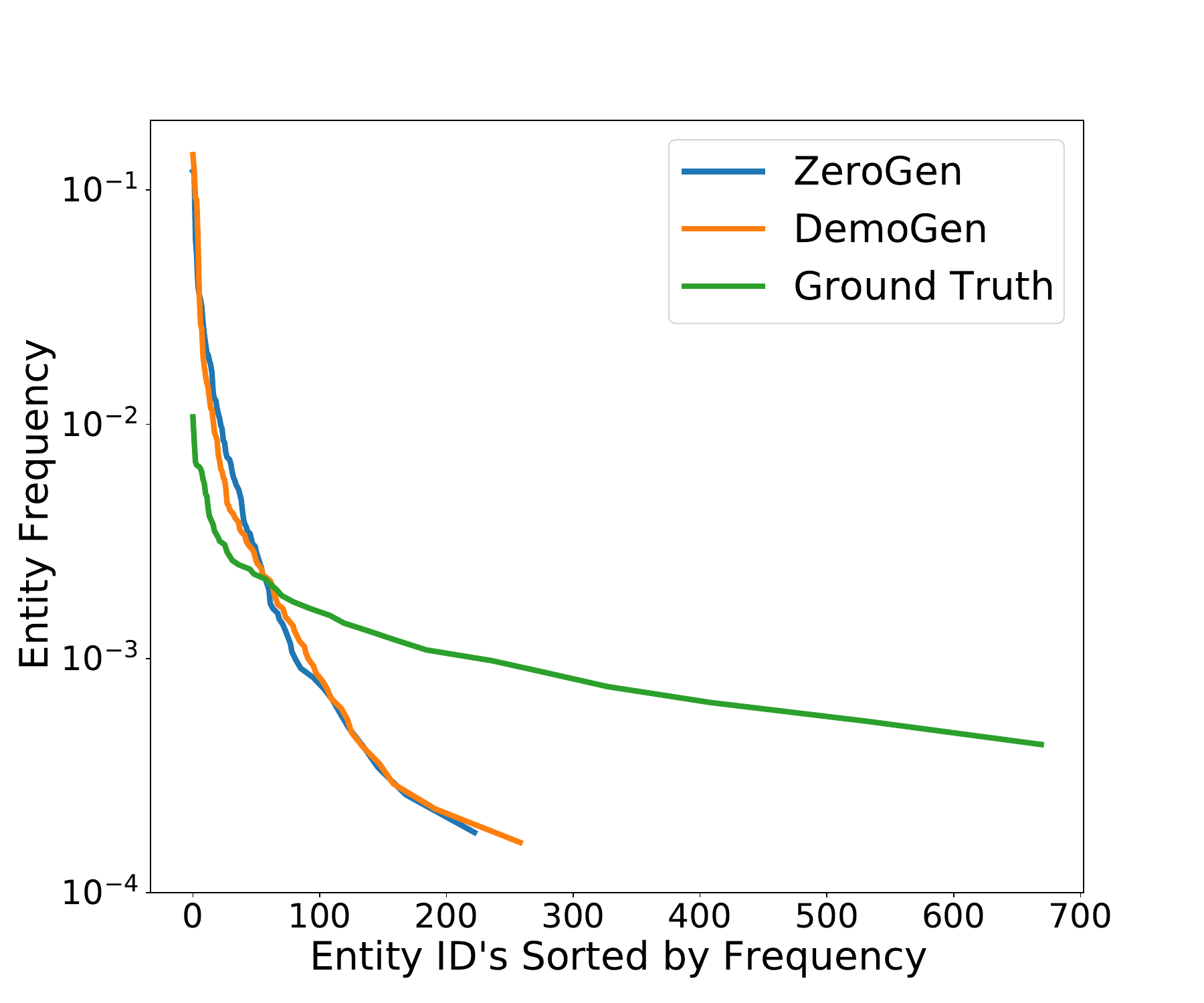}
		\label{fig:bc5cdr_disease_freq_bsl}
	}
 % \vspace{-ex}
	\caption{Preliminary Studies. (c) is from BC5CDR-Disease and is in log scale. \vspace{-1ex}}

\label{fig:prelim2}
\end{figure*}

\section{Preliminary Study}
\vspace{-0.5ex}

\label{sec:preliminary}
This section first presents the foundational setup of synthetic data generation. 
Then, we provide an in-depth investigation into the pitfalls of existing synthetic data generation methods. 
% We observe that these methods often  introduce distribution shifts and exhibit limited diversity, which can be suboptimal for improving  downstream performance.

% Subsequently, we delve into an examination of the limitations observed in the existing methods, particularly focusing on issues including distribution shift and limited diversity.

\vspace{-1ex}
\subsection{Problem Setup}
In this paper, we study synthetic data generation under the few-shot setting.
The input consists of a training set $\cD=\{(x_i,y_i)\}_{i=1}^K$, where $(x_i, y_i)$ represents an input text and its corresponding label $y_i \in \cY$ for the $i$-th example. $K$ denotes the total number of training samples, which is  kept at a very small value (5-shot per label). The primary objective is to harness the LLM $\cM$ to generate a synthetic dataset, denoted as $\tilde{\cD}=\{(\tilde{x_i},\tilde{y_i})\}_{i=1}^N$, where $N$ is the number of generated samples ($N \gg K$). 
We use $\rho(\cdot)$ to denote the generation process from the LLM.
% To use  for 
For each downstream task, we fine-tune a classifier $\cC_{\theta}$ \blue{(a moderate-size pre-trained language model)} parameterized by $\theta$ on the synthetic dataset $\tilde{\cD}$ for evaluating its quality.\footnote{While In-context Learning~\citep{brown2020language} can also be utilized, it is often hard to fit all generated instances into the context window, especially for datasets with high cardinality.}
% This deliberate design is aimed at leveraging $\cM$ to create a substantially augmented dataset for downstream tasks.

% The training set Dtrain = {(x, y)i}
% consists of K training samples per label where x =
% [x1, x2, . . . , xn] is a text sequence with n tokens.
\vspace{-0.5ex}
\subsection{Limitations of Existing  Methods}
\label{sec:limitations}
Denote the task-specific prompts for class label name $j$ as $p_j$, we take a closer look at the synthetic text data generated by two representative approaches: ZeroGen~\citep{ye2022zerogen}, which directly instructs LLMs for data generation as  $\tilde{\cD}_{\text{Zero}} \sim \rho_{j\sim\cY}(\cdot; p_j)$, and DemoGen~\citep{gpt3mix,meng2023tuning}, which augments the prompt with few-shot demonstrations $\cD$ as $\tilde{\cD}_{\text{Demo}} \sim \rho_{j\sim\cY}\left(\cdot; [p_j, \cD]\right)$.
% $\cD_{\text{train}}$
The prompt format of ZeroGen and DemoGen are in Appendix~\ref{sec:prompt_format_bsl}.
We observe that these methods often introduce \textit{distribution shifts} and exhibit \textit{limited diversity}, which can lead to suboptimal downstream performance. 
% The illustration is as follows, and we include additional embedding visualizations in Appendix~\ref{sec:add_prelim}.

\noindent \textbf{Distribution Shift.} An inherent issue when adapting LLMs to specific domains for text generation is the \emph{distribution shift}, given that LLMs are primarily trained on vast amounts of web text in general domains. 
% as there exists a large discrepancy between the embeddings of the ground truth data and the synthetic data. 
% \ran{sentencebert, more in appendix} 
To quantify the data distribution shift, we employ Central Moment Discrepancy (CMD)~\citep{zellinger2017central} to measure the gap between synthetic and real data across six clinical NLP datasets --- a high CMD value indicates a large gap between two distributions\footnote{Details of calculating CMD is in Appendix \ref{sec:cmd}.}. Figure \ref{fig:cmd-baseline} illustrates that both ZeroGen and DemoGen exhibit elevated CMD scores. 
% \hj{What are the two distributions used to measure CMD for the Ground Truth bar (if they are not 1.0)?}.
% This limitation remains evident even when incorporating few-shot demonstrations into the process, indicating a notable disparity between the ground truth data and synthetic data. 
Despite the inclusion of few-shot demonstrations in DemoGen, this limitation remains evident, indicating a notable disparity between the ground-truth and synthetic data.
%indicating substantial dissimilarity between the synthetic data and those of the real dataset. 
% We provide the t-SNE plots of the training data embeddings that exhibit similar patterns in Appendix~\ref{sec:add_prelim}.

% TSNE-embedding
% Case Study

\noindent \textbf{Limited Diversity.}
Clinical datasets in real-world scenarios often include rich domain knowledge that can be challenging to replicate in synthetic data. We evaluate synthetic dataset diversity by using both entity quantity and their normalized frequencies. The results are illustrated in Figures~\ref{fig:avg-entity-baseline} and \ref{fig:bc5cdr_disease_freq_bsl}. Our analysis reveals that datasets generated by ZeroGen and DemoGen exhibit a limited number of clinical entities, having a substantial discrepancy with the ground truth. 
Furthermore, it is highlighted that only a minority of potential entities and relations are frequently referenced across instances, while the majority are generated infrequently.

% Furthermore, these synthetic entities exhibit a \textit{long-tailed distribution} of normalized frequencies, highlighting that only a minority are frequently referenced across instances, while the majority of potential entities and relations are generated infrequently.
%  of synthetic datasets created with existing methods
To explicitly illustrate the limitations, we present a case study in Figure~\ref{fig:prelim1},  Appendix~\ref{sec:add_prelim}. 
% In this case study, we randomly select one sample from each class within the training set generated by ZeroGen and DemoGen. These selected samples are compared with the ground truth data from the MEDIQA-RQE dataset, which aims to predict whether a consumer health query can entail an existing Frequently Asked Question (FAQ). 
The comparison reveals that samples generated by ZeroGen and DemoGen lack  \textit{sufficient details} present in the ground truth data. 
Besides, the generated samples adhere to a more uniform style, while the ground truth encompasses various situations and writing styles, including urgent and informal inquiries.
% Furthermore, the generated samples lack the \textit{situation diversity} and \textit{style variability} present in the ground truth. The ground truth encompasses various situations and writing styles, including urgent and informal inquiries, while the generated samples adhere to a more uniform style and structure.

\begin{figure*}[t]
    \centering
    % \vspace{-ex}
    \includegraphics[width=0.97\linewidth]{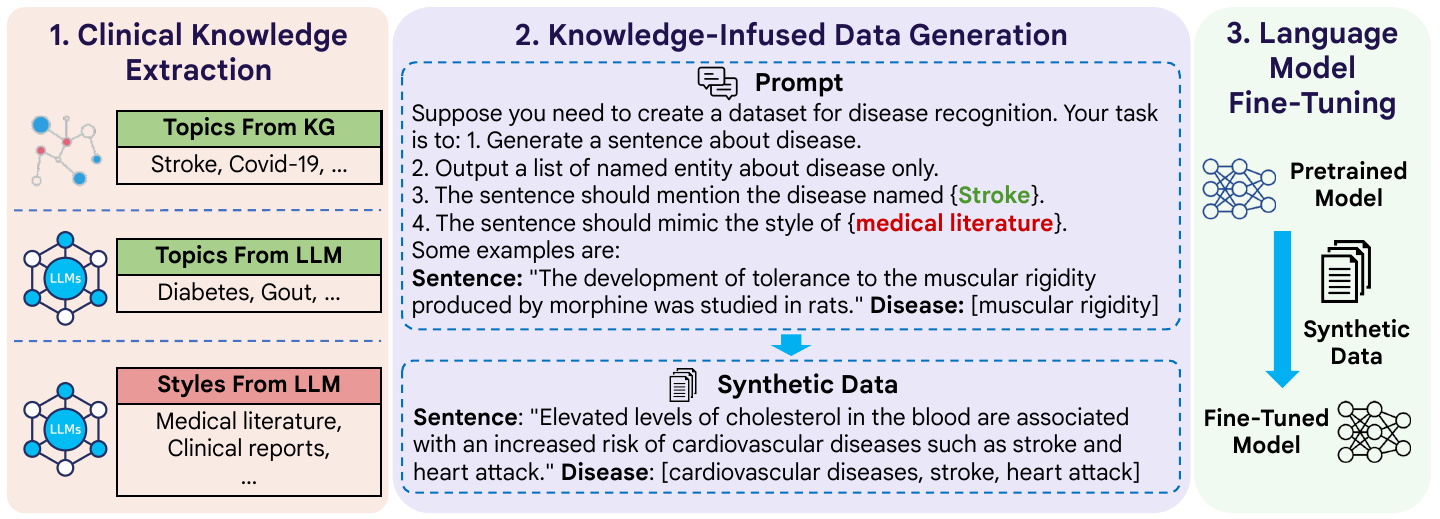}
    \caption{The overview of \ours. \vspace{-2ex}
    % The left orange panel illustrates the knowledge extraction part. The middle purple panel shows the synthetic data generation module. The right green one is the fine-tuning step.}
    }
    % \vspace{-ex}
    \label{fig:overall}
\end{figure*}
% \vspace{-0.5ex}
\section{Knowledge Infused Data Generation}
% \vspace{-0.5ex}
Section \ref{sec:preliminary} highlights the necessity of domain-tailored knowledge for clinical synthetic data generation. In pursuit of this, we present {\ours}, a knowledge-informed framework for clinical data generation. The overview of {\ours} is shown in Figure~\ref{fig:overall}. 
This two-step methodology harnesses the emergent capabilities of LLMs and external knowledge from KGs to facilitate the synthesis of clinical data, even with few-shot examples only.  
% \hj{Perhaps changing this last sentence (maybe it can be put somewhere else later when readers have finished Sec 4.1 and 4.2) to an overall description of the whole pipeline helps to understand the fine-tuning process (and the role of each module) better. I found it may be implicit in realizing that all our generated synthetic data are then used to fine-tune a pre-trained model at this point.}. 

\subsection{Clinical knowledge extraction}
Contrary to previous studies~\citep{ye2022zerogen,ye2022progen,meng2023tuning} which employ generic queries $p_j$ to prompt LLMs for text generation, {\ours} emphasizes refining clinically informed prompts. This approach aims to extract rich clinically relevant knowledge from parametric (\eg LLMs) or non-parametric sources (\eg knowledge graphs) and tailor it to clinical NLP tasks.
% Different from prior research~\citep{ye2022zerogen,ye2022progen,meng2022generating,meng2023tuning} that use the simple task-specific queries to prompt Language Model (LLM) for data generation,
% {\ours}'s primary focus is to optimize clinically informed prompts to better harvest the clinical-relevant knowledge from LLMs and adapt them to clinical NLP tasks.
% emphasize on the initial ground of contextual domain knowledge.
To realize this, our modeling contains two dimensions including \emph{clinical topics} $\cT$ and \emph{writing styles} $\cW$, which are integrated into the original prompts to infuse domain-specific knowledge. 
% added in revise
The \emph{Clinical topic} refers to a \emph{clinical entity} (e.g., disease) or \emph{relation} (e.g., the relationship between diseases and medications), which is usually a phrase, while the \emph{writing style} is a phrase that depicts the tone, and overall presentation of the text.
%%%%%%%%%
By composing different topics and writing styles together, {\ours} provide a diverse suite of prompts, resulting in a wider spectrum of text produced from the LLM $\cM$.
For details of prompt formats across various tasks, please see \textbf{Appendix~\ref{sec:prompt_format}}.
% Examples of the generated topics and writing styles are in \textbf{Appendix~\ref{sec:generated_details}}.

% harness the synergistic potential of Knowledge Graphs (KGs) and Large Language Models (LLMs) in constructing a candidate set enriched with prior knowledge, specifically (1) the sampling of pertinent entities or relations from external knowledge graphs (KGs), and (2) the extraction of relevant hints through queries to Language Models (LLMs).\ran{style}
% \vspace{-0.4ex}
\subsubsection{Clinical Topics Generation}
% \vspace{-0.2ex}
We provide two choices to generate clinical topics $\cT$-- one is to sample related entities or relations from external KG, and the other is to query relevant knowledge from LLM.

\noindent \textbf{Topics $\cT_{\operatorname{KG}}$ sampled from Non-Parametric KGs.} 
Healthcare KGs offer 
% comprehensive view of medical concepts and their relationships.
a rich collection of medical concepts and their complex relationships, 
% and serve as a promising tool for organizing 
which organizes medical knowledge in a structured way~\citep{li2022graph}. 
In our study, we employ the integrative biomedical knowledge hub (iBKH) as the KG~\citep{su2023biomedical} $\cG$ to generate topics $\cT_{\operatorname{KG}} \sim \operatorname{query}(\cG)$
due to its broad coverage over clinical entities. 
To illustrate, for the Disease Recognition task (NCBI,~\citet{ncbi-disease}), we extract all disease nodes $e$ from the iBKH to bolster the medical information as 
$\cT_{\operatorname{KG}}^{\operatorname{NCBI}}\sim \operatorname{query}(\cG_{\operatorname{disease}})$, 
% \hj{$\operatorname{query}(\cG_{\operatorname{disease}}, \cdot)$?}, 
$\cG_{\operatorname{disease}}=\{e\in \cG|\operatorname{type}(e)=\operatorname{disease}\}$.
% $\cT_{\operatorname{KG}}^{\operatorname{NCBI}} = \{e\}\sim\operatorname{query}(\cG, \operatorname{type}(e)=\operatorname{disease})$. 
As another example, we retrieve links between chemicals  $c$ and diseases $d$  for the chemical and disease relation extraction (CDR,~\citet{cdr_dataset}) as
$\cT_{\operatorname{KG}}^{\operatorname{CDR}}\sim \operatorname{query}(\cG_{\operatorname{relation\_{cd}}})$, 
% \hj{$\operatorname{query}(\cG_{\operatorname{relation\_{cd}}}, \cdot)$?}, 
$\cG_{\operatorname{relation\_{cd}}}=\{\langle c,r,d \rangle\in \cG|\operatorname{type}(r)=\operatorname{has\_{relation}}\}$. 
% $\cT_{\operatorname{KG}}^{\operatorname{CDR}} = \{\langle d_1,r,d_2 \rangle\}\sim\operatorname{query}(\cG, {r}=\operatorname{has\_relation})$. 
By injecting information from the KG into the data generation step, we ensure the generated samples are more contextually accurate and semantically rich.

\noindent \textbf{Topics $\cT_{\operatorname{LLM}}$ queried from Parametric LLMs.} 
Pre-trained on extensive text corpora such as medical literature, LLMs provide an alternative method for acquiring domain knowledge.
% Large Language Models (LLMs) offer another viable avenue for acquiring foundational domain knowledge, owing to their extensive training on diverse text corpora, including the vast expanse of medical literature. 
Specifically, we aim to harness the rich clinical domain knowledge encoded in ChatGPT (\texttt{gpt-3.5-turbo-0301}) to augment the prompt. 
The incorporated prior knowledge from LLMs focus on entity types that hold relevance within clinical text datasets, including \emph{diseases}, \emph{drugs}, \emph{symptoms}, and \emph{side effects}.
For each of entity types $e_i$, we prompt the LLMs by formulating inquiries $q(e_i)$, \eg, ``\emph{Suppose you are a clinician and want to collect a set of <Entity Type>. Could you list 300 entities about <Entity Type>?}''. These crafted conversational cues serve as effective prompts to retrieve clinically significant entities from the rich domain knowledge within LLMs as 
$\cT_{\operatorname{LLM}} \sim \rho\left(\cdot;q(e_i)\right)$. 
For each entity type, we generate 300 entities for synthetic data generation.

% \vspace{-0.4ex}
\subsubsection{Clinical Writing Styles Suggestion}
% \vspace{-0.1ex}
\textbf{Styles suggested by LLMs.} To address the limitations mentioned in Sec~\ref{sec:limitations} and introduce a diverse range of writing styles $\cW$ for synthetic samples, we leverage the powerful LLM to suggest candidate writing styles for each task. Specifically, for the task $i$, we incorporate task names $n_i$ into our prompts $p^{\operatorname{style}}_{i}$ (e.g., \emph{disease entity recognition}, \emph{recognizing text entailment}) and integrate few-shot demonstrations $d^{\operatorname{style}}_{i}$. We then engage LLM in suggesting several potential sources, speakers, or authors of the sentences as $\cW \sim \rho\left(\cdot; [p^{\operatorname{style}}_{i}, d^{\operatorname{style}}_{i}]\right)$. 
% \hj{since $n_i$ is included in $p^{\operatorname{style}}_{i}$ as illustrated above, here should be $\cW \sim \rho\left(\cdot; [p^{\operatorname{style}}_{i}, d^{\operatorname{style}}_{i}]\right)$?}
% See Appendix~\ref{sec: style_prompt} for a detailed prompt. 
Responses such as ``\emph{medical literature}" or ``\emph{patient-doctor dialogues}" are augmented into the prompts to imitate the writing styles found in real datasets. 
% \joyce{I'm not quite sure what this quite looks like. Is there an example in the appendix/supplemental part? Seems you're downplaying it here otherwise.}
% explicitly query 

% \vspace{-0.5ex}
\subsection{Knowledge-infused Data Generation}
% \vspace{-0.5ex}
% After building the candidate set with one of the previous approaches, we randomly sample one keyword each time and augment it into the prompt. 
% % An example of the prompt on xxx dataset is shown in Figure xxx. 
% For example, a keyword for the NCBI dataset could be ``\texttt{stroke}", then we enrich the prompt for querying ChatGPT by adding an additional sentence ``\texttt{generate a sentence about stroke}" (see Figure/Appendix for full prompt). 
With the generated topics and styles, the key challenge becomes how to leverage them to extract rich clinical information from the LLM for improving synthetic data quality.
% hat, if used wisely, can be potentially leveraged for generalizing over limited context.
% if they are guided effectively. 
Directly putting all the elements to enrich the prompt is often infeasible due to the massive size of entities.
% and over-complicated 
To balance informativeness as well as diversity, we propose a knowledge-infused strategy, where for each class label name $j\in\cY$, the collected clinical topics and writing styles serve as the base unit. 
% \hj{`class' is somewhat sudden and vague here, maybe `class label for each specific task'?}
% for leading LLMs towards clinical domains.  
% To elucidate, each extracted hinting keyword from the candidate set takes a role in shaping a clinically informed prompt structure. 
In each step, we randomly sample a topic $t \in \cT$ and a writing style $w \in \cW$ from the candidate set to augment the prompt for class $j \in \cY$  as 
$p^{\operatorname{Clin}}_j(t, w) = [p_j, t, w]$. 
Then, we use the augmented prompt $p^{\operatorname{Clin}}_j(t, w)$ together with the few-shot demonstrations $\cD$ to generate the synthetic dataset $\tilde{\cD}_{\operatorname{Clin}}$ as 
\begin{equation}
\setlength{\abovedisplayskip}{6pt}
\setlength{\belowdisplayskip}{6pt}
\tilde{\cD}_{\operatorname{Clin}} \sim \rho_{j\sim\cY, t\sim \cT, w\sim\cW}\left(\cdot; \left[p_j, t, w\right], \cD\right).
\nonumber
\end{equation}
% For instance, for the Disease Recognition (NCBI) task, consider a clinical entity like ``\texttt{stroke}" and a writing style such as ``\texttt{medical literature}". We enrich the prompt query for LLM by appending ``\texttt{generate a sentence about stroke in the style of medical literature}" as a generation guidance. 
% For a comprehensive view of the prompt formats across various tasks, please refer to Appendix~\ref{sec:prompt_format}. 
Despite its simplicity, this strategy enjoys several merits: 
(1) \emph{Clinical infusion}: the clinical context is incorporated into the prompts to directly guide data generation; 
(2) \emph{Diversity}: it encourages data  diversity via dynamically composing different entities and writing styles into prompts; 
(3) \emph{Flexibility}: it is compatible with different sources of $\cT$ and $\cW$ without reliance on specific knowledge formats. 
Consequently, the quality and clinical relevance of the generated synthetic data are enhanced.  
While some works focus on prompt optimization for data generation or other NLP tasks, they typically utilize a fixed prompt and optimize this prompt format, which is orthogonal to \ours{}.
% \hj{do we also need to mark other important references to Appendix bold, similarly? I felt like some prompt example references above also help to understand and are important}. 
% While several works also focus on prompt optimization for data generation or other NLP tasks, these approaches mainly use a \emph{fixed} prompt and focus on optimizing the prompt format and demonstrations, which is orthogonal to \ours{}. \textbf{We provide comparisons in Appendix~\ref{sec:diff_prompt_design}}. 
% For a comprehensive overview of prompt formats across different tasks,
% please refer to Appendix~\ref{sec:prompt_format}. 
%  unlike existing LLM-based data generation and prompt optimization methods which use a \emph{fixed} prompt, 

\subsection{Language Model Fine-tuning}
After generating synthetic data $\tilde{\cD}$, we fine-tune a pre-trained classifier $\cC_{\theta}$ for each downstream task. Following \citet{meng2023tuning}, we first fine-tune $\cC_{\theta}$ on $\cD$ with standard supervised training objectives on few-shot examples (denoted as $\ell(\cdot)$) in Stage 1, then on synthetic data $\tilde{\cD}$ in Stage 2 as  
% \vspace{-1ex}
% \begin{equation}
% \begin{small}
\begin{align}
\setlength{\abovedisplayskip}{6pt}
\setlength{\belowdisplayskip}{6pt}
    \label{eq:stage1}
    \theta^{(1)} &= \min_{\theta}~\mathbb{E}_{(x, y) \sim \cD} \ell\left( f(x; \theta), y \right), \nonumber \\
    \theta^{(2)} &=  \min_{\theta}~\mathbb{E}_{(\tilde{x}, \tilde{y}) \sim \tilde{\cD}} \ell\left( f(\tilde{x}; \theta), \tilde{y} \right),  \theta_{\text{init}} = \theta^{(1)}.
    \nonumber
\end{align} 
% \end{small}
% \end{equation}
It's important to highlight that we strictly follow a standard fine-tuning process and avoid using any extra techniques: (1) for standard classification tasks, $\ell(\cdot)$ is the cross-entropy loss; (2) for multi-label classification tasks, $\ell(\cdot)$ is the binary cross-entropy loss; 
(3) for token-level classification tasks, we stack an additional linear layer as the classification head and  $\ell(\cdot)$ is the token-level cross-entropy loss. 
% This is to ensure methodological consistency and transparency throughout the evaluation across different methods and tasks. 
The design of \emph{advanced learning objectives} as well as \emph{data mixing strategies}, while important, are orthogonal to the scope of this paper.

% It is important to underscore that, in pursuit of a rigorous quality evaluation of the synthetic data, we rigorously adhere to a conventional fine-tuning pipeline, without any auxiliary techniques. Our chosen learning objective is the minimization of the cross-entropy loss against the task-specific target, ensuring methodological consistency and transparency throughout the evaluation across different methods and tasks.

% \subsubsection{Style}

% \vspace{-0.7ex}
\section{Empirical Evaluation}
% \vspace{-0.4ex}
Given our focus on data generation, our major interest lies in faithfully evaluating different synthetic text generation approaches under few-shot scenarios, rather than competing in a ``\emph{state-of-the-art}" race with general few-shot NLP methods. 
The following questions particularly intrigue us:
\textbf{RQ1}: How does {\ours} perform when compared with baselines on different downstream tasks?  
% \textbf{RQ2}: How do factors such as LLM generators and the size of synthetic data affect the performance of {\ours}? 
\textbf{RQ2}: What impact do factors like LLM generators and synthetic data size have on the performance of {\ours}?
\textbf{RQ3}: How is the quality of the synthetic data generated by {\ours} and baselines?
% These questions are addressed in Sec~\ref{sec:model_perf}, Sec~\ref{sec:ablation} and Sec~\ref{sec:quality_analysis}, respectively.

\begin{table*}[t]
% \floatconts
  \caption{Experimental results aggregated by tasks. \textbf{Bold} and \underline{underline} denote the best and second-best results. $\dagger$: Models exclusive to NER tasks. $*$: Since the two $\dagger$ models only report results on two NER datasets, we report the average performance on those two datasets for a fair comparison. 
  "Supervised-Full" and "Supervised-Few" denote the results using the original dataset and using only the few-shot examples as training data, respectively.
  \vspace{-1ex}}
  \renewcommand\arraystretch{0.95}
  \resizebox{0.98\linewidth}{!}{
  \begin{tabular}{lcc|ccccc|ccccc}
  \toprule
  \multirow{3.5}{*}{\bf Task}  & \multicolumn{2}{c|}{\textit {Single-Sentence Tasks}} &  \multicolumn{5}{c|}{\textit{Sentence-Pair Tasks}} & \multicolumn{5}{c}{\textit{Token Classification Tasks}} \\
  \cmidrule(lr){2-3} \cmidrule(lr){4-8} \cmidrule(lr){9-13} 
  & \bfseries Text Class (2) & \bfseries RE (3) & \bfseries NLI (3) & \multicolumn{2}{c}{\textbf{Fact Verification (2)}} & \bfseries STS (1)& \bfseries QA (2)& \multicolumn{2}{c}{\textbf{NER (4)}} & \multicolumn{3}{c}{\textbf{MedAttr (1)}} \\
  % \midrule
  \cmidrule(lr){2-2} \cmidrule(lr){3-3} \cmidrule(lr){4-4} \cmidrule(lr){5-6} \cmidrule(lr){7-7} \cmidrule(lr){8-8} \cmidrule(lr){9-10} \cmidrule(lr){11-13}
  & F1 & F1 & Acc & Acc & F1 & Acc & Acc & F1 & F1-subset$^*$ & P & R & F1\\
  \midrule
  \multicolumn{12}{l}{\textbf{PubMedBERT$_{\texttt{Base}}$}} \\
  \midrule
  % \blue{Supervised-Full(SOTA)} \\
  Supervised-Full & 77.01 & 77.34 & 79.20 & 67.58 & 65.49 & 75.70 & 74.70 & 89.67 & 87.27 & --- & --- & ---\\
  Supervised-Few & 18.61 & 43.89 & 44.64 & 29.43 & 27.10 & 55.70& 54.74 & 39.41 & 34.12 & 38.11 & 43.82 & 40.77 \\
  \midrule
  DA-Word Sub~\shortcite{checklist} & 40.74 & 38.14 & 55.08 & 28.86 & 25.83 & 54.40 & 53.58& 44.30 & 40.41 & 40.25 & 47.65 & 43.64\\
  DA-Back Trans~\shortcite{uda} & 47.24 & --- & 54.30 & 32.15 & 28.04 & 55.80 & 53.28 & --- & --- & --- & --- & ---\\
  DA-Mixup~\shortcite{chen2020mixtext,seqmix} & 45.09 & 43.37 & 53.52 & 32.78 & 29.12 & 58.20 & 51.91 & 42.20 & 37.65 & 42.37 & 48.96 & 45.43\\
  DA-Transformer~\shortcite{melm,kumar2020data}  & 41.02 & 47.56 & 55.71 & 35.32 & 31.77 & 58.80& 56.36 & 44.75 & 39.66 & 37.82 & 44.28 & 40.80\\
  LightNER$^\dagger$~\shortcite{lightner} & --- & --- & --- & --- & --- & --- & --- &  ---- & 39.49 & --- & --- & ---\\
  % DA-MELM$^\dagger$& --- & --- & --- & --- & --- & --- & 44.75 & 39.66 & 37.82 & 44.28 & 40.80\\
  KGPC$^\dagger$~\shortcite{chen2023few} & --- & --- & --- & --- & --- & --- & --- &--- &  51.60 & --- & --- & ---\\
  \midrule
  ZeroGen~\shortcite{ye2022zerogen,meng2022generating} & 59.02 & 63.84 & 55.96 & 35.30 & 32.50 & 68.35 &  61.89 & 56.97 & 48.26 & 52.80 & 49.53 & 51.11\\
  DemoGen~\shortcite{meng2023tuning,gpt3mix} & 64.09 & 67.46 & 59.80 & 40.30 & 35.95 & 70.85 & 62.01 & 60.16 & 53.91 & 58.15 & 56.84 & 57.49\\
  ProGen~\shortcite{ye2022progen} & 65.16 & 67.23 & 59.57 & 37.71 & 34.54 & 69.30 & 60.74 & 60.49 & 55.11 & 57.76 & 58.57 & 58.16\\  
S3~\shortcite{wang2023lets} & 65.12 & 67.60 & 61.36 & 40.17 & 36.44 & 70.20 &  63.58 & 60.36 & 54.25 & 56.21 &	63.60 &	59.68 \\
  \midrule
  % \hline
  \rowcolor{teal!10} {\ours} w/ KG & \underline{67.15} & \underline{69.01} & \underline{64.89} & \underline{43.83} & \underline{39.43} & \underline{72.20} & \bf 71.49 & \textbf{64.26} & \textbf{60.11} & \textbf{71.75} & \underline{65.20} & \textbf{68.32}\\
  \rowcolor{teal!10} {\ours} w/ LLM & \textbf{67.82} & \textbf{70.06} & \textbf{67.24} & \textbf{46.50} & \textbf{41.46} & \textbf{73.30} & \underline{69.60} & \underline{63.17} & \underline{58.49} & \underline{68.19} & \textbf{66.79} & \underline{67.48}\\
  % \midrule
  \rowcolor{gray!15} Performance Gain & 4.08\% & 3.63\% & 9.58\% & 15.38\% & 13.77\% & 3.47\% & 12.44\% & 6.23\% & --- & --- & --- & 14.48\% \\
  \midrule
  \multicolumn{12}{l}{\textbf{PubMedBERT$_{\texttt{Large}}$}} \\
  \midrule
  Supervised-Full & 80.06 & 79.64 & 82.65 & 72.97 & 69.23 & 78.80 &80.37 &  90.15 & 87.68 & --- & --- & ---\\
  Supervised-Few & 17.86 & 52.68 & 50.00& 40.90 & 30.50 & 59.73 & 59.50& 42.84 & 37.57 & 41.30 & 45.02 & 43.08 \\
  \midrule
  DA-Word Sub~\shortcite{checklist} & 43.99 & 44.35 & 57.66 & 35.51 & 31.95 & 55.30 & 58.57& 46.67 & 43.70 & 46.77 & 43.52 & 45.09\\
  DA-Back Trans~\shortcite{uda} & 50.98 & --- & 58.39 & 34.12 & 31.36 & 56.40 &57.19 & --- & --- & --- & --- & ---\\
  DA-Mixup~\shortcite{chen2020mixtext,seqmix} & 46.74 & 50.97 & 57.35 & 34.01 & 31.10 & 58.50 & 56.68 & 46.69 & 43.01 & 41.25 & 52.09 & 46.04\\
  DA-Transformer~\shortcite{melm,kumar2020data} & 44.41 & 46.12 & 58.94 & 35.09 & 30.95 & 58.10 & 59.30 & 46.94 & 43.50 & 43.36 & 45.78 & 44.54\\
  % LightNER$^\dagger$~\shortcite{lightner} & --- & --- & --- & --- & --- & --- & --- & --- &  --- & --- & --- & ---\\
  % DA-MELM$^\dagger$ & --- & --- & --- & --- & --- & --- & 46.94 & 43.50 & 43.36 & 45.78 & 44.54\\
  % KGPC$^\dagger$~\shortcite{chen2023few} & --- & --- & --- & --- & --- & --- & ---& --- & --- & --- & --- & ---\\
  \midrule
  ZeroGen~\shortcite{ye2022zerogen,meng2022generating} & 61.51 & 65.18 & 63.47 & 41.12 & 36.10 & 72.69 &66.02 & 57.79 & 49.10 & 54.04 & 51.40 & 52.69\\
  DemoGen~\shortcite{meng2023tuning,gpt3mix} & 64.97 & 68.65 & 64.58 & 42.61 & 38.69 & 74.37 & 65.04 & 61.43 & 55.61 & 62.67 & 61.02 & 61.83\\
  ProGen~\shortcite{ye2022progen} & 65.01 & 69.23 & 63.32 & 42.79 & 38.63 & 74.90 & 63.27 & 62.47 & 57.31 & 57.21 & 63.70 & 60.28\\
 S3~\shortcite{wang2023lets} & 64.33 & 69.65 & 65.07 & 41.76 & 37.72 & 73.20 & 66.33 & 61.97 & 56.29 & 63.07	&62.72&	62.89 \\

  \midrule
  % \hline
  \rowcolor{teal!10} {\ours} w/ KG & \underline{66.76} & \underline{71.47} & \textbf{70.90} & \underline{48.62} & \underline{42.45} & \underline{75.40} & \bf 73.94 & \textbf{65.48} & \textbf{62.23} & \underline{70.96} & \textbf{69.66} & \textbf{70.30}\\
  \rowcolor{teal!10} {\ours} w/ LLM & \textbf{67.61} & \textbf{72.81} & \underline{70.50} & \textbf{49.51} & \textbf{43.72} & \textbf{76.21} &  \underline{73.40} &\underline{65.36} & \underline{61.89} & \textbf{71.61} & \underline{66.86} & \underline{69.15}\\
  \rowcolor{gray!15} {Performance Gain} & 4.00\% & 4.54\% & 8.96\% & 15.70\% & 13.00\% & 3.47\% & 11.47\% & 1.76\% & --- & --- & --- & 11.78\% \\
  \bottomrule
  \end{tabular}
   }
  \label{tab:main-table}
\end{table*}

\subsection{Experiment Setup}
We conduct experiments in the few-shot settings with 5 examples for each class. We employ ChatGPT~\citep{chatgpt} (\texttt{gpt-3.5-turbo-0301}) as the LLM generator $\cM$\footnote{Studies on using Medical LLMs are in Appendix \ref{sec:med_llm_gen}.} and \textbf{maintain the same amount of synthetic training data for both {\ours} and baselines for a fair comparison.} The pre-trained PubMedBERT~\citep{gu2021domain} is then applied to fine-tune on the synthetic data for both {\ours} and baselines, where we consider both the \texttt{Base} and \texttt{Large} model.

\noindent \textbf{Datasets and Tasks.}
We undertake a comprehensive evaluation of \textbf{18 datasets} across a diverse array of tasks in clinical NLP benchmarks~\citep{blue,fries2022bigbio}: 2 text classification, 3 relation extraction (RE), 3 natural language inference (NLI), 2 fact verification, 2 question answering (QA), 1 sentence similarity (STS), 4 Named Entity Recognition (NER), and 1 attribute extraction datasets. 
Please see Appendix~\ref{sec:dataset_description} for descriptions and the statistics of each dataset.

% The evaluation tasks and datasets are summarized in Table \ref{tab:datastats}. Note that the number of training samples indicates the size of the \textit{original} training set. Please see Appendix~\ref{sec:dataset_description} for detailed dataset descriptions.

\noindent  \textbf{Baselines.}
We compare {\ours} with \textbf{10 baselines} in total, including
6 data augmentation and 4 LLM-based data generation techniques. 
See Appendix~\ref{sec:baseline_details} for their descriptions. 

\noindent \textbf{Implementation Details.}
\label{sec:implementation_details}
For implementation, we use PyTorch~\citep{paszke2019pytorch} and HuggingFace~\citep{wolf2019huggingface}. For each dataset, we randomly sample 5 examples from each class to provide few-shot demonstrations and keep a validation set of the same size. 
During the data generation process when we call the ChatGPT APIs~\citep{chatgpt}, we set the parameter $\operatorname{top\_p}=1.0$ and temperature $t=1.0$ to balance between the quality of the generated text as well as diversity~\citep{chung-etal-2023-increasing,yu2023large}\footnote{We do not further increase $t$, as previous analysis  \citep{chung-etal-2023-increasing,yu2023large} has shown that increasing $t$ to larger value does not help with additional performance gain.}. 
In the experiments, We generate 5000 synthetic training data for both {\ours} and the baselines and report the average performance over 3 random seeds for all the results. 
With the generated synthetic dataset, we follow the common few-shot learning setting~\citep{perez2021true} to train all the models for 6 epochs and use the model with the best performance on the validation set for evaluation. 
During the PubMedBERT fine-tuning, we adopt AdamW~\citep{loshchilov2017decoupled} for optimization with a linear warmup of the first 5\% steps and linear learning rate decay. The learning rate is set to 2e-5 for \texttt{Base} and 1e-5 for \texttt{Large}, and the maximum number of tokens per sequence is 256.

\begin{figure*}[t]
\vspace{-2.5ex}
    \centering
    \begin{minipage}{0.48\textwidth}
        \centering
        \subfigure[HOC]{
            \includegraphics[width=0.48\textwidth]{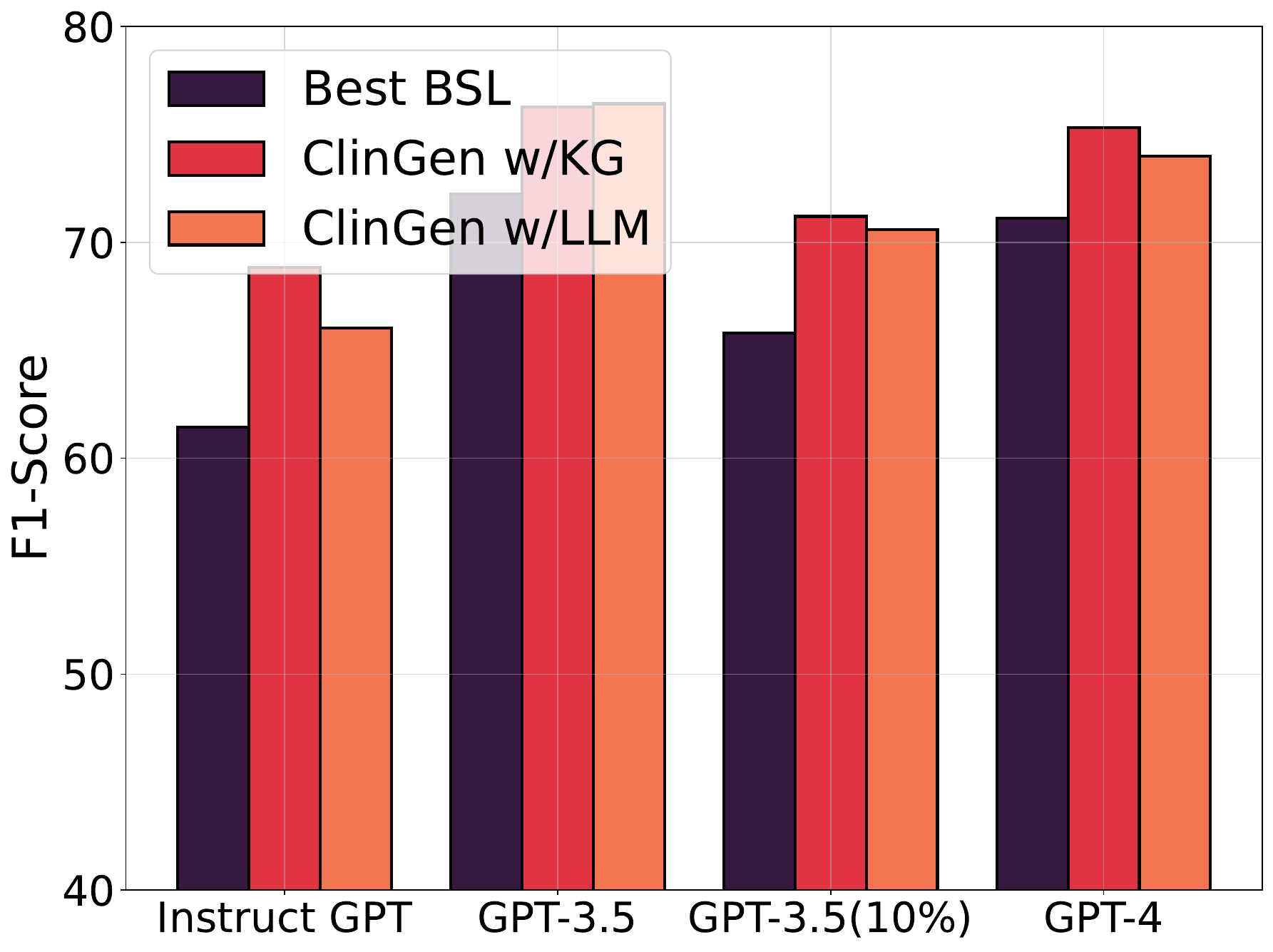}
            \label{fig:generator-HOC}
        } \hspace{-3mm}
        \subfigure[MEDIQA-RQE]{
            \includegraphics[width=0.48\textwidth]{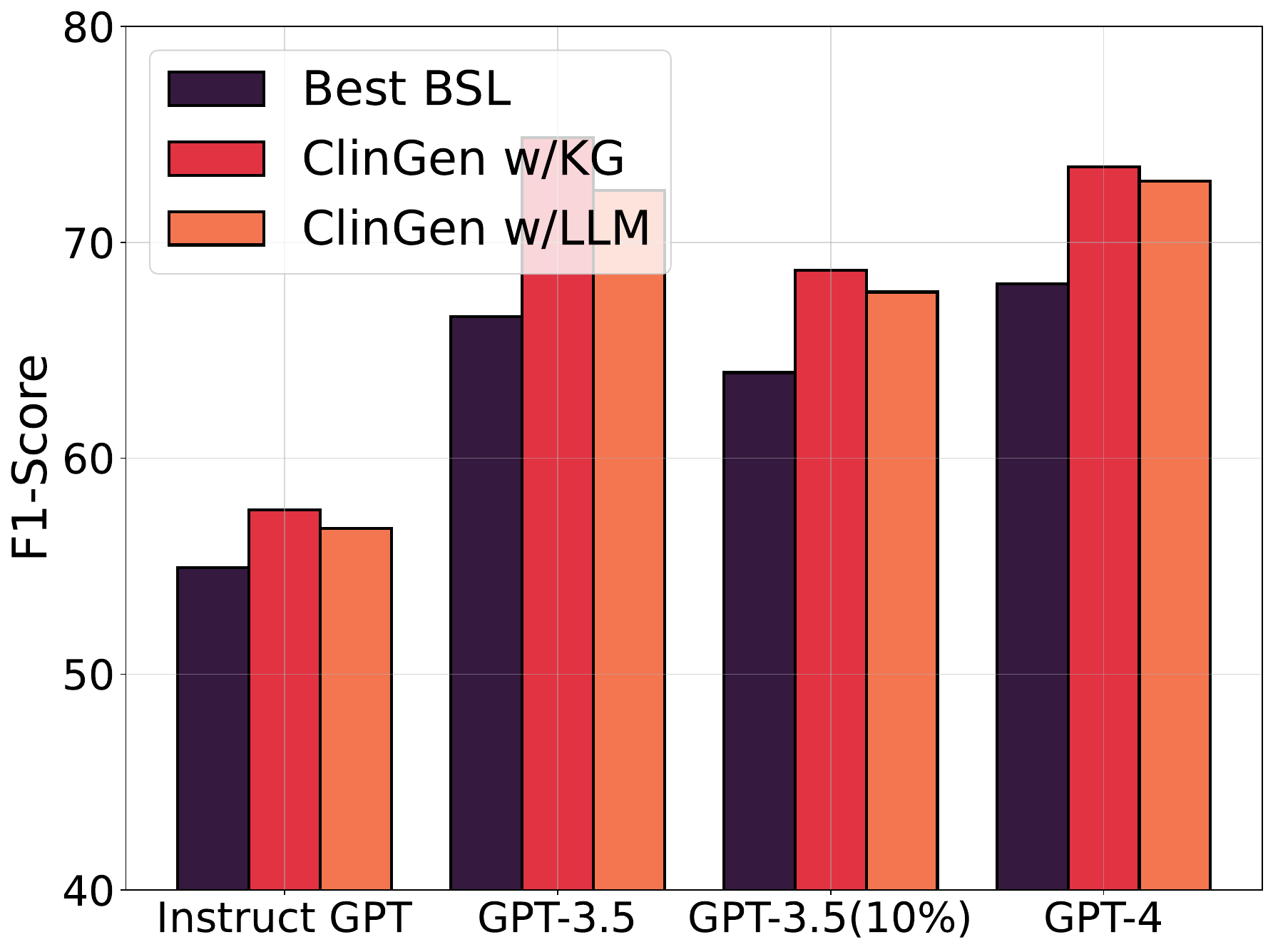}
            \label{fig:generator-MEDIQA-RQE}
        }
        \vspace{-2ex}
        \RawCaption{\caption{Different generators at \texttt{Base}.}\label{fig:generator}}
    \end{minipage}%
    % \begin{minipage}{0.33\textwidth}
    %     \centering
    %     \vspace{3.5mm}
    %     \includegraphics[width=0.96\textwidth]{figures/size-HOC.pdf}
    %     \caption{\textit{Accuracy vs. Confidence score.}}
    %     \label{fig:confscore}
    % \end{minipage}
    \begin{minipage}{0.48\textwidth}
        \centering
        \subfigure[HOC]{
            \includegraphics[width=0.5\textwidth]{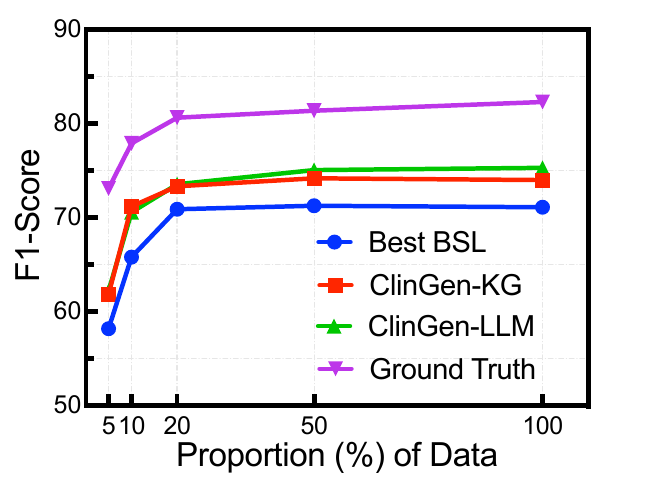}
            \label{fig:size-HOC}
        } \hspace{-6mm}
        \subfigure[MEDIQA-RQE]{
            \includegraphics[width=0.5\textwidth]{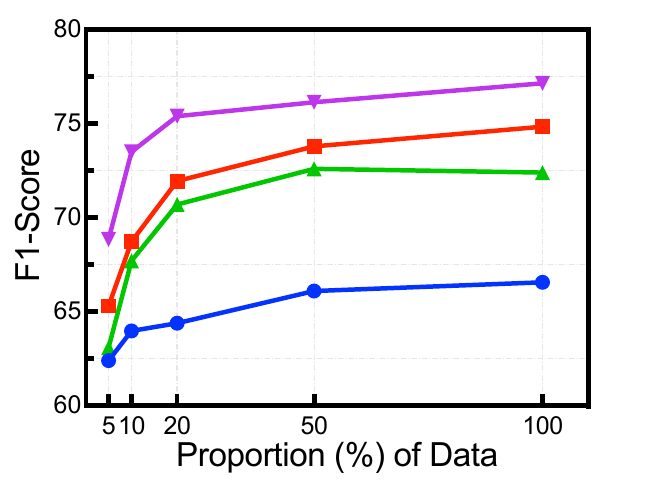}
            \label{fig:size-Mediqa-rqe}
        }
        \vspace{-2ex}
        \RawCaption{\caption{Different proportion of data at \texttt{Base}.}\label{fig:size-synthetic}}
    \end{minipage}%
    \vspace{-0.5ex}
\end{figure*}

\begin{table*}[t]
% \floatconts
% \vspace{-1ex}
  \caption{Comparison between prompting LLM for inference and {\ours} at \texttt{Large} scale.\vspace{-1ex}}
    \renewcommand\arraystretch{0.95}
  \resizebox{0.99\linewidth}{!}{
  \begin{tabular}{lcccccccccccccc}
  \toprule
  & \bfseries HOC & \multicolumn{3}{c}{\textbf{GAD}} & \bfseries ChemProt & \bfseries MEDIQA-RQE & \multicolumn{2}{c}{\textbf{PUBHEALTH}} & \multicolumn{3}{c}{\textbf{NCBI-Disease}} & \multicolumn{3}{c}{\textbf{CASI}}\\
  % \midrule
  \cmidrule(lr){2-2} \cmidrule(lr){3-5} \cmidrule(lr){6-6} \cmidrule(lr){7-7} \cmidrule(lr){8-9} \cmidrule(lr){10-12} \cmidrule(lr){13-15}
  & F1 & P & R & F1 & F1 & ACC & ACC & F1 & P & R & F1 & P & R & F1\\
  \midrule
  
  ChatGPT Inference~\scriptsize{(\citeauthor{chatgpt})} & 68.76 & 84.21 & \textbf{97.46} & 90.35 & 49.42 & 74.31 & \textbf{69.50} & \textbf{52.47} & 46.62 & 52.31 & 49.30 & 48.82 & \textbf{74.75} & 59.07\\
  \blue{PMC-LLaMa-13B Inference}~\scriptsize{(\citeauthor{wu2023pmcllama})} & 50.07 & 89.61 & 81.18 & 85.19  & 33.35 & 52.17 & 48.01 & 32.84 & 27.11 & 23.97 & 25.44 & 56.38 & 36.87 & 41.58
  \\
 \blue{MedAlpaca-13B Inference}~\scriptsize{(\citeauthor{han2023medalpaca})} & \blue{40.44} & \blue{71.95} & \blue{72.48} & \blue{72.21} & \blue{31.29} & \blue{58.12} & \blue{55.40} & \blue{34.63} & \blue{44.69} & \blue{31.16} & \blue{27.85} & \blue{52.51} & \blue{49.16} & \blue{51.64}
 \\
  \midrule
  % \hline
  \rowcolor{teal!10} {\ours} w/ KG & 77.71 & 94.30 & 89.09 & \textbf{91.62} & 60.12 & \textbf{79.92} & 50.20 & 41.26 & \textbf{62.46} & \textbf{64.08} & \textbf{63.26} & 70.96 & 69.66 & \textbf{70.30} \\
  \rowcolor{teal!10} {\ours} w/ LLM & \textbf{78.14} & \textbf{95.08} & 86.14 & 90.39 & \textbf{63.05} & 77.36 & 52.96 & 43.31 & 61.12 & 60.16 & 60.64 & \textbf{71.61} & 66.86 & 69.15 \\
  \bottomrule
  \end{tabular}
  }
  \label{tab:gpt_inference}
  \vspace{-1ex}
\end{table*}

\subsection{Model Performance with Synthetic Data}
\label{sec:model_perf}
Table~\ref{tab:main-table} summarizes the experimental results. 
% We also show the performance using the original training data and few-shot examples, denoted as ``\emph{Supervised-Full}" and ``\emph{Supervised-Few}", respectively.
Due to space limits, we report the average performance over all datasets for each task, but provide the detailed results for each dataset in Tables~\ref{tab:single-sent}, \ref{tab:sent-pair}, \ref{tab:token-class} in Appendix~\ref{sec:more_experimental_results}. 
Based on the experimental results, we have the following findings:

\noindent $\diamond$ Our approach, {\ours}, consistently outperforms the baselines across all tasks. The average performance gain over all \textit{main} metrics is 8.7\% at \texttt{Base} scale and 7.7\% at \texttt{Large} scale. 
LLM-based methods outperform traditional DA techniques, showcasing their ability to capture task-specific information from a few examples. 
% LLM-based methods outperform traditional DA techniques, demonstrating their few-shot learning capability. 
DemoGen and ProGen's gains over ZeroGen highlight the positive impact of few-shot examples.
% The performance gains of DemoGen and ProGen over ZeroGen further highlight the positive impact of few-shot examples on data generation.
Despite being one of the most powerful data generation approaches, S3's gains are marginal in the few-shot setting due to its reliance on large validation sets.
% S3, one of the most powerful data generation approach so far, relies on a large set of validation set for effective extrapoliation.
% Under the challenging few-shot setting, S3 only yield a marginal performance gain. 

\noindent $\diamond$ In \textit{token classification tasks}, {\ours} performs better with KG compared to LLM due to the better alignment between the task's target and the generated domain knowledge, where the extracted topics serve as direct labels. 
Conversely, single-sentence and sentence-pair tasks favor LLM-based knowledge extraction. 
This could be because (1) These tasks prioritize sentence comprehension over specific terminologies, and some specialized terms might even impede LLM comprehension. (2) KGs \emph{may not} always contain the required information, e.g., certain relations in chemical/protein relation extraction tasks, limiting performance gains.
% The \textit{single-sentence} and \textit{sentence-pair tasks}, on the other hand, display an advantage for the LLM-based knowledge extraction. 
% This can be attributed to two potential reasons: first, these tasks prioritize understanding entire sentences over specific terminologies, and some specialized terms might even impede LLM comprehension. Second, KGs may not always contain the required information. For example, in a RE task involving chemicals and proteins, some types of relations are absent from the KG, thus the performance gain is rather limited.
% as this information is absent from the KG, we have to extract drug-gene relations as a substitute.

\noindent $\diamond$ Some DA methods are task-specific, limiting their generalizability. For example, LightNER and KGPC are designed for NER. It is also non-trivial to apply Back Translation to NER or RE, as it requires locating related entities in the generated sentence accurately.
In contrast, {\ours} is flexible and can be readily applied to various tasks.

\subsection{Ablation and Parameter Studies}
\label{sec:ablation}
\noindent \textbf{Effect of Different LLM Generators.}
To investigate the impact of various LLMs on {\ours},  
% we leverage other models in the GPT-family as the text generator. Specifically, 
we utilize InstructGPT (\texttt{text-curie-001})~\citep{ouyang2022training} and GPT-4~\citep{gpt4}. Note that we only generate 500 samples in the GPT-4 setting due to budget constraints, but we provide the results of GPT-3.5 with same amount of synthetic samples for a fair comparison. 
From Figure~\ref{fig:generator} we observe that {\ours} generally outperforms the best baseline in all settings.
% except for the NCBI-Disease dataset. 
Additionally, we observe generally improved performance with larger models, as they often have better capabilities to follow our designed instructions for the given prompts. See Appendix~\ref{sec:add_ablation_para} for more results.

\noindent \textbf{Effect of Size of Synthetic Data.}
In Figure~\ref{fig:size-synthetic} (and more in Appendix~\ref{sec:add_ablation_para}), we study the effect of the size of synthetic data. The result shows that {\ours} consistently outperforms the best baseline, using only around 10\% of the synthetic examples. This illustrates that incorporating domain knowledge and increasing the diversity of the prompts could be an effective way to improve the sample efficiency and narrow the gap between the performance of synthetic and ground-truth datasets.
% \paragraph{How Many Clean Data Points is Synthetic Data Worth?}

% \begin{table}[hbtp]
% % \floatconts
%   \caption{Ablation studies on topic extraction and style suggestion at \texttt{Base} scale.}
%   \resizebox{0.5\linewidth}{!}{
%   \begin{tabular}{lcc|cc|cc|cc}
%   \toprule
%   & \multicolumn{2}{c}{\textbf{HOC}} & \multicolumn{2}{c}{\textbf{CDR}} & \multicolumn{2}{c}{\textbf{MEDIQA-RQE}} & \multicolumn{2}{c}{\textbf{NCBI-Disease}}\\
%   % \midrule
%   \cmidrule(lr){2-3} \cmidrule(lr){4-5} \cmidrule(lr){6-7} \cmidrule(lr){8-9}
%   & w/ KG & w/ LLM & w/ KG & w/ LLM & w/ KG & w/ LLM & w/ KG & w/ LLM \\
%   \midrule
%   {\ours} & \textbf{76.28} & \textbf{76.42} & \textbf{61.74} & \textbf{63.34} & \textbf{74.85} & \textbf{72.40} & \textbf{59.46} & \textbf{55.95} \\
%   w/o Styles & 73.25 & 74.40 & 59.10 & 60.15 & 67.21 & 66.50 & 57.97 & 54.70 \\
%   % w/o Topics & 70.86 & 70.86 & 58.51 & 58.51 & 64.87 & 64.87 & 57.30 & 57.30\\
%   w/o Topics & \multicolumn{2}{c|}{70.86} & \multicolumn{2}{c|}{58.51} & \multicolumn{2}{c|}{64.87} & \multicolumn{2}{c}{55.09} \\
%   \bottomrule
%   \end{tabular}
%   }
%   \label{tab:ablation}
% \end{table}
\begin{table}[tp]
% \floatconts
% \vspace{-1ex}
  \caption{Ablation studies on topic extraction and style suggestion at \texttt{Base} scale. \vspace{-1ex}}
  \resizebox{\linewidth}{!}{
  \begin{tabular}{lcc|cc|cc|cc}
  \toprule
  & \multicolumn{2}{c}{\textbf{HOC}} & \multicolumn{2}{c}{\textbf{CDR}} & \multicolumn{2}{c}{\textbf{MEDIQA-RQE}} & \multicolumn{2}{c}{\textbf{NCBI-Disease}}\\
  % \midrule
  \cmidrule(lr){2-3} \cmidrule(lr){4-5} \cmidrule(lr){6-7} \cmidrule(lr){8-9}
  & w/ KG & w/ LLM & w/ KG & w/ LLM & w/ KG & w/ LLM & w/ KG & w/ LLM \\
  \midrule
  \rowcolor{teal!10} {\ours} & \textbf{76.28} & \textbf{76.42} & \textbf{61.74} & \textbf{63.34} & \textbf{74.85} & \textbf{72.40} & \textbf{59.46} & \textbf{55.95} \\
  w/o Styles & 73.25 & 74.40 & 59.10 & 60.15 & 67.21 & 66.50 & 57.97 & 54.70 \\
  % w/o Topics & 70.86 & 70.86 & 58.51 & 58.51 & 64.87 & 64.87 & 57.30 & 57.30\\
  w/o Topics & \multicolumn{2}{c|}{70.86} & \multicolumn{2}{c|}{58.51} & \multicolumn{2}{c|}{69.86} & \multicolumn{2}{c}{55.09} \\
  \bottomrule
  \end{tabular}
  }
  \label{tab:ablation}
  \vspace{-1ex}
\end{table}

\begin{figure*}[tp]
	\centering
	\vspace{-1ex}
	\subfigure[t-SNE plot]{
		\includegraphics[width=0.24\linewidth]{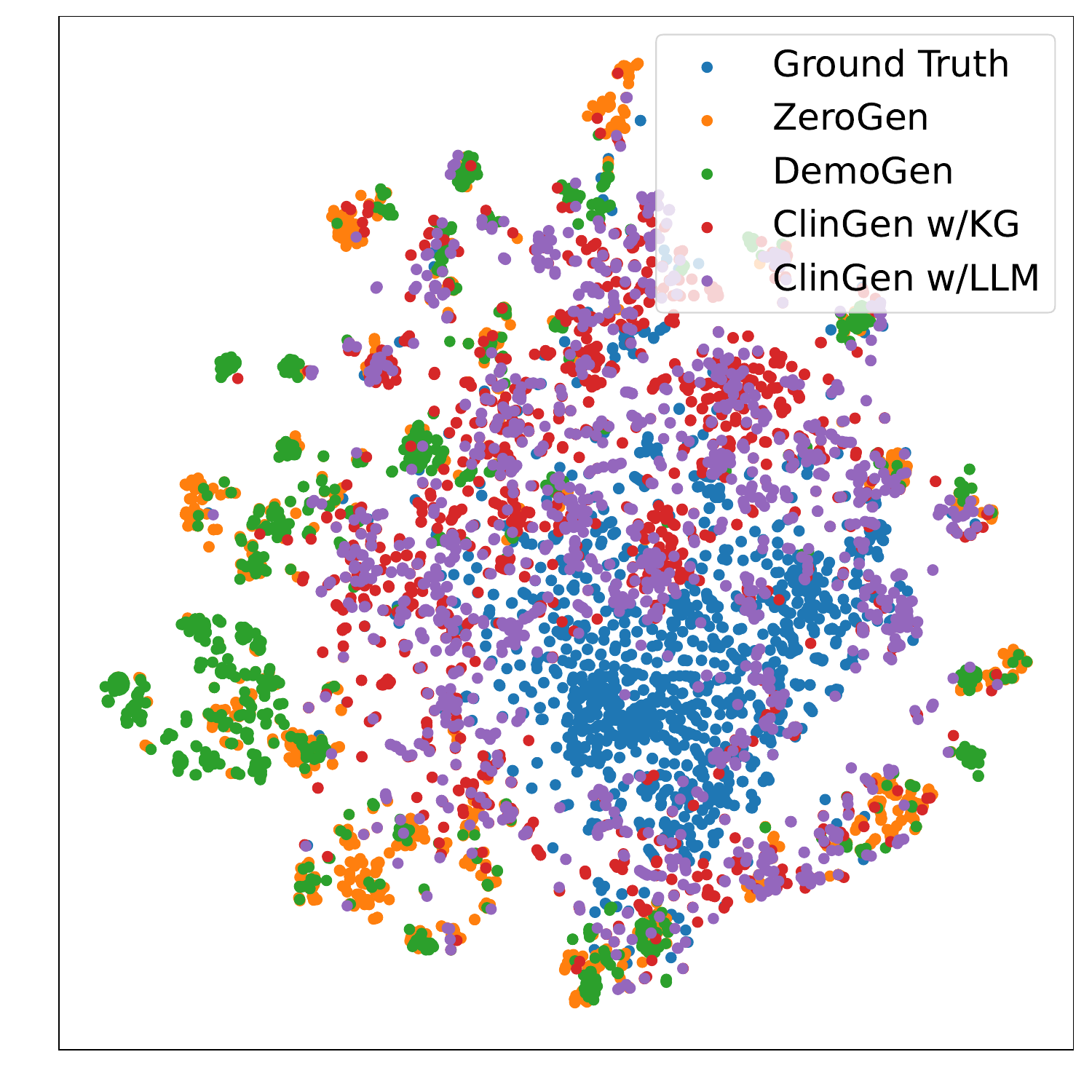}
		\label{fig:bc5cdr_disease_sentencebert_ours}
	} %\hfill
         \hspace{-0.2ex}
	% \subfigure[MEDIQA-RQE]{
	% 	\includegraphics[width=0.25\linewidth]{figures/mediqa_rqe_sentencebert_bsl.pdf}
	% 	\label{fig:mediqa_rqe_sentencebert_bsl}
	% }\hspace{-1.5ex}
     \subfigure[Case study of generated examples]{
		\includegraphics[width=0.65\linewidth]{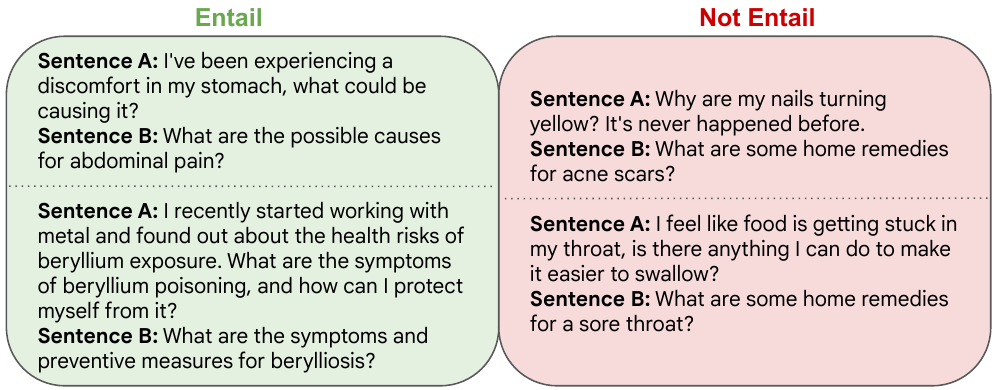}
		\label{fig:case_study_llm}
	}
	\caption{Data distribution and diversity measures on {\ours}. (a) is from BC5CDR-Disease and (b) is from MEDIQA-RQE using {\ours} with LLM. \vspace{-2ex}}
	% \vspace{-2ex}
\label{fig:quality_ana1}
\end{figure*}

 \begin{figure*}[tp]
	\centering
	\vspace{-1ex}
	\subfigure[CMD]{
		\includegraphics[width=0.362\linewidth]{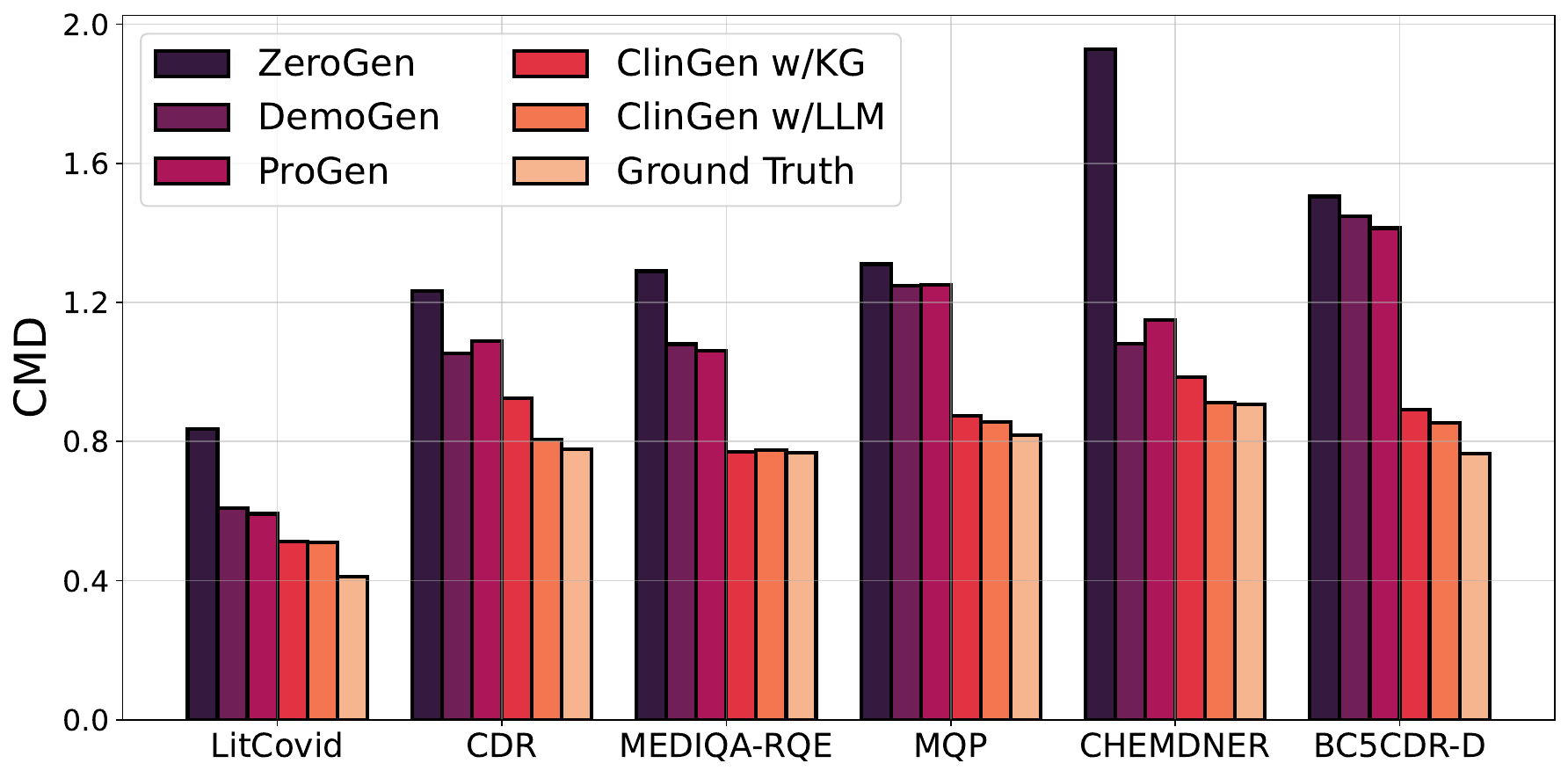}
		\label{fig:cmd-all}
	} %\hfill
         \hspace{-1.5ex}
	% \subfigure[MEDIQA-RQE]{
	% 	\includegraphics[width=0.25\linewidth]{figures/mediqa_rqe_sentencebert_bsl.pdf}
	% 	\label{fig:mediqa_rqe_sentencebert_bsl}
	% }\hspace{-1.5ex}
     \subfigure[Entity Coverage]{
		\includegraphics[width=0.362\linewidth]{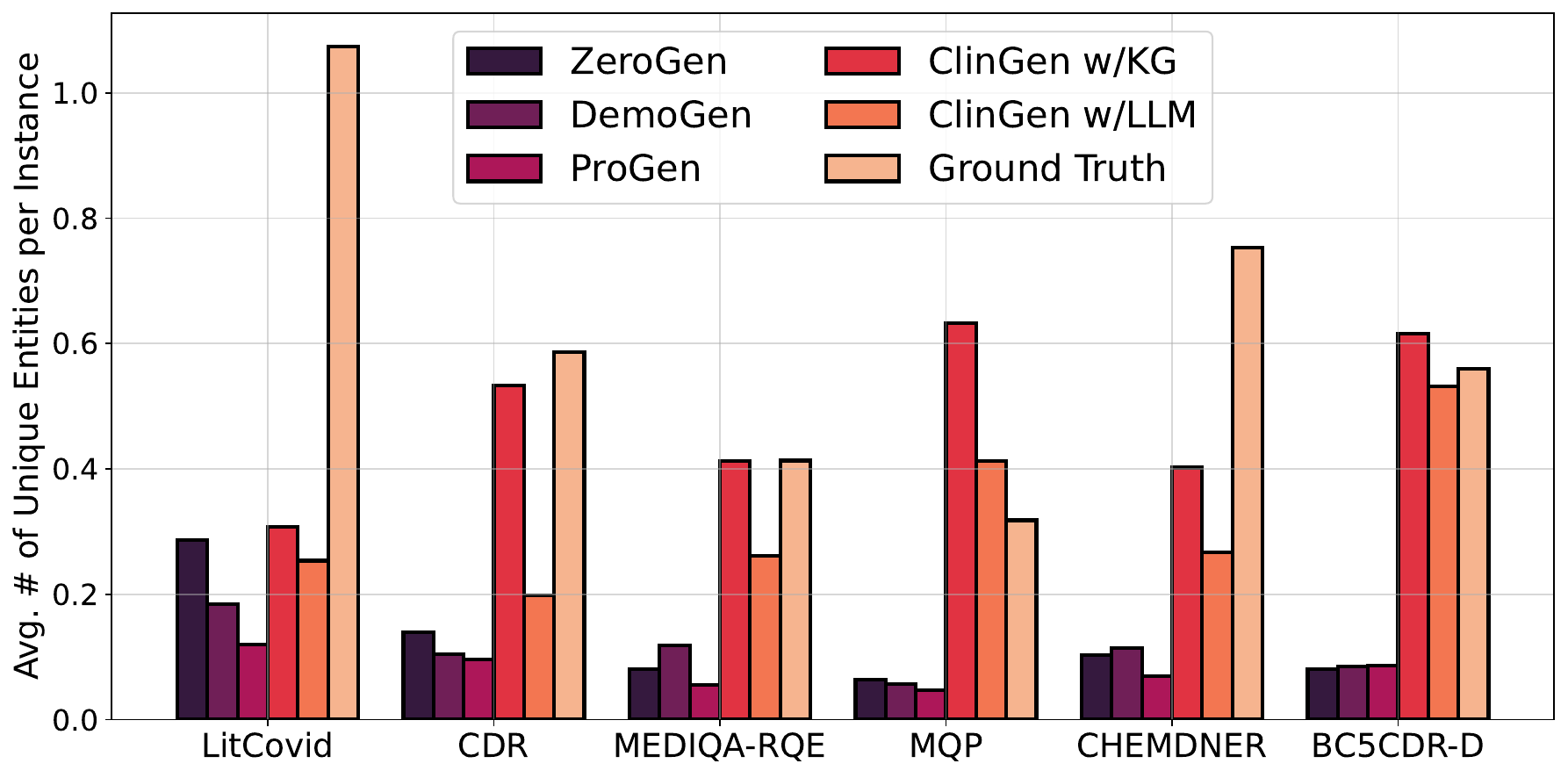}
		\label{fig:avg-entity-all}
	}
 \hspace{-1.5ex}
      \subfigure[Entity Frequency]{
		\includegraphics[width=0.248\linewidth]{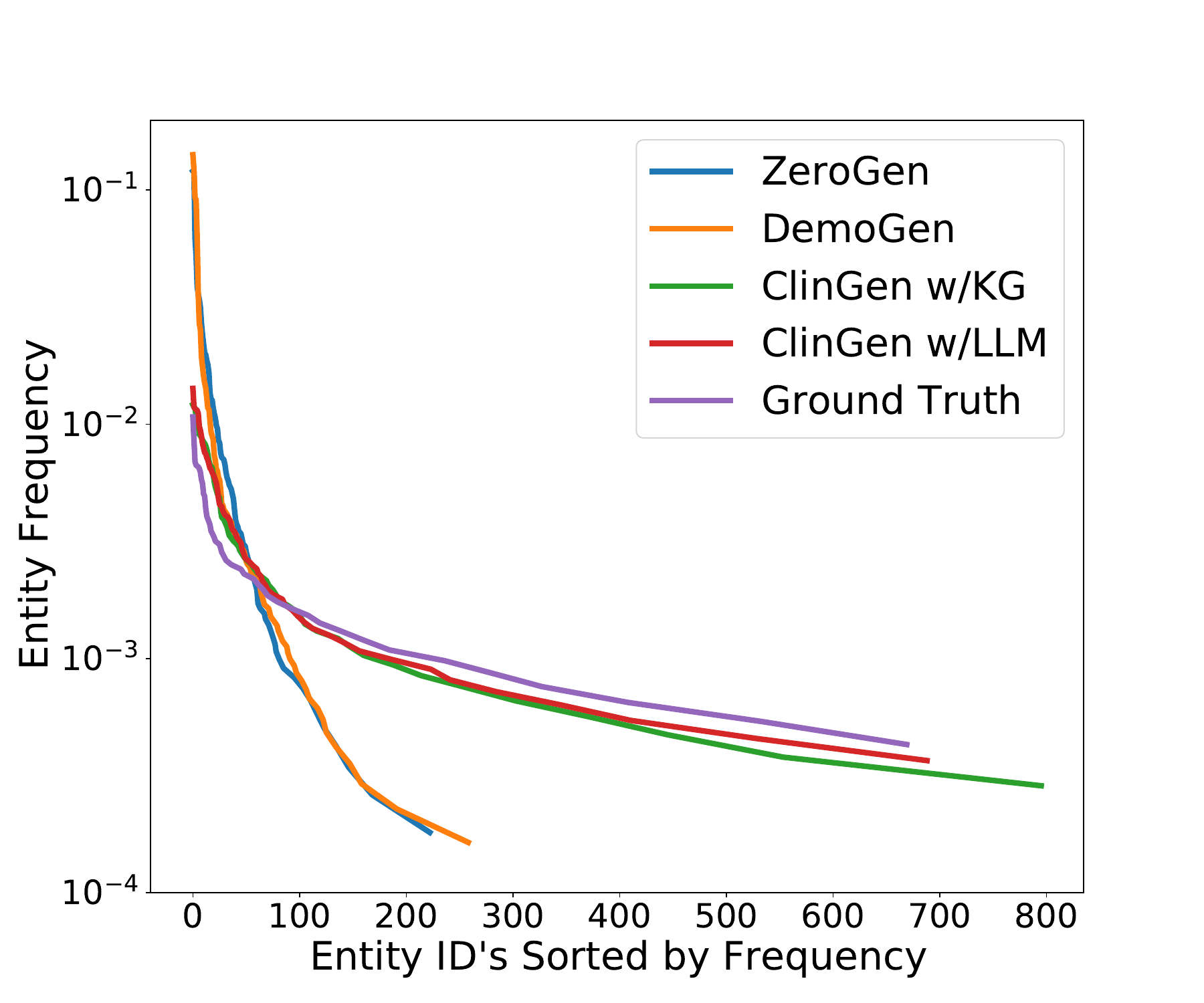}
		\label{fig:bc5cdr_disease_freq}
	}
	\caption{Data distribution and diversity measures on {\ours}. (c) is from BC5CDR-Disease.\vspace{-2ex}}
	\vspace{-1.5ex}
\label{fig:quality_ana2}
\end{figure*}

\noindent  \textbf{Comparison with few-shot inference via prompting LLM.}
% How Does the Value of Synthetic Data Compare to Clean Data Points? 
We also evaluate the performance of 5-shot in-context learning with ChatGPT \blue{and 3 medical LLMs, namely PMC-LLaMa-13b~\citep{wu2023pmcllama}, MedAlpaca-13b~\citep{han2023medalpaca}}. Due to budget limits, we  run experiments on datasets with few testing samples for each task. 
As presented in Table~\ref{tab:gpt_inference}, {\ours} at PubMedBERT$_{\texttt{Large}}$ scale achieves better results on 5 out of 6 datasets than ChatGPT few-shot learning, which uses $\sim 530 \times$ more parameters. 
One exception is for PUBHEALTH, as it requires complex reasoning abilities that PubMedBERT$_{\texttt{Large}}$ may not fully possess. 
Three medical LLMs, on the other hand, perform less effectively than both {\ours} and GPT-3.5 due to fewer parameters, limited reasoning capabilities, and training on a general medical corpus unsuited for the tasks.
Overall, {\ours} offers cost-effective and time-efficient advantages. 
While it entails a one-time investment in both money and time for synthetic training data generation, subsequent prediction relying on a moderate-sized model is much more efficient. 
Besides, the continued use of ChatGPT for inference on new testing data 
incurs ongoing time and financial costs, while our model requires zero additional costs for new data. 
% The price information is exhibited in Appendix \ref{sec:apd_cost}.

\noindent \textbf{Effect of Topic Extraction and Style Suggestion.}
We inspect different components of {\ours} in Table~\ref{tab:ablation}. It is observed that both Topics Extraction and Style Suggestion contribute to model performance as they enhance the relevance of generated samples to domain knowledge and introduce greater diversity. Different from the other datasets, MEDIQA-RQE shows more performance gain incorporating writing style than topics. It is because NLI tasks focus on capturing the relationships between two sentences while incorporating additional knowledge entities does not directly help the model improve the reasoning ability.

% \joyce{Style seemingly plays a key role in MEDIQA-RQE. Thoughts as to why?}

% \vspace{-0.7ex}
\section{Quality Analysis of the Synthetic Data}
% \vspace{-0.4ex}
\label{sec:quality_analysis}
\textbf{Data Distribution Measures.}
% In this section, we present the data distribution and diversity measurement of the synthetic dataset generated by {\ours}.
Figure~\ref{fig:bc5cdr_disease_sentencebert_ours} shows the t-SNE plot of data generated by {\ours} and baselines compared with the ground truth. This visualization demonstrates that {\ours} exhibits a greater overlap with the ground truth, indicating a similar distribution as the original dataset.
In addition, as depicted in Figure \ref{fig:cmd-all}, the embedding of \textit{\ours} aligns more closely with the ground truth distribution than other baselines across all six datasets, further justifying the efficacy of {\ours} for mitigating the distribution shift issue.
% Notably, \textit{\ours} w/ LLM tends to replicate the ground truth more precisely than \textit{\ours} w/ KG. 

\begin{table}[t]
% \floatconts
% \vspace{-1ex}
  \caption{Average Pairwise Similarity. \vspace{-3ex}}
  \resizebox{0.92\linewidth}{!}{
  \begin{tabular}{lcccc}
  \toprule
  % & \multicolumn{2}{c}{\textbf{HOC}} & \multicolumn{2}{c}{\textbf{CDR}} & \multicolumn{2}{c}{\textbf{MEDIQA-RQE}} & \multicolumn{2}{c}{\textbf{NCBI-Disease}}\\
  & \bfseries HOC & \bfseries CDR & \bfseries MEDIQA-RQE & \bfseries NCBI-Disease \\
  % \midrule
  % \cmidrule(lr){2-3} \cmidrule(lr){4-5} \cmidrule(lr){6-7} \cmidrule(lr){8-9}
  % & Avg.Sim. & Avg.Sim. & Avg.Sim. & Avg.Sim. \\
  \midrule
  ZeroGen & 0.512 & 0.469  &   0.277 &  0.528  \\
  DemoGen & 0.463 &    0.377   & 0.289  &  0.281  \\
  ProGen  & 0.481 &   0.321   & 0.290  &  0.357  \\
  \rowcolor{teal!10} {\ours} w/ KG  & 0.440 & \textbf{0.291}   & \textbf{0.243} &  0.180    \\
  \rowcolor{teal!10} {\ours} w/ LLM & \textbf{0.432}  & 0.338  &  0.255 &  \textbf{0.155} \\
  % \midrule
  Ground truth   & 0.265  & 0.268  &  0.164 &  0.262  \\
  \bottomrule
  \end{tabular}
  }
  \label{table:aps}
  \vspace{-1ex}
\end{table}

\noindent \textbf{Diversity Measures.}
\label{sec:diversity_measures}
Table~\ref{table:aps} calculates the average cosine similarity for sample pairs using SentenceBERT embeddings.
Compared to baselines, the dataset generated with {\ours} exhibits lower cosine similarity and the average similarity is close to that of the ground truth training data, which shows {\ours} could render more diverse data.  
% Moreover, while {\ours} w/ KG envelops more entities, {\ours} w/ LLM delivers a data distribution that is most congruent with the ground truth, further verifying the observations from Figure \ref{fig:cmd-all} and \ref{fig:avg-entity-all}.
% Moreover, Figure \ref{fig:avg-entity-all} highlights that \ours covers a broader range of entities in comparison to baselines. We find that {\ours} w/ KG captures a larger variety of entities than {\ours} w/ LLM, because KG tends to cover more extensive knowledge, including relatively uncommon information that may not be present in LLMs.
% attributed to the rich information, especially rarer knowledge, contained in the knowledge graph.
% Figure \ref{fig:bc5cdr_disease_freq} reflects that the entity frequency distribution of {\ours} is more in line with the ground truth, having a relatively balanced distribution among all entities. This ensures that {\ours} generates synthetic data with a wide range of diverse topics.

Moreover, Figure \ref{fig:avg-entity-all} highlights {\ours} covers a broader range of entities than baselines, with {\ours} w/ KG capturing more entities due to KGs' extensive knowledge. Figure \ref{fig:bc5cdr_disease_freq} reflects {\ours} has a more balanced entity frequency distribution aligned with ground truth, ensuring diverse topic coverage.

\begin{table*}[t]
% \floatconts
  \caption{The average cost (in US dollars) of running {\ours} on various datasets per 1000 samples, compared with prompting GPT-3.5 for inference and DemoGen.}
  \resizebox{0.8\linewidth}{!}{
  \begin{tabular}{lccccccc}
  \toprule
  & \bfseries HOC & \bfseries GAD & \bfseries ChemProt & \bfseries MEDIQA-RQE & \bfseries PUBHEALTH & \bfseries NCBI-Disease & \bfseries CASI\\
  % \midrule
  \midrule
  GPT-3.5 Inference & 1.09 & 1.05 & 5.75 & 2.15 & 2.80 & 0.90 & 1.30 \\ 
  % \hline
  DemoGen & 0.59 & 0.66 & 1.35 & 0.81 & 0.92 & 1.12 & 1.28 \\
  \rowcolor{teal!10} {\ours} w/ KG & 0.65 & 0.73 & 1.47 & 0.86 & 1.01 & 1.41 & 1.55 \\
  \rowcolor{teal!10} {\ours} w/ LLM & 0.72 & 0.84 & 1.51 & 0.90 & 1.34 & 1.49 & 1.62 \\
  \bottomrule
  \end{tabular}
  }
  \label{tab:money_cost}
  \vspace{-1ex}
\end{table*}

% rare diseases

\noindent \textbf{Case Study.}
In Figure~\ref{fig:case_study_llm}, we present a case study of examples generated by {\ours} with LLM on MEDIQA-RQE dataset, which consists of consumer health queries. The examples reveal that the sentences generated by {\ours} include more extensive contextual information compared with the baseline. These sentences closely resemble the queries people might pose in real-life scenarios.

\noindent \textbf{Study on Factual Consistency.} 
A human evaluation was carried out to assess the factual accuracy of the generated outputs across six representative tasks: LitCovid, CDR, Mediqa-RQE, MQP, PubHealth, and BC5CDR. For each task, a sample of 100 examples per class was randomly selected. Medical students then examine the generated text and evaluate its factuality. The findings from this rigorous human study revealed no instances of misinformation or hallucinated content in the randomly sampled examples, verifying the system's reliability in generating factually sound outputs.

% Many of the tasks do not require deep factual knowledge. For example, Mediqa-RQE involves generating short medical queries, while BC5CDR requires generating sentences mentioning diseases, neither of which necessitates factual claims.
% The language model used (GPT-3.5) has been specifically aligned to improve factuality and reduce hallucinations [1]. Prior work [2] has shown it can provide reliable medical information without significant misinformation.
% In summary, our study did not reveal factual consistency issues, both due to the nature of the tasks and the strong factual abilities of the LLM employed. We will conduct a more comprehensive evaluation for future work.

\noindent \textbf{Monetary Cost}
\label{sec:apd_cost}
We display the monetary cost of {\ours} for calling the OpenAI APIs, with a comparison with prompting GPT-3.5 for direct inference and DemoGen. From the values shown in Table~\ref{tab:money_cost}, we observe that inference via GPT-3.5 generally has a higher cost, as it needs to input all the testing samples for prompting. In contrast, DemoGen has a relatively lower cost, because it does not include the topics and writing styles to the prompts as {\ours} does.

% \vspace{-0.8ex}
\section{Conclusion}
% \vspace{-0.6ex}
% Large language models (LLMs) inherently capture significant clinical knowledge. 
% In this work, we focus on an effective and generic approach for clinical text data generation with LLMs. 
% We thoroughly examine existing synthetic data generation methods for clinical tasks and identify significant issues, such as large distribution shifts and limited diversity in the generated data. 
% To address these challenges, we proposed a novel clinical knowledge-infuse framework, {\ours}, where clinical knowledge from both non-parametric KGs and parametric LLMs are leveraged to contextualize the clinical data generation. Specifically, clinical topic knowledge and real-world writing styles are elicited to serve as the foundation to create domain-specific prompts. 
% Through extensive empirical evaluations across 7 clinical NLP tasks spanning 16 datasets and comparing against 9 baseline methods from various categories, our results consistently demonstrate that {\ours}-generated data improves task performance, closely aligns with real data distributions, and significantly enhances data diversity compared to existing approaches.
% We anticipate this knowledge-infused clinical data generation paradigm be readily adapted for any future clinical text tasks, serving as a versatile approach for clinical NLP studies.  

In this work, we study clinical text data generation using LLMs. We thoroughly assess existing methods for clinical data generation and identify issues including distribution shifts and limited diversity. 
To tackle these challenges, we introduce {\ours}, a  framework that leverages clinical knowledge from non-parametric KGs and parametric LLMs. 
This empowers data generation by utilizing clinical topic knowledge and real-world writing styles in domain-specific prompts. 
Our extensive empirical evaluations across 8 clinical NLP tasks and 18 datasets, compared to 10 baseline methods, consistently show that {\ours} improves task performance, aligns closely with real data, and enhances data diversity. 
We expect {\ours} can be seamlessly incorporated into a broad suite of clinical text tasks to advance clinical NLP research.
% \clearpage
\section*{Acknowledgement}
We thank the anonymous reviewers and area chairs for valuable feedbacks. 
This research was partially supported by the Emory Global Diabetes Center of the Woodruff Sciences Center, Emory University. Research reported in this publication was supported by the National Institute Of Diabetes And Digestive And Kidney Diseases of the National Institutes of Health under Award Number K25DK135913. 
The research also receives partial support by the National Science Foundation under Award Number IIS-2145411. The content is solely the responsibility of the authors and does not necessarily represent the official views of the National Institutes of Health.
We also thank Microsoft for providing research credits under the Accelerating Foundation Models Research Program.

\section*{Limitation}
In this work, we propose {\ours} to better harness the LLM for synthetic text data generation. 
Despite its strong performance, we mainly verify their efficacy from their empirical performance, sample diversity, and distribution gaps. There are still some limitations to this work:

\noindent \textbf{Factuality of LLM-generated Text}. One issue with LLM-based synthetic data generation is the phenomenon of \emph{hallucination}, wherein the model generates information that does not ground in reality~\citep{zhang2023siren}. This can lead to the propagation of misinformation, which may have negative impacts on the clinical domain. However, we have conducted a human study to justify that \emph{our generated synthetic data does not suffer from the issue of misinformation}.

\noindent \textbf{Application to other type of clinical data}.
Apart from text, there are other types of clinical data: 
For example, EHR data falls within a distinct modality (i.e. tabular data) from textual data, which may require different methodologies and approaches~\citep{wornow2023shaky}. 
% Nonetheless, we are aware of the capabilities of LLMs in this context. Recent studies~\citep{hegselmann2023tabllm,borisov2022language} have explored transforming tabular data into text to harness the power of LLMs, which yields promising results and shows the potential of LLMs for structured data generation. 

\section*{Ethics Consideration}
On specific issue is about patient privacy. To eliminate this concern, we carefully select the five few-shot demonstrations to ensure they are fully free from any Protected Health Information (PHI) related to patients.  We also make a deliberate effort to \emph{avoid any instructions} that can potentially extract sensitive patient information within the prompts. 
In addition, we have opted out of human review for the data by completing the Azure OpenAI Additional Use Case Form\footnote{\url{https://aka.ms/oai/additionalusecase}}. This allows us to use the Azure OpenAI service while ensuring Microsoft does not have access to patient data.

\bibliography{custom}

\begin{thebibliography}{74}
\expandafter\ifx\csname natexlab\endcsname\relax\def\natexlab#1{#1}\fi

\bibitem[{Abacha and Demner-Fushman(2016)}]{abacha2016recognizing}
Asma~Ben Abacha and Dina Demner-Fushman. 2016.
\newblock Recognizing question entailment for medical question answering.
\newblock In \emph{AMIA Annual Symposium Proceedings}, volume 2016, page 310.

\bibitem[{Agrawal et~al.(2022)Agrawal, Hegselmann, Lang, Kim, and Sontag}]{agrawal2022large}
Monica Agrawal, Stefan Hegselmann, Hunter Lang, Yoon Kim, and David Sontag. 2022.
\newblock \href {https://aclanthology.org/2022.emnlp-main.130} {Large language models are few-shot clinical information extractors}.
\newblock In \emph{Proceedings of the 2022 Conference on Empirical Methods in Natural Language Processing}, pages 1998--2022, Abu Dhabi, United Arab Emirates. Association for Computational Linguistics.

\bibitem[{Baker et~al.(2015)Baker, Silins, Guo, Ali, H{\"o}gberg, Stenius, and Korhonen}]{hoc}
Simon Baker, Ilona Silins, Yufan Guo, Imran Ali, Johan H{\"o}gberg, Ulla Stenius, and Anna Korhonen. 2015.
\newblock Automatic semantic classification of scientific literature according to the hallmarks of cancer.
\newblock \emph{Bioinformatics}, 32(3):432--440.

\bibitem[{Ben~Abacha et~al.(2019)Ben~Abacha, Shivade, and Demner-Fushman}]{mediqa-nli}
Asma Ben~Abacha, Chaitanya Shivade, and Dina Demner-Fushman. 2019.
\newblock \href {https://doi.org/10.18653/v1/W19-5039} {Overview of the {MEDIQA} 2019 shared task on textual inference, question entailment and question answering}.
\newblock In \emph{Proceedings of the 18th BioNLP Workshop and Shared Task}, pages 370--379, Florence, Italy. Association for Computational Linguistics.

\bibitem[{Bravo et~al.(2015)Bravo, Pi{\~{n}}ero, Queralt-Rosinach, Rautschka, and Furlong}]{gad}
{\`{A}}lex Bravo, Janet Pi{\~{n}}ero, N{\'{u}}ria Queralt-Rosinach, Michael Rautschka, and Laura~I Furlong. 2015.
\newblock Extraction of relations between genes and diseases from text and large-scale data analysis: implications for translational research.
\newblock \emph{{BMC} Bioinformatics}, 16(1).

\bibitem[{Brown et~al.(2020)Brown, Mann, Ryder, Subbiah, Kaplan, Dhariwal, Neelakantan, Shyam, Sastry, Askell et~al.}]{brown2020language}
Tom~B. Brown, Benjamin Mann, Nick Ryder, Melanie Subbiah, Jared Kaplan, Prafulla Dhariwal, Arvind Neelakantan, Pranav Shyam, Girish Sastry, Amanda Askell, et~al. 2020.
\newblock \href {https://proceedings.neurips.cc/paper/2020/hash/1457c0d6bfcb4967418bfb8ac142f64a-Abstract.html} {Language models are few-shot learners}.
\newblock In \emph{Advances in Neural Information Processing Systems 33: Annual Conference on Neural Information Processing Systems 2020}.

\bibitem[{Chen et~al.(2020)Chen, Yang, and Yang}]{chen2020mixtext}
Jiaao Chen, Zichao Yang, and Diyi Yang. 2020.
\newblock \href {https://doi.org/10.18653/v1/2020.acl-main.194} {{M}ix{T}ext: Linguistically-informed interpolation of hidden space for semi-supervised text classification}.
\newblock In \emph{Proceedings of the 58th Annual Meeting of the Association for Computational Linguistics}, pages 2147--2157, Online. Association for Computational Linguistics.

\bibitem[{Chen et~al.(2023)Chen, Wang, Lin, Zhao, and Yang}]{chen2023few}
Peng Chen, Jian Wang, Hongfei Lin, Di~Zhao, and Zhihao Yang. 2023.
\newblock Few-shot biomedical named entity recognition via knowledge-guided instance generation and prompt contrastive learning.
\newblock \emph{Bioinformatics}, 39(8):btad496.

\bibitem[{Chen et~al.(2021)Chen, Allot, Leaman, Do{\u{g}}an, and Lu}]{litcovid}
Qingyu Chen, Alexis Allot, Robert Leaman, Rezarta~Islamaj Do{\u{g}}an, and Zhiyong Lu. 2021.
\newblock Overview of the biocreative vii litcovid track: multi-label topic classification for covid-19 literature annotation.
\newblock In \emph{Proceedings of the BioCreative challenge evaluation workshop}.

\bibitem[{Chen et~al.(2022{\natexlab{a}})Chen, Li, Deng, Tan, Xu, Huang, Si, Chen, and Zhang}]{lightner}
Xiang Chen, Lei Li, Shumin Deng, Chuanqi Tan, Changliang Xu, Fei Huang, Luo Si, Huajun Chen, and Ningyu Zhang. 2022{\natexlab{a}}.
\newblock \href {https://aclanthology.org/2022.coling-1.209} {{L}ight{NER}: A lightweight tuning paradigm for low-resource {NER} via pluggable prompting}.
\newblock In \emph{Proceedings of the 29th International Conference on Computational Linguistics}, pages 2374--2387, Gyeongju, Republic of Korea. International Committee on Computational Linguistics.

\bibitem[{Chen et~al.(2022{\natexlab{b}})Chen, Zhang, Xie, Deng, Yao, Tan, Huang, Si, and Chen}]{chen2022knowprompt}
Xiang Chen, Ningyu Zhang, Xin Xie, Shumin Deng, Yunzhi Yao, Chuanqi Tan, Fei Huang, Luo Si, and Huajun Chen. 2022{\natexlab{b}}.
\newblock Knowprompt: Knowledge-aware prompt-tuning with synergistic optimization for relation extraction.
\newblock In \emph{Proceedings of the ACM Web conference 2022}, pages 2778--2788.

\bibitem[{Chintagunta et~al.(2021)Chintagunta, Katariya, Amatriain, and Kannan}]{chintagunta2021medically}
Bharath Chintagunta, Namit Katariya, Xavier Amatriain, and Anitha Kannan. 2021.
\newblock \href {https://doi.org/10.18653/v1/2021.nlpmc-1.9} {Medically aware {GPT}-3 as a data generator for medical dialogue summarization}.
\newblock In \emph{Proceedings of the Second Workshop on Natural Language Processing for Medical Conversations}, pages 66--76, Online. Association for Computational Linguistics.

\bibitem[{Chung et~al.(2022)Chung, Hou, Longpre, Zoph, Tay, Fedus, Li, Wang, Dehghani, Brahma et~al.}]{chung2022scaling}
Hyung~Won Chung, Le~Hou, Shayne Longpre, Barret Zoph, Yi~Tay, William Fedus, Eric Li, Xuezhi Wang, Mostafa Dehghani, Siddhartha Brahma, et~al. 2022.
\newblock \href {https://arxiv.org/abs/2210.11416} {Scaling instruction-finetuned language models}.
\newblock \emph{ArXiv preprint}, abs/2210.11416.

\bibitem[{Chung et~al.(2023)Chung, Kamar, and Amershi}]{chung-etal-2023-increasing}
John Chung, Ece Kamar, and Saleema Amershi. 2023.
\newblock \href {https://doi.org/10.18653/v1/2023.acl-long.34} {Increasing diversity while maintaining accuracy: Text data generation with large language models and human interventions}.
\newblock In \emph{Proceedings of the 61st Annual Meeting of the Association for Computational Linguistics (Volume 1: Long Papers)}, pages 575--593, Toronto, Canada. Association for Computational Linguistics.

\bibitem[{Cui et~al.(2023)Cui, Lu, Wang, Xu, Ma, Yu, Yu, Kan, Fu, Ling, Ho, Wang, and Yang}]{cui2023a}
Hejie Cui, Jiaying Lu, Shiyu Wang, Ran Xu, Wenjing Ma, Shaojun Yu, Yue Yu, Xuan Kan, Tianfan Fu, Chen Ling, Joyce Ho, Fei Wang, and Carl Yang. 2023.
\newblock \href {https://openreview.net/forum?id=CZCktJoBRh} {A survey on knowledge graphs for healthcare: Resources, application progress, and promise}.
\newblock In \emph{ICML 3rd Workshop on Interpretable Machine Learning in Healthcare (IMLH)}.

\bibitem[{Dogan et~al.(2014)Dogan, Leaman, and Lu}]{ncbi-disease}
Rezarta~Islamaj Dogan, Robert Leaman, and Zhiyong Lu. 2014.
\newblock Ncbi disease corpus: A resource for disease name recognition and concept normalization.
\newblock \emph{Journal of biomedical informatics}, 47:1--10.

\bibitem[{Fries et~al.(2022)Fries, Weber, Seelam, Altay, Datta, Garda, Kang, Su, Kusa, Cahyawijaya, Barth, Ott et~al.}]{fries2022bigbio}
Jason~Alan Fries, Leon Weber, Natasha Seelam, Gabriel Altay, Debajyoti Datta, Samuele Garda, Myungsun Kang, Ruisi Su, Wojciech Kusa, Samuel Cahyawijaya, Fabio Barth, Simon Ott, et~al. 2022.
\newblock Bigbio: A framework for data-centric biomedical natural language processing.
\newblock In \emph{Thirty-sixth Conference on Neural Information Processing Systems Datasets and Benchmarks Track}.

\bibitem[{Giorgi et~al.(2023)Giorgi, Toma, Xie, Chen, An, Zheng, and Wang}]{giorgi2023clinical}
John Giorgi, Augustin Toma, Ronald Xie, Sondra Chen, Kevin~R An, Grace~X Zheng, and Bo~Wang. 2023.
\newblock \href {https://arxiv.org/abs/2305.02220} {Clinical note generation from doctor-patient conversations using large language models: Insights from mediqa-chat}.
\newblock \emph{ArXiv preprint}, abs/2305.02220.

\bibitem[{Gu et~al.(2021)Gu, Tinn, Cheng, Lucas, Usuyama, Liu, Naumann, Gao, and Poon}]{gu2021domain}
Yu~Gu, Robert Tinn, Hao Cheng, Michael Lucas, Naoto Usuyama, Xiaodong Liu, Tristan Naumann, Jianfeng Gao, and Hoifung Poon. 2021.
\newblock Domain-specific language model pretraining for biomedical natural language processing.
\newblock \emph{ACM Transactions on Computing for Healthcare (HEALTH)}, 3(1):1--23.

\bibitem[{Han et~al.(2023)Han, Adams, Papaioannou, Grundmann, Oberhauser, L{\"o}ser, Truhn, and Bressem}]{han2023medalpaca}
Tianyu Han, Lisa~C Adams, Jens-Michalis Papaioannou, Paul Grundmann, Tom Oberhauser, Alexander L{\"o}ser, Daniel Truhn, and Keno~K Bressem. 2023.
\newblock \href {https://arxiv.org/abs/2304.08247} {Medalpaca--an open-source collection of medical conversational ai models and training data}.
\newblock \emph{ArXiv preprint}, abs/2304.08247.

\bibitem[{Ive et~al.(2020)Ive, Viani, Kam, Yin, Verma, Puntis, Cardinal, Roberts, Stewart, and Velupillai}]{ive2020generation}
Julia Ive, Natalia Viani, Joyce Kam, Lucia Yin, Somain Verma, Stephen Puntis, Rudolf~N Cardinal, Angus Roberts, Robert Stewart, and Sumithra Velupillai. 2020.
\newblock Generation and evaluation of artificial mental health records for natural language processing.
\newblock \emph{NPJ digital medicine}, 3(1):69.

\bibitem[{Jin et~al.(2019)Jin, Dhingra, Liu, Cohen, and Lu}]{jin2019pubmedqa}
Qiao Jin, Bhuwan Dhingra, Zhengping Liu, William Cohen, and Xinghua Lu. 2019.
\newblock \href {https://doi.org/10.18653/v1/D19-1259} {{P}ub{M}ed{QA}: A dataset for biomedical research question answering}.
\newblock In \emph{Proceedings of the 2019 Conference on Empirical Methods in Natural Language Processing and the 9th International Joint Conference on Natural Language Processing (EMNLP-IJCNLP)}, pages 2567--2577, Hong Kong, China. Association for Computational Linguistics.

\bibitem[{Kang et~al.(2021)Kang, Perotte, Tang, Ta, and Weng}]{kang2021umls}
Tian Kang, Adler Perotte, Youlan Tang, Casey Ta, and Chunhua Weng. 2021.
\newblock Umls-based data augmentation for natural language processing of clinical research literature.
\newblock \emph{Journal of the American Medical Informatics Association}, 28(4):812--823.

\bibitem[{Khot et~al.(2018)Khot, Sabharwal, and Clark}]{khot2018scitail}
Tushar Khot, Ashish Sabharwal, and Peter Clark. 2018.
\newblock \href {https://www.aaai.org/ocs/index.php/AAAI/AAAI18/paper/view/17368} {Scitail: {A} textual entailment dataset from science question answering}.
\newblock In \emph{Proceedings of the Thirty-Second {AAAI} Conference on Artificial Intelligence, (AAAI-18)}, pages 5189--5197. {AAAI} Press.

\bibitem[{Kotonya and Toni(2020)}]{PUBHEALTH}
Neema Kotonya and Francesca Toni. 2020.
\newblock \href {https://doi.org/10.18653/v1/2020.emnlp-main.623} {Explainable automated fact-checking for public health claims}.
\newblock In \emph{Proceedings of the 2020 Conference on Empirical Methods in Natural Language Processing (EMNLP)}, pages 7740--7754, Online. Association for Computational Linguistics.

\bibitem[{Krallinger et~al.(2015)Krallinger, Rabal, Leitner, Vazquez, Salgado, Lu, Leaman, Lu, Ji, Lowe, Sayle, Batista-Navarro et~al.}]{chemdner}
Martin Krallinger, Obdulia Rabal, Florian Leitner, Miguel Vazquez, David Salgado, Zhiyong Lu, Robert Leaman, Yanan Lu, Donghong Ji, Daniel~M. Lowe, Roger~A. Sayle, Riza Batista-Navarro, et~al. 2015.
\newblock The chemdner corpus of chemicals and drugs and its annotation principles.
\newblock \emph{Journal of Cheminformatics}, 7(1):S2.

\bibitem[{Kumar et~al.(2020)Kumar, Choudhary, and Cho}]{kumar2020data}
Varun Kumar, Ashutosh Choudhary, and Eunah Cho. 2020.
\newblock \href {https://aclanthology.org/2020.lifelongnlp-1.3} {Data augmentation using pre-trained transformer models}.
\newblock In \emph{Proceedings of the 2nd Workshop on Life-long Learning for Spoken Language Systems}, pages 18--26, Suzhou, China. Association for Computational Linguistics.

\bibitem[{Lee et~al.(2023)Lee, Goldberg, and Kohane}]{lee2023ai}
Peter Lee, Carey Goldberg, and Isaac Kohane. 2023.
\newblock \emph{The AI Revolution in Medicine: GPT-4 and Beyond}.
\newblock Pearson Education, Limited.

\bibitem[{Li et~al.(2016)Li, Sun, Johnson, Sciaky, Wei, Leaman, Davis, Mattingly, Wiegers, and Lu}]{bc5cdr}
Jiao Li, Yueping Sun, Robin~J. Johnson, Daniela Sciaky, Chih{-}Hsuan Wei, Robert Leaman, Allan~Peter Davis, Carolyn~J. Mattingly, Thomas~C. Wiegers, and Zhiyong Lu. 2016.
\newblock Biocreative {V} {CDR} task corpus: a resource for chemical disease relation extraction.
\newblock \emph{Database J. Biol. Databases Curation}, 2016.

\bibitem[{Li et~al.(2022)Li, Huang, and Zitnik}]{li2022graph}
Michelle~M Li, Kexin Huang, and Marinka Zitnik. 2022.
\newblock Graph representation learning in biomedicine and healthcare.
\newblock \emph{Nature Biomedical Engineering}, 6(12):1353--1369.

\bibitem[{Li et~al.(2023)Li, Zhu, Lu, and Yin}]{li-etal-2023-synthetic}
Zhuoyan Li, Hangxiao Zhu, Zhuoran Lu, and Ming Yin. 2023.
\newblock \href {https://doi.org/10.18653/v1/2023.emnlp-main.647} {Synthetic data generation with large language models for text classification: Potential and limitations}.
\newblock In \emph{Proceedings of the 2023 Conference on Empirical Methods in Natural Language Processing}, pages 10443--10461, Singapore. Association for Computational Linguistics.

\bibitem[{Liu et~al.(2022{\natexlab{a}})Liu, Swayamdipta, Smith, and Choi}]{liu2022wanli}
Alisa Liu, Swabha Swayamdipta, Noah~A. Smith, and Yejin Choi. 2022{\natexlab{a}}.
\newblock \href {https://aclanthology.org/2022.findings-emnlp.508} {{WANLI}: Worker and {AI} collaboration for natural language inference dataset creation}.
\newblock In \emph{Findings of the Association for Computational Linguistics: EMNLP 2022}, pages 6826--6847, Abu Dhabi, United Arab Emirates. Association for Computational Linguistics.

\bibitem[{Liu et~al.(2022{\natexlab{b}})Liu, Liu, Lu, Welleck, West, Le~Bras, Choi, and Hajishirzi}]{liu-etal-2022-generated}
Jiacheng Liu, Alisa Liu, Ximing Lu, Sean Welleck, Peter West, Ronan Le~Bras, Yejin Choi, and Hannaneh Hajishirzi. 2022{\natexlab{b}}.
\newblock \href {https://doi.org/10.18653/v1/2022.acl-long.225} {Generated knowledge prompting for commonsense reasoning}.
\newblock In \emph{Proceedings of the 60th Annual Meeting of the Association for Computational Linguistics (Volume 1: Long Papers)}, pages 3154--3169, Dublin, Ireland. Association for Computational Linguistics.

\bibitem[{Liu et~al.(2023)Liu, Wang, and Liu}]{liu2023utility}
Jialin Liu, Changyu Wang, and Siru Liu. 2023.
\newblock Utility of chatgpt in clinical practice.
\newblock \emph{Journal of Medical Internet Research}, 25:e48568.

\bibitem[{Loshchilov and Hutter(2019)}]{loshchilov2017decoupled}
Ilya Loshchilov and Frank Hutter. 2019.
\newblock \href {https://openreview.net/forum?id=Bkg6RiCqY7} {Decoupled weight decay regularization}.
\newblock In \emph{7th International Conference on Learning Representations, {ICLR} 2019, New Orleans, LA, USA, May 6-9, 2019}.

\bibitem[{McCreery et~al.(2020)McCreery, Katariya, Kannan, Chablani, and Amatriain}]{mqp}
Clara~H. McCreery, Namit Katariya, Anitha Kannan, Manish Chablani, and Xavier Amatriain. 2020.
\newblock \href {https://dl.acm.org/doi/10.1145/3394486.3412861} {Effective transfer learning for identifying similar questions: Matching user questions to {COVID-19} faqs}.
\newblock In \emph{{KDD} '20: The 26th {ACM} {SIGKDD} Conference on Knowledge Discovery and Data Mining, Virtual Event, CA, USA, August 23-27, 2020}, pages 3458--3465. {ACM}.

\bibitem[{Meng et~al.(2022)Meng, Huang, Zhang, and Han}]{meng2022generating}
Yu~Meng, Jiaxin Huang, Yu~Zhang, and Jiawei Han. 2022.
\newblock Generating training data with language models: Towards zero-shot language understanding.
\newblock In \emph{Advances in Neural Information Processing Systems}.

\bibitem[{Meng et~al.(2023)Meng, Michalski, Huang, Zhang, Abdelzaher, and Han}]{meng2023tuning}
Yu~Meng, Martin Michalski, Jiaxin Huang, Yu~Zhang, Tarek Abdelzaher, and Jiawei Han. 2023.
\newblock Tuning language models as training data generators for augmentation-enhanced few-shot learning.
\newblock In \emph{International Conference on Machine Learning}, pages 24457--24477. PMLR.

\bibitem[{Mesk{\'o} and Topol(2023)}]{mesko2023imperative}
Bertalan Mesk{\'o} and Eric~J Topol. 2023.
\newblock The imperative for regulatory oversight of large language models (or generative ai) in healthcare.
\newblock \emph{NPJ Digital Medicine}, 6(1):120.

\bibitem[{Mishra et~al.(2022)Mishra, Khashabi, Baral, Choi, and Hajishirzi}]{mishra-etal-2022-reframing}
Swaroop Mishra, Daniel Khashabi, Chitta Baral, Yejin Choi, and Hannaneh Hajishirzi. 2022.
\newblock \href {https://doi.org/10.18653/v1/2022.findings-acl.50} {Reframing instructional prompts to {GPT}k{'}s language}.
\newblock In \emph{Findings of the Association for Computational Linguistics: ACL 2022}, pages 589--612, Dublin, Ireland. Association for Computational Linguistics.

\bibitem[{Moon et~al.(2014)Moon, Pakhomov, Liu, Ryan, and Melton}]{claim}
Sungrim Moon, Serguei Pakhomov, Nathan Liu, James~O Ryan, and Genevieve~B Melton. 2014.
\newblock A sense inventory for clinical abbreviations and acronyms created using clinical notes and medical dictionary resources.
\newblock \emph{Journal of the American Medical Informatics Association}, 21(2):299--307.

\bibitem[{OpenAI(2023{\natexlab{a}})}]{gpt4}
OpenAI. 2023{\natexlab{a}}.
\newblock Gpt-4 technical report.
\newblock \emph{arXiv}.

\bibitem[{OpenAI(2023{\natexlab{b}})}]{chatgpt}
OpenAI. 2023{\natexlab{b}}.
\newblock \href {https://openai.com/blog/chatgpt} {Introducing chatgpt}.

\bibitem[{Ouyang et~al.(2022)Ouyang, Wu, Jiang, Almeida, Wainwright, Mishkin, Zhang, Agarwal, Slama, Ray et~al.}]{ouyang2022training}
Long Ouyang, Jeffrey Wu, Xu~Jiang, Diogo Almeida, Carroll Wainwright, Pamela Mishkin, Chong Zhang, Sandhini Agarwal, Katarina Slama, Alex Ray, et~al. 2022.
\newblock Training language models to follow instructions with human feedback.
\newblock \emph{Advances in Neural Information Processing Systems}, 35:27730--27744.

\bibitem[{Paszke et~al.(2019)Paszke, Gross, Massa, Lerer, Bradbury, Chanan, Killeen, Lin et~al.}]{paszke2019pytorch}
Adam Paszke, Sam Gross, Francisco Massa, Adam Lerer, James Bradbury, Gregory Chanan, Trevor Killeen, Zeming Lin, et~al. 2019.
\newblock \href {https://proceedings.neurips.cc/paper/2019/hash/bdbca288fee7f92f2bfa9f7012727740-Abstract.html} {Pytorch: An imperative style, high-performance deep learning library}.
\newblock In \emph{Advances in Neural Information Processing Systems 32: Annual Conference on Neural Information Processing Systems 2019}, pages 8024--8035.

\bibitem[{Peng et~al.(2019)Peng, Yan, and Lu}]{blue}
Yifan Peng, Shankai Yan, and Zhiyong Lu. 2019.
\newblock \href {https://doi.org/10.18653/v1/W19-5006} {Transfer learning in biomedical natural language processing: An evaluation of {BERT} and {ELM}o on ten benchmarking datasets}.
\newblock In \emph{Proceedings of the 18th BioNLP Workshop and Shared Task}, pages 58--65, Florence, Italy. Association for Computational Linguistics.

\bibitem[{Perez et~al.(2021)Perez, Kiela, and Cho}]{perez2021true}
Ethan Perez, Douwe Kiela, and Kyunghyun Cho. 2021.
\newblock \href {https://proceedings.neurips.cc/paper/2021/hash/5c04925674920eb58467fb52ce4ef728-Abstract.html} {True few-shot learning with language models}.
\newblock In \emph{Advances in Neural Information Processing Systems 34: Annual Conference on Neural Information Processing Systems 2021, NeurIPS 2021, December 6-14, 2021, virtual}, pages 11054--11070.

\bibitem[{Reimers and Gurevych(2019)}]{reimers2019sentence}
Nils Reimers and Iryna Gurevych. 2019.
\newblock \href {https://doi.org/10.18653/v1/D19-1410} {Sentence-{BERT}: Sentence embeddings using {S}iamese {BERT}-networks}.
\newblock In \emph{Proceedings of the 2019 Conference on Empirical Methods in Natural Language Processing and the 9th International Joint Conference on Natural Language Processing (EMNLP-IJCNLP)}, pages 3982--3992, Hong Kong, China. Association for Computational Linguistics.

\bibitem[{Ribeiro et~al.(2020)Ribeiro, Wu, Guestrin, and Singh}]{checklist}
Marco~Tulio Ribeiro, Tongshuang Wu, Carlos Guestrin, and Sameer Singh. 2020.
\newblock \href {https://doi.org/10.18653/v1/2020.acl-main.442} {Beyond accuracy: Behavioral testing of {NLP} models with {C}heck{L}ist}.
\newblock In \emph{Proceedings of the 58th Annual Meeting of the Association for Computational Linguistics}, pages 4902--4912, Online. Association for Computational Linguistics.

\bibitem[{Sarrouti et~al.(2021)Sarrouti, Ben~Abacha, Mrabet, and Demner-Fushman}]{healthver}
Mourad Sarrouti, Asma Ben~Abacha, Yassine Mrabet, and Dina Demner-Fushman. 2021.
\newblock \href {https://doi.org/10.18653/v1/2021.findings-emnlp.297} {Evidence-based fact-checking of health-related claims}.
\newblock In \emph{Findings of the Association for Computational Linguistics: EMNLP 2021}, pages 3499--3512, Punta Cana, Dominican Republic. Association for Computational Linguistics.

\bibitem[{Shivade(2017)}]{mednli}
Chaitanya Shivade. 2017.
\newblock \href {https://doi.org/10.13026/C2RS98} {Mednli — a natural language inference dataset for the clinical domain}.

\bibitem[{Singhal et~al.(2023)Singhal, Azizi, Tu, Mahdavi, Wei, Chung, Scales, Tanwani, Cole-Lewis, Pfohl et~al.}]{singhal2022large}
Karan Singhal, Shekoofeh Azizi, Tao Tu, S~Sara Mahdavi, Jason Wei, Hyung~Won Chung, Nathan Scales, Ajay Tanwani, Heather Cole-Lewis, Stephen Pfohl, et~al. 2023.
\newblock Large language models encode clinical knowledge.
\newblock \emph{Nature}.

\bibitem[{Su et~al.(2023)Su, Hou, Zhou, Rajendran, Maasch, Abedi, Zhang, Bai, Cuturrufo, Guo et~al.}]{su2023biomedical}
Chang Su, Yu~Hou, Manqi Zhou, Suraj Rajendran, Jacqueline~RMA Maasch, Zehra Abedi, Haotan Zhang, Zilong Bai, Anthony Cuturrufo, Winston Guo, et~al. 2023.
\newblock Biomedical discovery through the integrative biomedical knowledge hub (ibkh).
\newblock \emph{Iscience}, 26(4).

\bibitem[{Taboureau et~al.(2010)Taboureau, Nielsen, Audouze, Weinhold, Edsg{\"a}rd, Roque, Kouskoumvekaki, Bora et~al.}]{chemprot}
Olivier Taboureau, Sonny~Kim Nielsen, Karine Audouze, Nils Weinhold, Daniel Edsg{\"a}rd, Francisco~S Roque, Irene Kouskoumvekaki, Alina Bora, et~al. 2010.
\newblock Chemprot: a disease chemical biology database.
\newblock \emph{Nucleic acids research}, 39:D367--D372.

\bibitem[{Tang et~al.(2023)Tang, Han, Jiang, and Hu}]{tang2023does}
Ruixiang Tang, Xiaotian Han, Xiaoqian Jiang, and Xia Hu. 2023.
\newblock \href {https://arxiv.org/abs/2303.04360} {Does synthetic data generation of llms help clinical text mining?}
\newblock \emph{ArXiv preprint}, abs/2303.04360.

\bibitem[{Tsatsaronis et~al.(2015)Tsatsaronis, Balikas, Malakasiotis, Partalas, Zschunke, Alvers, Weissenborn, Krithara, Petridis, Polychronopoulos et~al.}]{bioasq}
George Tsatsaronis, Georgios Balikas, Prodromos Malakasiotis, Ioannis Partalas, Matthias Zschunke, Michael~R Alvers, Dirk Weissenborn, Anastasia Krithara, Sergios Petridis, Dimitris Polychronopoulos, et~al. 2015.
\newblock An overview of the bioasq large-scale biomedical semantic indexing and question answering competition.
\newblock \emph{BMC bioinformatics}, 16(1):1--28.

\bibitem[{Tu et~al.(2023)Tu, Azizi, Driess, Schaekermann, Amin, Chang, Carroll, Lau, Tanno, Ktena et~al.}]{tu2023towards}
Tao Tu, Shekoofeh Azizi, Danny Driess, Mike Schaekermann, Mohamed Amin, Pi-Chuan Chang, Andrew Carroll, Chuck Lau, Ryutaro Tanno, Ira Ktena, et~al. 2023.
\newblock \href {https://arxiv.org/abs/2307.14334} {Towards generalist biomedical ai}.
\newblock \emph{ArXiv preprint}, abs/2307.14334.

\bibitem[{Wang et~al.(2023)Wang, Zhou, and Sachan}]{wang2023lets}
Ruida Wang, Wangchunshu Zhou, and Mrinmaya Sachan. 2023.
\newblock \href {https://openreview.net/forum?id=GSNoZKqHgO} {Let's synthesize step by step: Iterative dataset synthesis with large language models by extrapolating errors from small models}.
\newblock In \emph{The 2023 Conference on Empirical Methods in Natural Language Processing}.

\bibitem[{Wang et~al.(2024)Wang, Li, Wang, Bai, Luo, Zhang, Jojic, Xing, and Hu}]{wang2023promptagent}
Xinyuan Wang, Chenxi Li, Zhen Wang, Fan Bai, Haotian Luo, Jiayou Zhang, Nebojsa Jojic, Eric~P Xing, and Zhiting Hu. 2024.
\newblock \href {https://openreview.net/forum?id=22pyNMuIoa} {Promptagent: Strategic planning with language models enables expert-level prompt optimization}.
\newblock In \emph{The Twelfth International Conference on Learning Representations}.

\bibitem[{Wei et~al.(2016)Wei, Peng, Leaman, Davis, Mattingly, Li, Wiegers, and Lu}]{cdr_dataset}
Chih-Hsuan Wei, Yifan Peng, Robert Leaman, Allan~Peter Davis, Carolyn~J Mattingly, Jiao Li, Thomas~C Wiegers, and Zhiyong Lu. 2016.
\newblock Assessing the state of the art in biomedical relation extraction: overview of the biocreative v chemical-disease relation (cdr) task.
\newblock \emph{Database}, 2016.

\bibitem[{Wolf et~al.(2019)Wolf, Debut, Sanh, Chaumond, Delangue, Moi, Cistac, Rault, Louf, Funtowicz et~al.}]{wolf2019huggingface}
Thomas Wolf, Lysandre Debut, Victor Sanh, Julien Chaumond, Clement Delangue, Anthony Moi, Pierric Cistac, Tim Rault, R{\'e}mi Louf, Morgan Funtowicz, et~al. 2019.
\newblock \href {https://arxiv.org/abs/1910.03771} {Huggingface's transformers: State-of-the-art natural language processing}.
\newblock \emph{ArXiv preprint}, abs/1910.03771.

\bibitem[{Wornow et~al.(2023)Wornow, Xu, Thapa, Patel, Steinberg, Fleming, Pfeffer, Fries, and Shah}]{wornow2023shaky}
Michael Wornow, Yizhe Xu, Rahul Thapa, Birju Patel, Ethan Steinberg, Scott Fleming, Michael~A Pfeffer, Jason Fries, and Nigam~H Shah. 2023.
\newblock \href {https://arxiv.org/abs/2303.12961} {The shaky foundations of clinical foundation models: A survey of large language models and foundation models for emrs}.
\newblock \emph{ArXiv preprint}, abs/2303.12961.

\bibitem[{Wu et~al.(2023)Wu, Zhang, Zhang, Wang, and Xie}]{wu2023pmcllama}
Chaoyi Wu, Xiaoman Zhang, Ya~Zhang, Yanfeng Wang, and Weidi Xie. 2023.
\newblock \href {https://arxiv.org/abs/2304.14454} {Pmc-llama: Further finetuning llama on medical papers}.
\newblock \emph{ArXiv preprint}, abs/2304.14454.

\bibitem[{Xie et~al.(2020)Xie, Dai, Hovy, Luong, and Le}]{uda}
Qizhe Xie, Zihang Dai, Eduard~H. Hovy, Thang Luong, and Quoc Le. 2020.
\newblock \href {https://proceedings.neurips.cc/paper/2020/hash/44feb0096faa8326192570788b38c1d1-Abstract.html} {Unsupervised data augmentation for consistency training}.
\newblock In \emph{Advances in Neural Information Processing Systems 33: Annual Conference on Neural Information Processing Systems 2020}.

\bibitem[{Xu et~al.(2023)Xu, Yu, Ho, and Yang}]{xu2023weakly}
Ran Xu, Yue Yu, Joyce Ho, and Carl Yang. 2023.
\newblock Weakly-supervised scientific document classification via retrieval-augmented multi-stage training.
\newblock In \emph{Proceedings of the 46th International ACM SIGIR Conference on Research and Development in Information Retrieval}, pages 2501--2505.

\bibitem[{Ye et~al.(2022{\natexlab{a}})Ye, Gao, Li, Xu, Feng, Wu, Yu, and Kong}]{ye2022zerogen}
Jiacheng Ye, Jiahui Gao, Qintong Li, Hang Xu, Jiangtao Feng, Zhiyong Wu, Tao Yu, and Lingpeng Kong. 2022{\natexlab{a}}.
\newblock \href {https://aclanthology.org/2022.emnlp-main.801} {{Z}ero{G}en: Efficient zero-shot learning via dataset generation}.
\newblock In \emph{Proceedings of the 2022 Conference on Empirical Methods in Natural Language Processing}, pages 11653--11669, Abu Dhabi, United Arab Emirates. Association for Computational Linguistics.

\bibitem[{Ye et~al.(2022{\natexlab{b}})Ye, Gao, Wu, Feng, Yu, and Kong}]{ye2022progen}
Jiacheng Ye, Jiahui Gao, Zhiyong Wu, Jiangtao Feng, Tao Yu, and Lingpeng Kong. 2022{\natexlab{b}}.
\newblock \href {https://aclanthology.org/2022.findings-emnlp.269} {{P}ro{G}en: Progressive zero-shot dataset generation via in-context feedback}.
\newblock In \emph{Findings of the Association for Computational Linguistics: EMNLP 2022}, pages 3671--3683, Abu Dhabi, United Arab Emirates. Association for Computational Linguistics.

\bibitem[{Yoo et~al.(2021)Yoo, Park, Kang, Lee, and Park}]{gpt3mix}
Kang~Min Yoo, Dongju Park, Jaewook Kang, Sang-Woo Lee, and Woomyoung Park. 2021.
\newblock \href {https://doi.org/10.18653/v1/2021.findings-emnlp.192} {{GPT}3{M}ix: Leveraging large-scale language models for text augmentation}.
\newblock In \emph{Findings of the Association for Computational Linguistics: EMNLP 2021}, pages 2225--2239, Punta Cana, Dominican Republic. Association for Computational Linguistics.

\bibitem[{Yu et~al.(2023)Yu, Zhuang, Zhang, Meng, Ratner, Krishna, Shen, and Zhang}]{yu2023large}
Yue Yu, Yuchen Zhuang, Jieyu Zhang, Yu~Meng, Alexander Ratner, Ranjay Krishna, Jiaming Shen, and Chao Zhang. 2023.
\newblock \href {https://openreview.net/forum?id=6hZIfAY9GD} {Large language model as attributed training data generator: A tale of diversity and bias}.
\newblock In \emph{Thirty-seventh Conference on Neural Information Processing Systems Datasets and Benchmarks Track}.

\bibitem[{Zellinger et~al.(2017)Zellinger, Grubinger, Lughofer, Natschl{\"{a}}ger, and Saminger{-}Platz}]{zellinger2017central}
Werner Zellinger, Thomas Grubinger, Edwin Lughofer, Thomas Natschl{\"{a}}ger, and Susanne Saminger{-}Platz. 2017.
\newblock \href {https://openreview.net/forum?id=SkB-\_mcel} {Central moment discrepancy {(CMD)} for domain-invariant representation learning}.
\newblock In \emph{5th International Conference on Learning Representations}.

\bibitem[{Zhang et~al.(2020)Zhang, Yu, and Zhang}]{seqmix}
Rongzhi Zhang, Yue Yu, and Chao Zhang. 2020.
\newblock \href {https://doi.org/10.18653/v1/2020.emnlp-main.691} {{S}eq{M}ix: Augmenting active sequence labeling via sequence mixup}.
\newblock In \emph{Proceedings of the 2020 Conference on Empirical Methods in Natural Language Processing (EMNLP)}, pages 8566--8579, Online. Association for Computational Linguistics.

\bibitem[{Zhang et~al.(2023)Zhang, Li, Cui, Cai, Liu, Fu, Huang, Zhao, Zhang, Chen et~al.}]{zhang2023siren}
Yue Zhang, Yafu Li, Leyang Cui, Deng Cai, Lemao Liu, Tingchen Fu, Xinting Huang, Enbo Zhao, Yu~Zhang, Yulong Chen, et~al. 2023.
\newblock \href {https://arxiv.org/abs/2309.01219} {Siren's song in the ai ocean: A survey on hallucination in large language models}.
\newblock \emph{ArXiv preprint}, abs/2309.01219.

\bibitem[{Zhou et~al.(2022)Zhou, Li, He, Bing, Cambria, Si, and Miao}]{melm}
Ran Zhou, Xin Li, Ruidan He, Lidong Bing, Erik Cambria, Luo Si, and Chunyan Miao. 2022.
\newblock \href {https://doi.org/10.18653/v1/2022.acl-long.160} {{MELM}: Data augmentation with masked entity language modeling for low-resource {NER}}.
\newblock In \emph{Proceedings of the 60th Annual Meeting of the Association for Computational Linguistics (Volume 1: Long Papers)}, pages 2251--2262, Dublin, Ireland. Association for Computational Linguistics.

\bibitem[{Zhou et~al.(2023)Zhou, Muresanu, Han, Paster, Pitis, Chan, and Ba}]{zhou2023large}
Yongchao Zhou, Andrei~Ioan Muresanu, Ziwen Han, Keiran Paster, Silviu Pitis, Harris Chan, and Jimmy Ba. 2023.
\newblock \href {https://openreview.net/forum?id=92gvk82DE-} {Large language models are human-level prompt engineers}.
\newblock In \emph{The Eleventh International Conference on Learning Representations}.

\end{thebibliography}
\clearpage

\appendix
\section{Details on the Calculation of CMD}
\label{sec:cmd}
We introduce the Central Moment Discrepancy (CMD)~\citep{zellinger2017central}, which is a widely used metric to measure the domain shift in the area of domain-invariant representation learning. Let $X=\left(x_1, \ldots, x_n\right)$ and $Y=\left(y_1, \ldots, y_n\right)$ be bounded feature vectors independent and identically distributed from two probability distributions $p$ and $q$. The central moment discrepancy metric (CMD) is defined by
\begin{align}
   \operatorname{CMD}(p, q)&=\frac{1}{|b-a|}\|\mathbb{E}(X)-\mathbb{E}(Y)\|_2 \nonumber \\
   &+\sum_{k=2}^{\infty} \frac{1}{|b-a|^k}\left\|c_k(X)-c_k(Y)\right\|_2 
   \nonumber
\end{align} 
where $\mathbb{E}(X)$ is the expectation of $X$, and
\begin{equation}
    c_k(X)=\left(\mathbb{E}\left(\prod_{i=1}^N\left(X_i-\mathbb{E}\left(X_i\right)\right)^{r_i}\right)\right)_{\substack{r_1+\ldots+r_N=k \\ r_1, \ldots, r_n \geq 0}}
    \nonumber
\end{equation}
is the central moment vector of order $k$.
To estimate the CMD efficiently without infinite-order calculation, we follow~\citep{zellinger2017central} and use a $K$-order approximation of CMD as 
\begin{align}
\operatorname{CMD}_k(p, q)&=\frac{1}{|b-a|}\|\mathbf{E}(X)-\mathbf{E}(Y)\|_2
 \nonumber\\
&+\sum_{k=2}^K \frac{1}{|b-a|^k}\left\|C_k(X)-C_k(Y)\right\|_2
   \nonumber
\end{align} 
where $\mathbf{E}(X)=\frac{1}{|X|} \sum_{x \in X} x$ is the empirical expectation vector computed on the sample $X$ and $C_k(X)=\mathbf{E}\left((x-\mathbf{E}(X))^k\right)$ is the vector of all $k^{\text {th }}$ order sample central moments of the coordinates of $X$\footnote{The implementation of CMD 
 is available at \url{https://gist.github.com/yusuke0519/724aa68fc431afadb0cc7280168da17b}}.  
 To adapt CMD in our work, we set $K=5$, and use the embedding from SentenceBERT~\citep{reimers2019sentence} to calculate the embedding $X, Y$ for instances. 
\section{Additional Preliminary Studies}
\label{sec:add_prelim}

\begin{figure*}[t]
	\centering
	% \vspace{-2ex}
	\subfigure[LitCovid]{
		\includegraphics[width=0.31\linewidth]{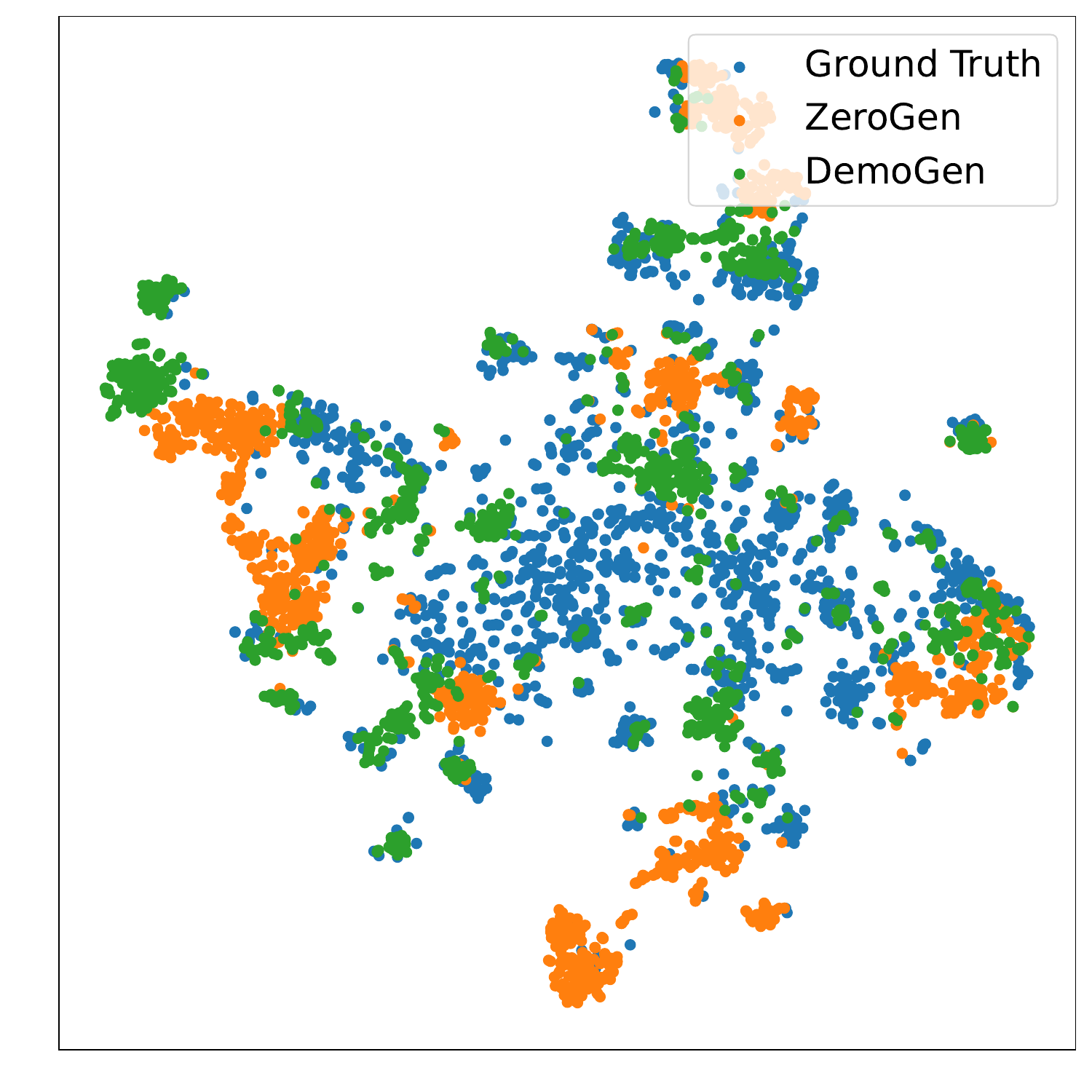}
	} %\hfill
         % \hspace{-1.5ex}
     \subfigure[GAD]{
		\includegraphics[width=0.31\linewidth]{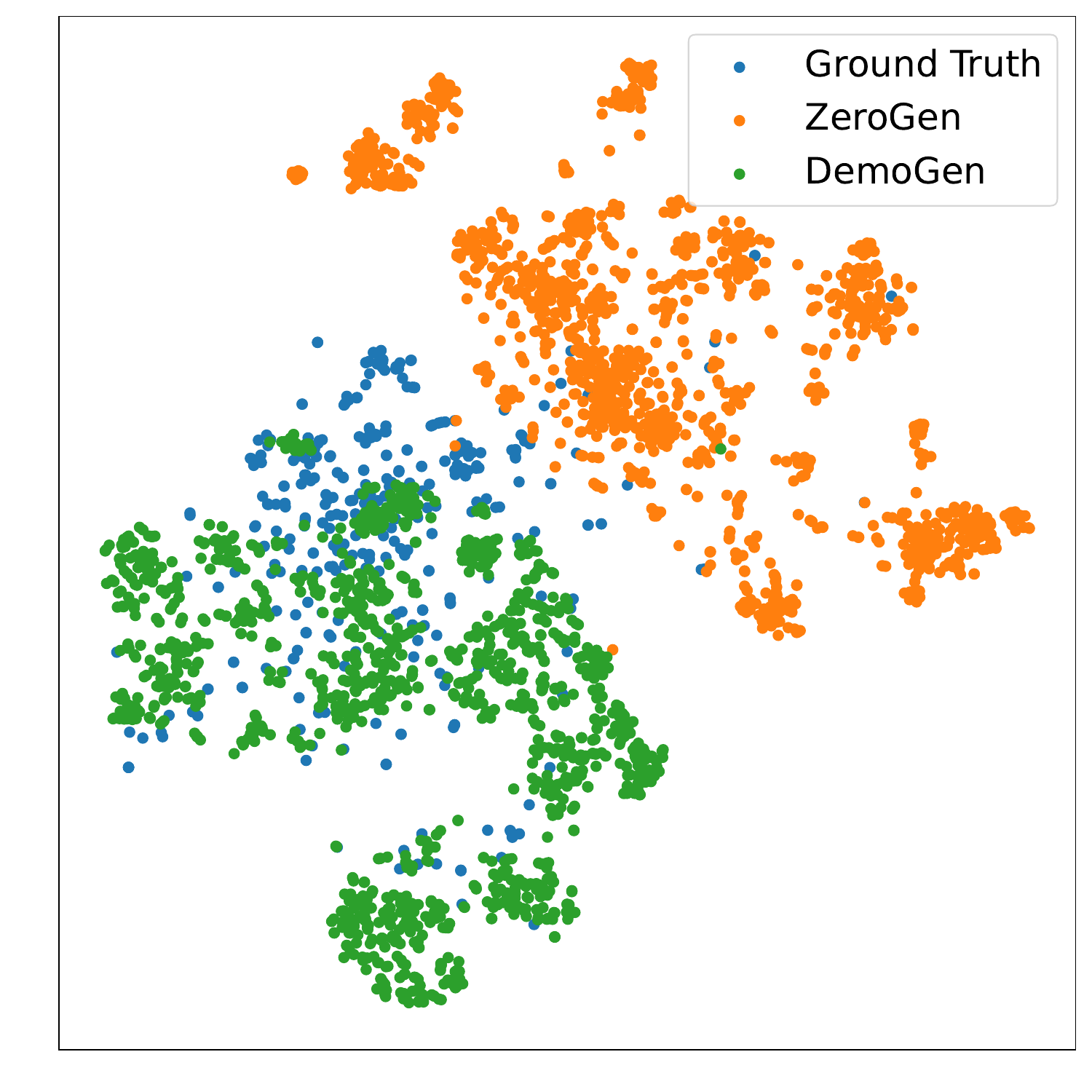}
	}
        % \hspace{-1.5ex}
      \subfigure[CDR]{
		\includegraphics[width=0.31\linewidth]{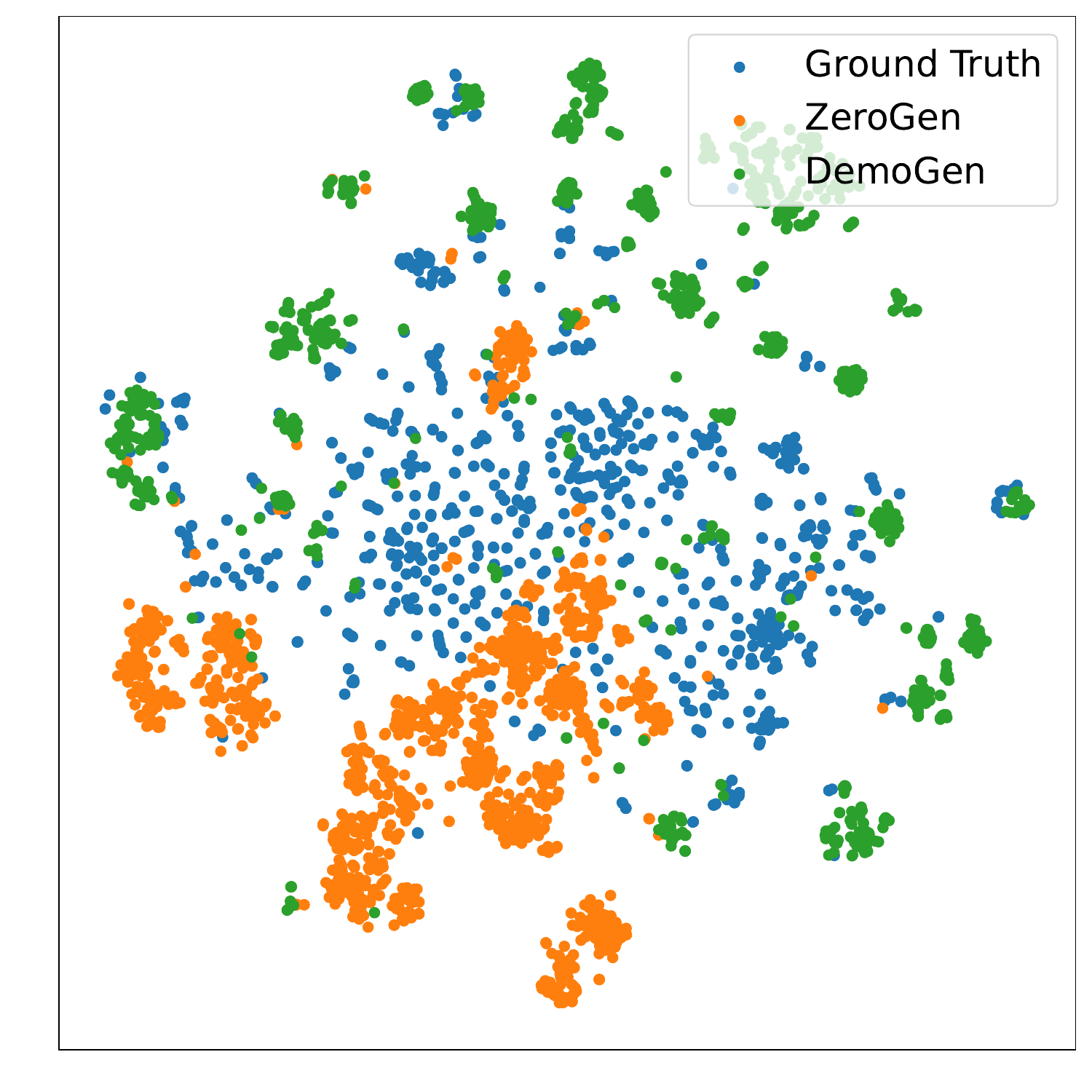}
	}
 	\subfigure[MEDIQA-RQE]{
		\includegraphics[width=0.31\linewidth]{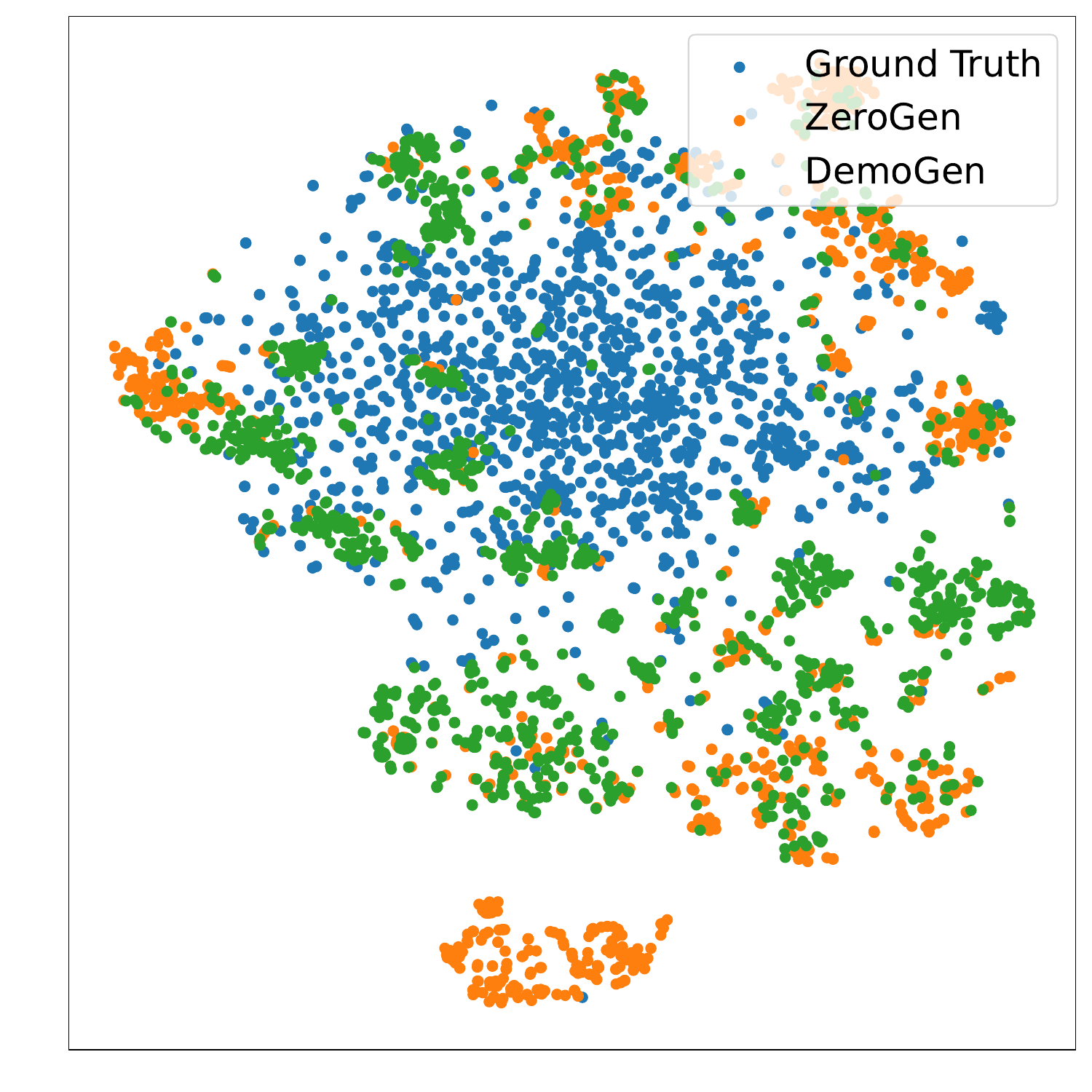}
	} %\hfill
         % \hspace{-1.5ex}
     \subfigure[MQP]{
		\includegraphics[width=0.31\linewidth]{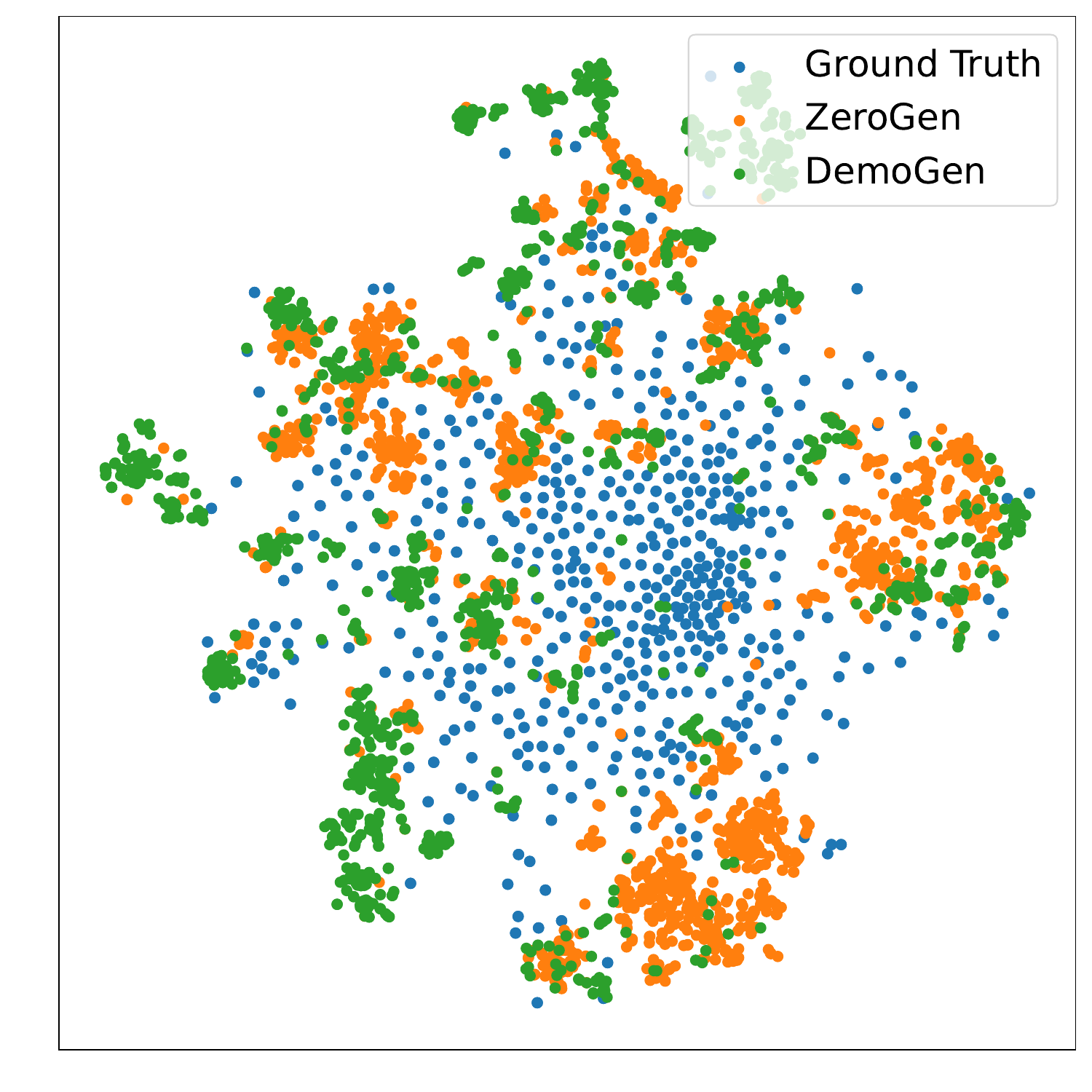}
	}
        % \hspace{-1.5ex}
      \subfigure[CHEMDNER]{
		\includegraphics[width=0.31\linewidth]{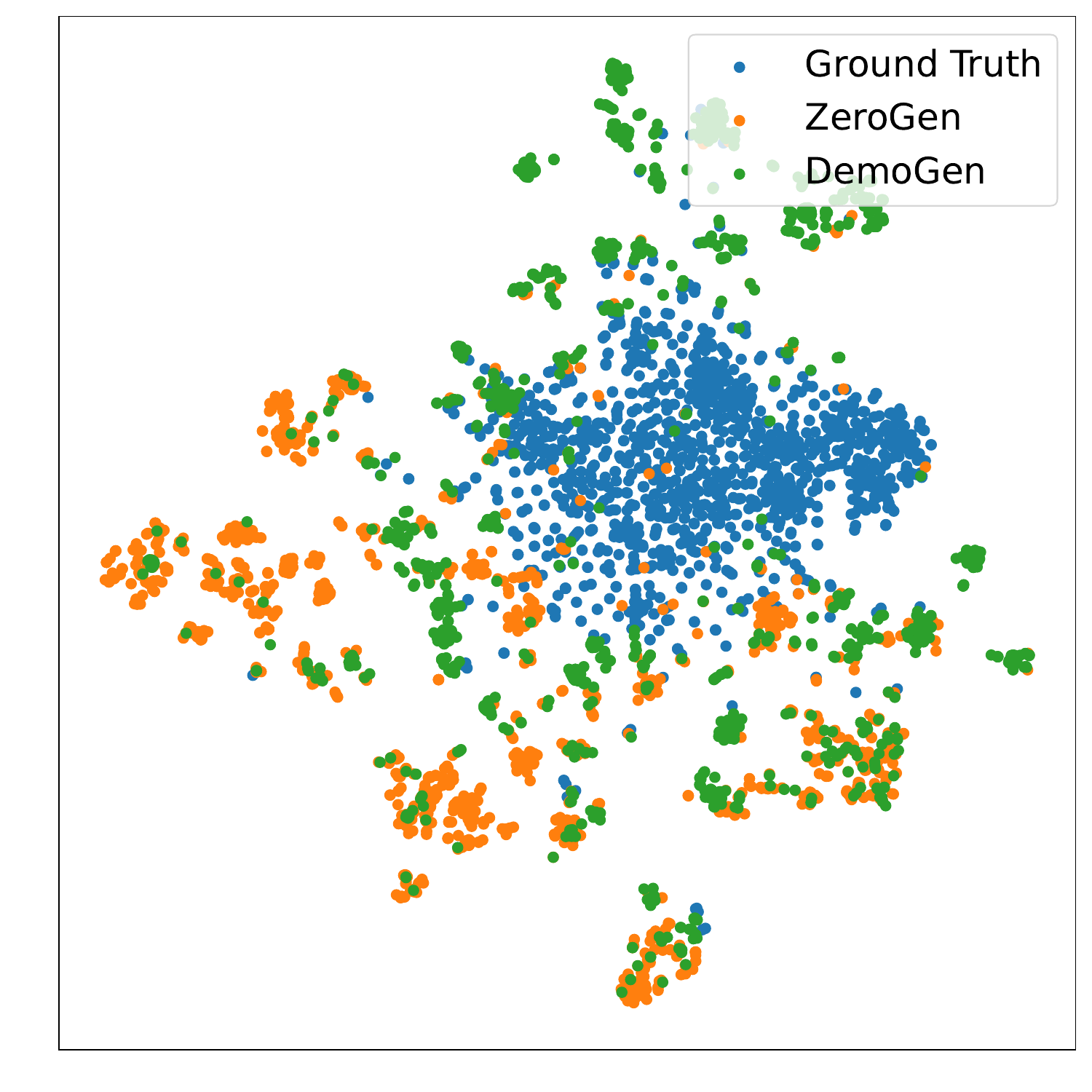}
	}
	\caption{The t-SNE plots of datasets generated by ZeroGen and DemoGen compared with the ground truth.}
\label{fig:add_prelim_tsne}
\end{figure*}

\begin{figure*}[t]
	\centering
	% \vspace{-2ex}
	\subfigure[LitCovid]{
		\includegraphics[width=0.31\linewidth]{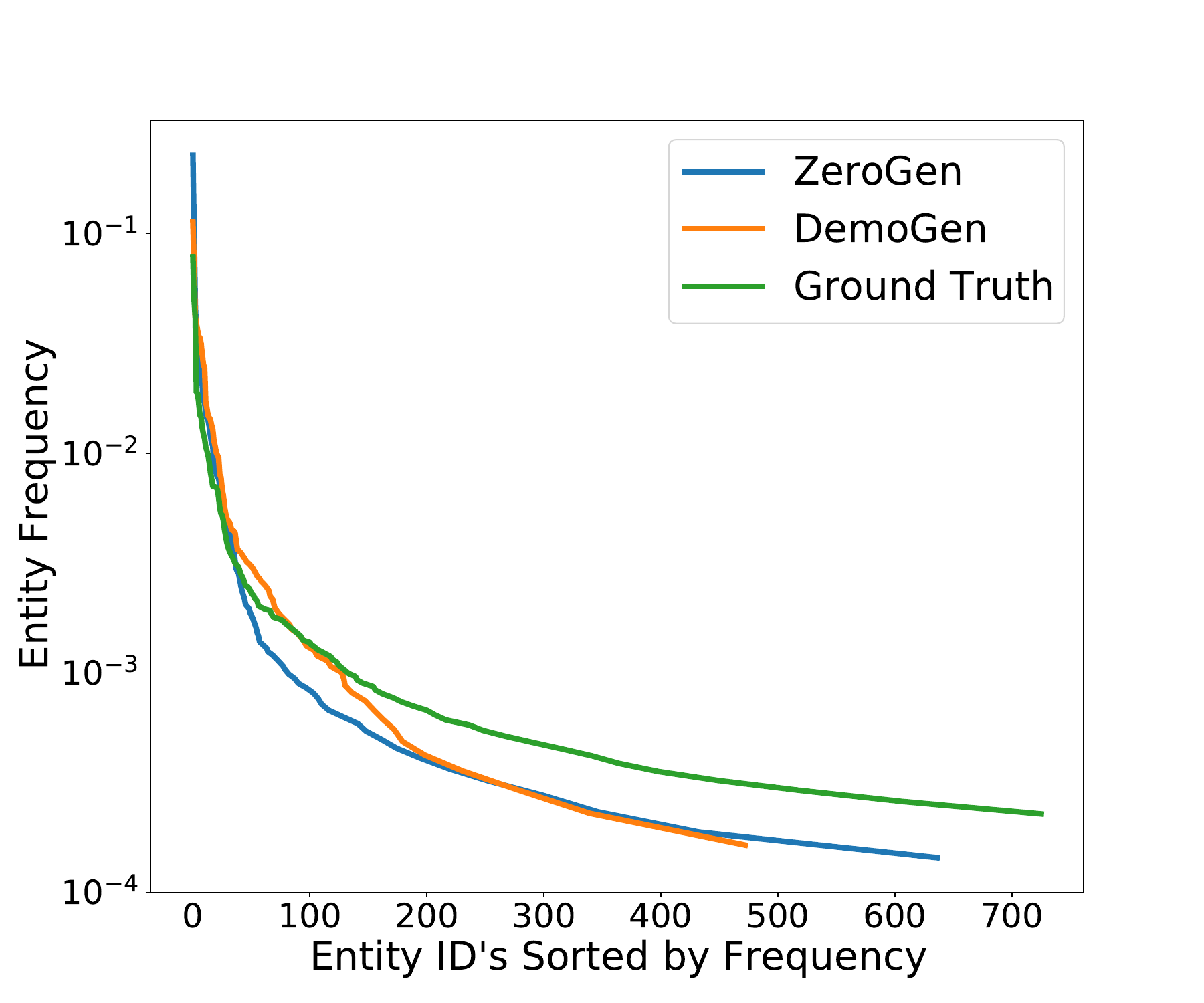}
	} %\hfill
         % \hspace{-1.5ex}
     \subfigure[GAD]{
		\includegraphics[width=0.31\linewidth]{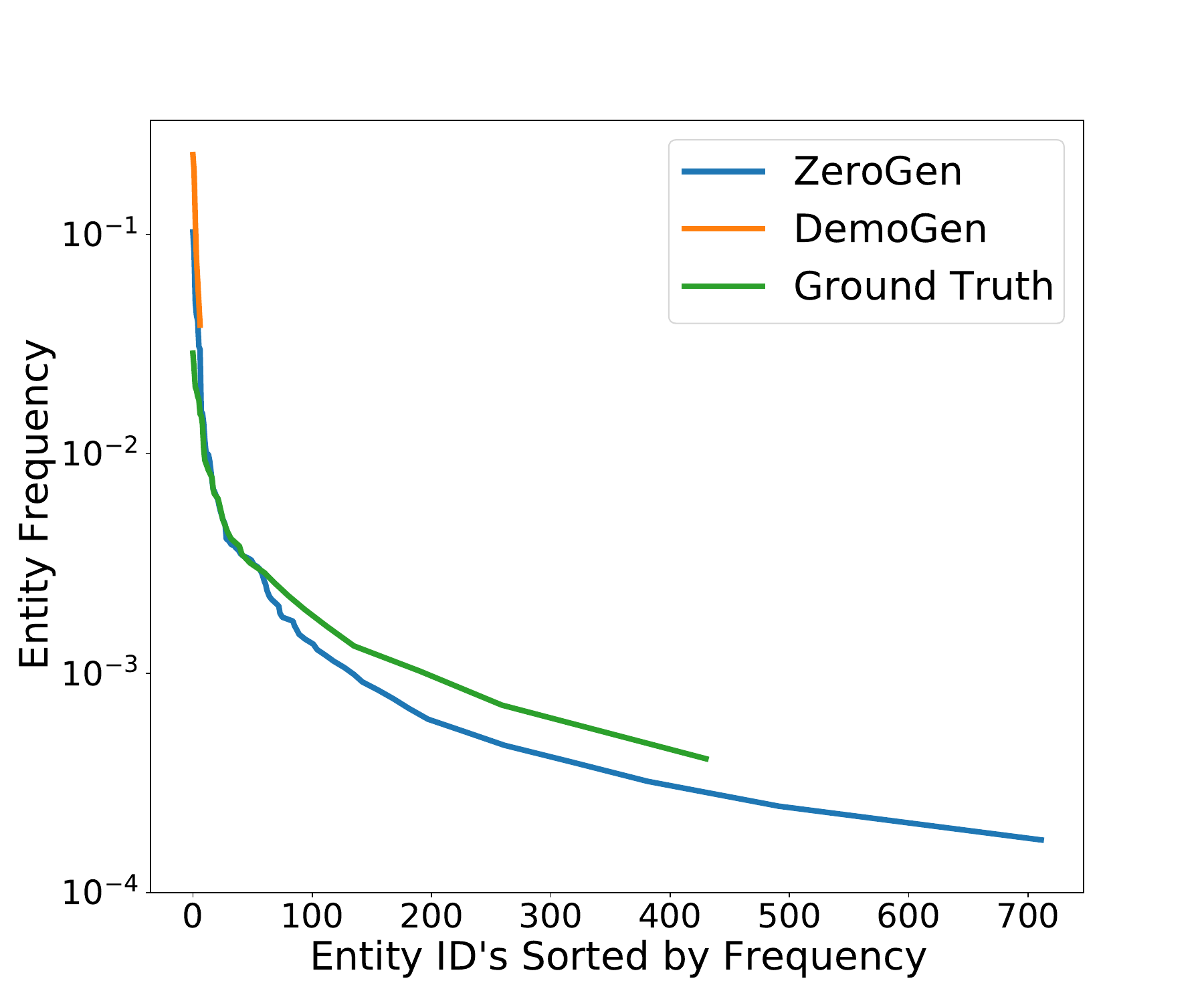}
	}
        % \hspace{-1.5ex}
      \subfigure[CDR]{
		\includegraphics[width=0.31\linewidth]{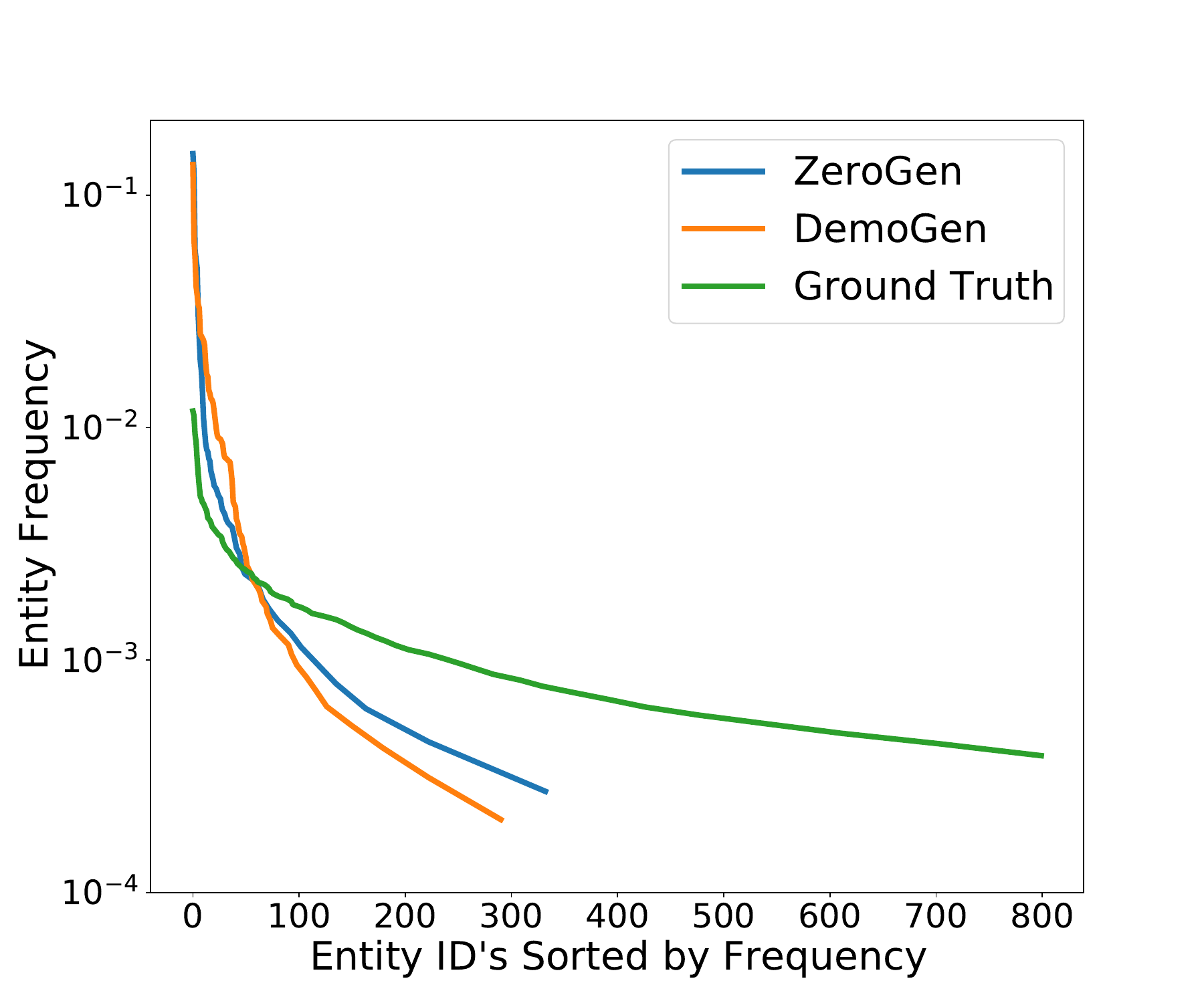}
	}
 	\subfigure[MEDIQA-RQE]{
		\includegraphics[width=0.31\linewidth]{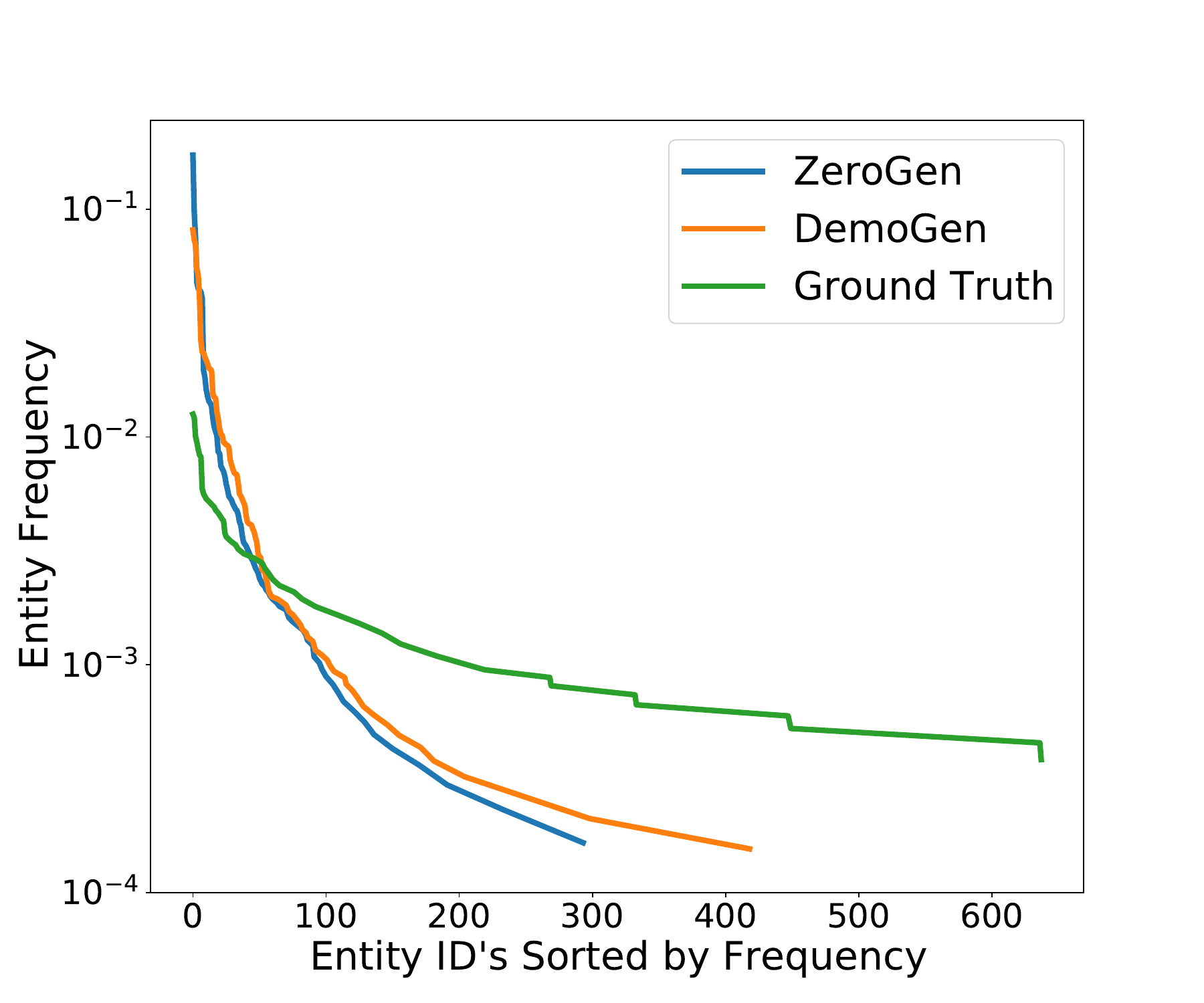}
	} %\hfill
         % \hspace{-1.5ex}
     \subfigure[MQP]{
		\includegraphics[width=0.31\linewidth]{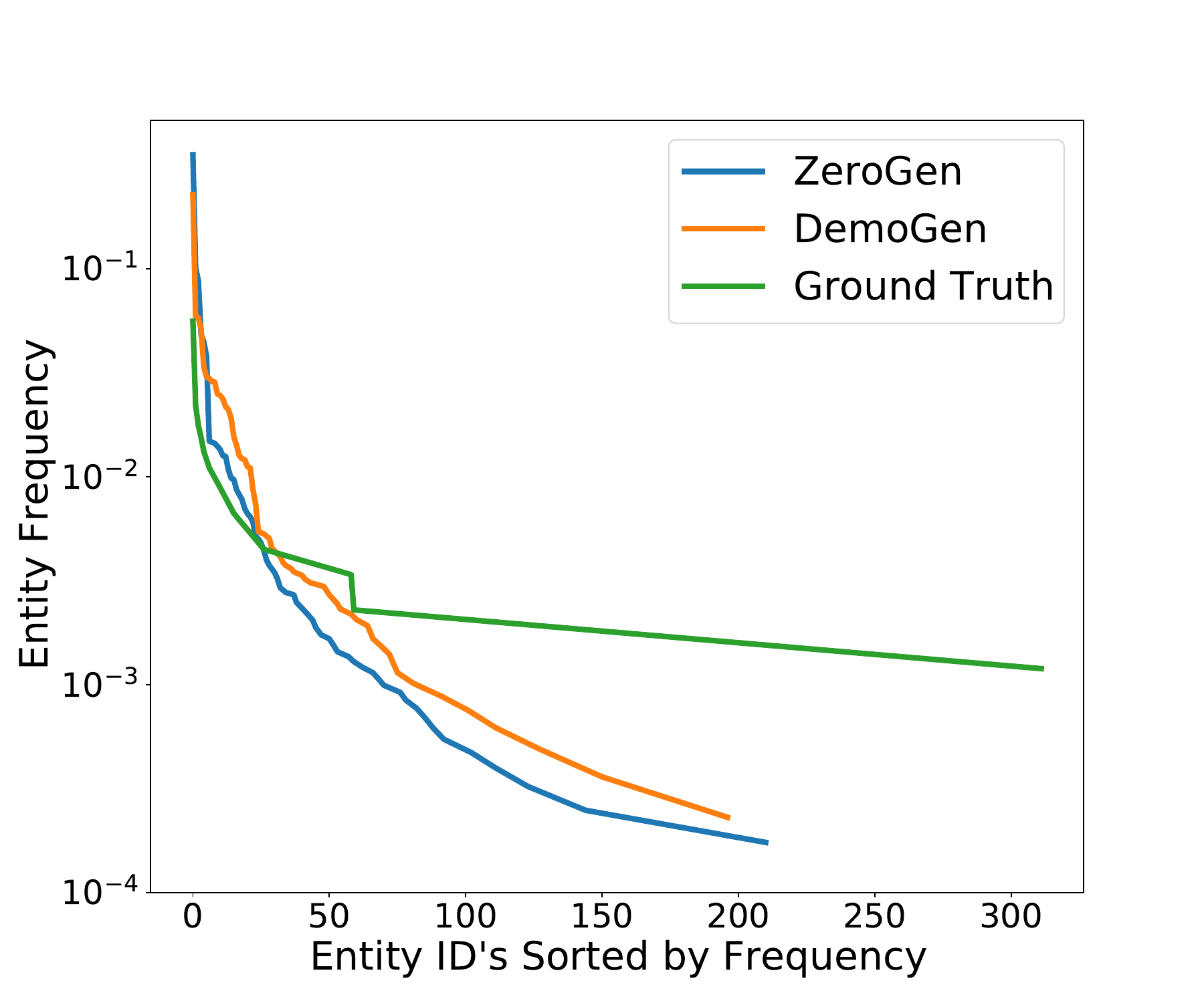}
	}
        % \hspace{-1.5ex}
      \subfigure[CHEMDNER]{
		\includegraphics[width=0.31\linewidth]{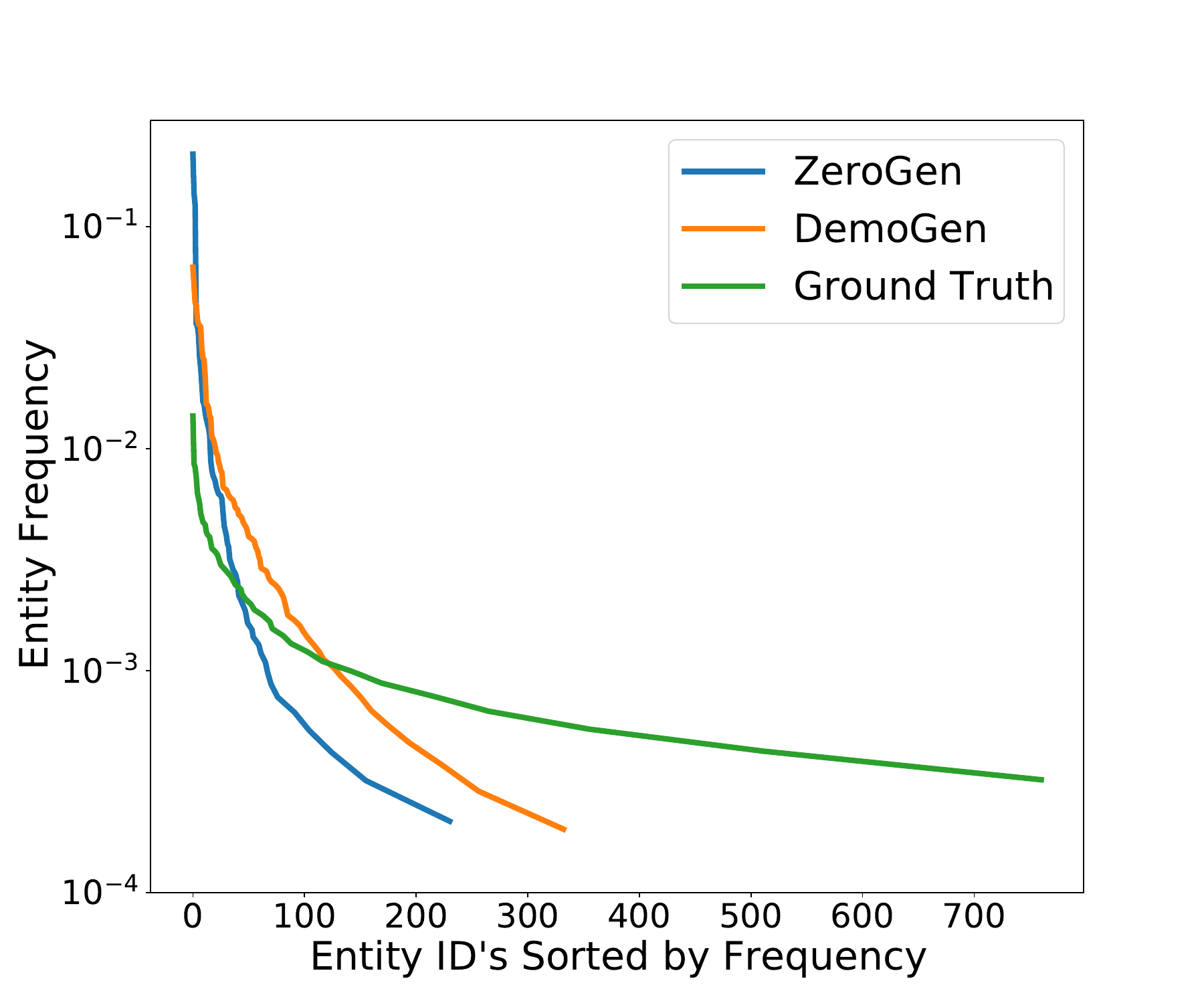}
	}
	\caption{The regularized entity frequencies of datasets generated by ZeroGen and DemoGen compared with the ground truth in log scale.}
\label{fig:add_prelim_freq}
\end{figure*}

We present additional preliminary studies of the t-SNE plots in Figure~\ref{fig:add_prelim_tsne} and the regularized entity frequencies in Figure~\ref{fig:add_prelim_freq}.
In Figure \ref{fig:add_prelim_tsne}, we visualize the embeddings\footnote{We employ SentenceBERT~\citep{reimers2019sentence} as the text encoder.} of both the ground truth training data and synthetic datasets generated via two representative methods. Overall, these methods use generic prompts (see Appendix~\ref{sec:prompt_format_bsl} for details) with minimal domain-specific constraints.
These results further justify the distribution shift issue mentioned in section \ref{sec:limitations}, demonstrating that the limited diversity as well as the distribution shift issue generally exists for a broad range of clinical NLP tasks.
% \vspace{-1.6ex}

\begin{figure*}[t!]
	\centering
	% \vspace{-3ex}
	% \subfigure[t-SNE plot]{
	% 	\includegraphics[width=0.28\linewidth]{figures/bc5cdr_disease_sentencebert_bsl.pdf}
	% 	\label{fig:bc5cdr_disease_sentencebert_bsl}
	% } %\hfill
 %         \hspace{-2.3ex}
	% \subfigure[MEDIQA-RQE]{
	% 	\includegraphics[width=0.25\linewidth]{figures/mediqa_rqe_sentencebert_bsl.pdf}
	% 	\label{fig:mediqa_rqe_sentencebert_bsl}
	% }\hspace{-1.5ex}
		\includegraphics[width=0.9\linewidth]{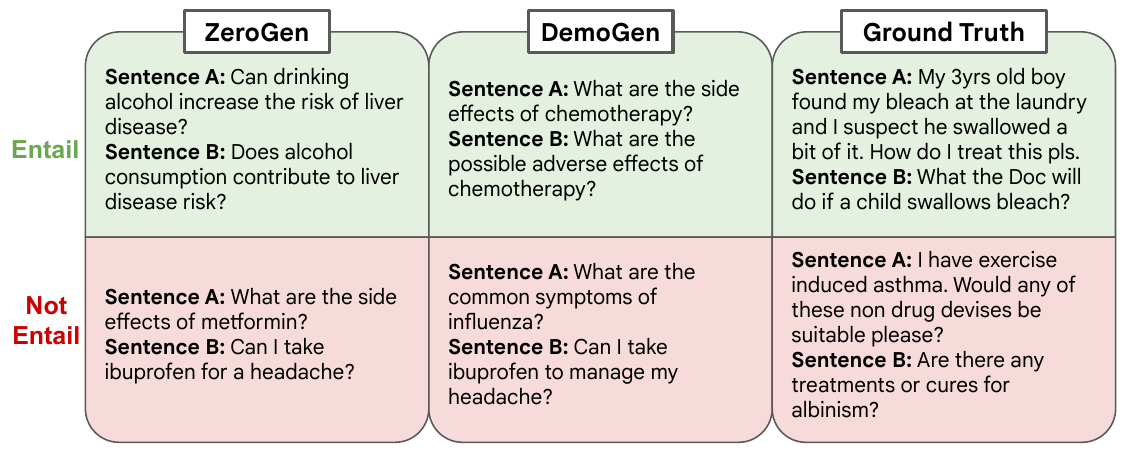}
		% \label{fig:case_study_prelim}
	\caption{Case study of generated samples by existing methods ZeroGen and DemoGen.}
\label{fig:prelim1}
% \vspace{-3ex}
\end{figure*}

Figure~\ref{fig:prelim1} shows a case study, where we randomly select one sample from each class within the training set generated by ZeroGen and DemoGen. These selected samples are compared with the ground truth data from the MEDIQA-RQE dataset, which aims to predict whether a consumer health query can entail an existing Frequently Asked Question (FAQ). 
It is evident that the samples generated by ZeroGen and DemoGen exhibit a limited range of writing styles and tend to follow a specific template, whereas the ground truth sample contains more contextual elements that are typically encountered in real-life scenarios.

\section{Dataset Description}
\label{sec:dataset_description}

% \begin{table}[hbtp]
% \floatconts
%     {tab:datasets}
%     {\caption{Dataset statistics. For \# of hyperedges in MIMIC-III, the first number indicates the hyperedges without labels, while the second one indicates ones with labels.}}
%     {
%     \resizebox{0.8\linewidth}{!}{
%     \begin{tabular}{lcc}
%     \toprule
%     \bfseries Stats & \bfseries MIMIC-III    & \bfseries \dataset          \\ 
%     \midrule 
%     \# of diagnosis & 846 & 7915\\
%     \# of medication & 4525 & 489\\
%     \# of procedure & 2032 & 4321\\
%     \# of service & 20 & ---\\
%     \# of hyperedges & 36875/12353 & 36611 \\
%     \bottomrule
%     \end{tabular}
%     }
%     }
% \end{table}

\begin{table*}[t]
% \floatconts
  \caption{Dataset statistics. We do not count the non-entity/non-relation class for relation extraction and token classification tasks to align with existing works. P and R stand for Precision and Recall. Metrics in \textbf{bold} are considered as the main metrics. $*$ is not allowed to put into GPT and $\dagger$ does not provide training data, so we sample few-shot examples from the SciTail~\citep{khot2018scitail} instead.}
  \label{tab:datastats}
  
  {
  \resizebox{0.95\linewidth}{!}{
  \begin{tabular}{lcccc}
  \toprule
  \bfseries Corpus & \bfseries Tasks & \bfseries \#Class & \bfseries \#Train/\#Test & \bfseries Metrics \\
  \midrule
  \multicolumn{5}{l}{\textbf{Single-Sentence Tasks}}\\
  \midrule
  LitCovid~\citep{litcovid} & Text Classification & 7 & 24960/6238 & \textbf{F1}\\
  HOC~\citep{hoc} & Text Classification & 10 & 3091/898 &\textbf{F1}\\
  GAD~\citep{gad} & Relation Extraction (RE) & 1 & 4750/350 & P, R, \textbf{F1}\\
  CDR~\citep{cdr_dataset} & Relation Extraction (RE) & 1 & 8431/2522 &  P, R, \textbf{F1}\\
  ChemProt~\citep{chemprot} & Relation Extraction (RE) & 5 & 8793/1087 & \textbf{F1}\\
  \midrule
  \multicolumn{5}{l}{\textbf{Sentence-Pair Tasks}}\\
  \midrule
  MedNLI$^*$~\citep{mednli} & Natural Language Inference (NLI) & 3 & 11232/1422 & \textbf{Acc}\\
  MEDIQA-NLI$^\dagger$~\citep{mediqa-nli} & Natural Language Inference (NLI) & 3 & -/405 & \textbf{Acc}\\
  MEDIQA-RQE~\citep{abacha2016recognizing} & Natural Language Inference (NLI) & 2 & 8588/302 & \textbf{Acc}\\
  PUBHEALTH~\citep{PUBHEALTH} & Fact Verification & 4 & 9804/1231 & Acc, \textbf{F1}\\
  HealthVer~\citep{healthver} & Fact Verification & 3 & 10591/1824 & Acc, \textbf{F1}\\
  MQP~\citep{mqp} & Sentences Similarity (STS) & 2 & 10/3033 & \textbf{Acc} \\
  PubmedQA~\citep{jin2019pubmedqa} & Question Answering (QA) & 2 & 500/500 & \textbf{Acc} \\
  BioASQ~\citep{bioasq} & Question Answering (QA) & 2 & 670/140 &   \textbf{Acc}\\
  \midrule
  \multicolumn{5}{l}{\textbf{Token Classification Tasks}}\\
  \midrule
  BC5CDR-Disease~\citep{bc5cdr} & Named Entity Recognition (NER) & 1 & 4882/5085 & P, R, \textbf{F1}\\
  BC5CDR-Chemical~\citep{bc5cdr} & Named Entity Recognition (NER) & 1 & 4882/5085 & P, R, \textbf{F1}\\
  NCBI-Disease~\citep{ncbi-disease} & Named Entity Recognition (NER) & 1 & 5336/921 &  P, R, \textbf{F1}\\
  CHEMDNER~\citep{chemdner} & Named Entity Recognition (NER) & 1 & 14522/12430 &  P, R, \textbf{F1}\\
  CASI~\citep{agrawal2022large,claim} & Attribute Extraction & 6 & 5/100 & \textbf{F1}\\
  \bottomrule
  \end{tabular}
  }}
\end{table*}

The evaluation tasks and datasets are summarized in Table \ref{tab:datastats}. Note that the number of training samples indicates the size of the \textit{original} training set. Specifically, we consider the following datasets:

\begin{itemize}[leftmargin=0.2cm]

\item \textbf{Single-Sentence Tasks}
\begin{itemize}[label=$\circ$]
% \noindent$\bullet$ \textbf{Text Classification}: 
\item \uline{Text Classification}: 

\begin{itemize}[leftmargin=0.2cm]
    \item The \textit{LitCovid} dataset~\citep{litcovid} consists of COVID-19-related publications from PubMed. The task is to predict the topics of the sentences, including ``Epidemic Forecasting", ``Treatment", ``Prevention", ``Mechanism", ``Case Report", ``Transmission", and ``Diagnosis".
    \item The \textit{HOC} dataset~\citep{hoc} also extracts sentences from PubMed articles, each annotated at the sentence level. The task is to predict the topics of the sentences, including ``evading growth suppressors", ``tumor promoting inflammation", ``enabling replicative immortality", ``cellular energetics", ``resisting cell death", ``activating invasion and metastasis", genomic instability and mutation", ``inducing angiogenesis", ``sustaining proliferative signaling", and ``avoiding immune destruction".
\end{itemize}

% Our evaluation aligns with prior research~\citep{blue}, focusing on the document-level F1-score for the multi-label text classification task.

% \noindent$\bullet$ \textbf{Relation Extraction}: 
\item \uline{Relation Extraction}:
\begin{itemize}[leftmargin=0.2cm]
    \item The \textit{GAD}~\citep{gad} dataset is to predict whether there is a relation between the given disease and gene in the sentences. Note that the original annotation for this dataset is Noisy. To remedy this issue, we \emph{relabel} 350 examples from the original test set to form a clean subset for faithful evaluation.
    \item The \textit{CDR}~\citep{cdr_dataset} dataset is to predict whether the provided chemical can induce the disease in the sentences.
    \item The \textit{ChemProt}~\citep{chemprot} dataset focuses on the chemical-protein relations, and the labels include ``Upregulator", ``Downregulator", ``Agonist", ``Antagonist", ``Product\_of" and ``No relation".
\end{itemize}
% The relation extraction tasks involve identifying relationships and their associated types among entities within sentences. These tasks encompass the detection of gene-disease relation in the \textit{GAD}~\citep{gad} dataset, chemical-disease relation in the \textit{CDR}~\citep{cdr_dataset} dataset, and chemical-protein interactions in the \textit{ChemProt}~\citep{chemprot} dataset. 
% Performance evaluation hinges on a comparison of the predicted relation types against the annotated data, utilizing metrics precision, recall, and F1-score.
\end{itemize}

\item \textbf{Sentence-Pair Tasks}
\begin{itemize}[label=$\circ$]
% \noindent$\bullet$ \textbf{Natural Language Inference (NLI)}: 
\item \uline{Natural Language Inference (NLI)}: 

\begin{itemize}[leftmargin=0.2cm]
    \item The \textit{MedNLI}~\citep{mednli} dateset consists of sentences pairs derived from MIMIC-III, where we predict the relations between the sentences. The labels include ``entailment", ``neutral" and ``contradiction".
    \item The \textit{MEDIQA-NLI}~\citep{mediqa-nli} dataset  comprises text-hypothesis pairs. Their relations include ``entailment", ``neutral" and ``contradiction".
    \item The \textit{MEDIQA-RQE}~\citep{abacha2016recognizing} dataset contains NIH consumer health question pairs, and the task is to recognize if the first question can entail the second one. 
\end{itemize}
% To identify inference relations between sentences, we leveraged three clinical NLI datasets: \textit{MedNLI*}~\citep{mednli} with 14,049 pairs derived from MIMIC-III, \textit{MEDIQA-NLI}~\citep{mediqa-nli} comprising 405 text-hypothesis pairs, and \textit{MEDIQA-RQE}~\citep{abacha2016recognizing} with 230 NIH consumer health question pairs for recognizing question entailment (RQE). 
% We evaluate NLI and RQE tasks using accuracy as our primary evaluation metric.

% \noindent$\bullet$ \textbf{Fact Verification}: 
\item \uline{Fact Verification}:
\begin{itemize}[leftmargin=0.2cm]
    \item The \textit{PUBHEALTH}~\citep{PUBHEALTH} encompasses claims paired with journalist-crafted explanations. The task is to predict the relations between the claim and evidence, including ``Refute", ``Unproven", ``Support", and ``Mixture".
    \item The \textit{HealthVer}~\citep{healthver} contains evidence-claim pairs from search engine snippets regarding COVID-19 questions. The relations between claims and evidences are chosen from ``Refute", ``Unproven", and ``Support".
\end{itemize}

\item \uline{Question Answering (QA)}:
\begin{itemize}[leftmargin=0.2cm]
   \item The \textit{PubmedQA} task \citep{jin2019pubmedqa} entails responding to inquiries regarding the abstracts of biomedical research papers.
\item The \textit{BioASQ} task \citep{bioasq} spans multiple question types, including factoid, list, summary, and yes/no questions derived from expert-reviewed biomedical research papers.
\end{itemize}

% In verifying claims against credible evidence, our investigation delves into two real-world evidence-based fact-checking datasets: \textit{PUBHEALTH}~\citep{PUBHEALTH}
% encompassing 11.8K claims paired with journalist-crafted explanations, and \textit{HealthVer}~\citep{healthver} containing 14,330 evidence-claim pairs from search engine snippets regarding COVID-19 questions.
% Our evaluation of the verification prediction task is based on accuracy and F1-score as the key performance metrics.

% \noindent$\bullet$ \textbf{Sentence Similarity (STS)}: 
\item \uline{Sentence Similarity (STS)}:
\begin{itemize}[leftmargin=0.2cm]
    \item the \textit{MQP}~\citep{mqp} dataset comprises a collection of medical question pairs designed for identifying semantically similar questions. The task is to predict whether the two questions are equivalent or not.
\end{itemize}
% \textit{MQP}~\citep{mqp} dataset comprises a collection of 4,567 unique medical question pairs designed for identifying semantically similar questions. Our evaluation of model performance involves comparing the predicted similarity scores with the ground truth labels, and we report accuracy as the comparison metric.
\end{itemize}

\item \textbf{Token Classification Tasks}
\begin{itemize}[label=$\circ$]
% \noindent$\bullet$ \textbf{Named Entity Recognition (NER)}: 
\item \uline{Named Entity Recognition (NER)}:
\begin{itemize}[leftmargin=0.2cm]
    \item The \textit{BC5CDR-Disease}~\citep{bc5cdr} is to recognize diseases in the sentences.
    \item The \textit{BC5CDR-Chemical}~\citep{bc5cdr} is to recognize chemicals in the sentences.
    \item The \textit{NCBI-Disease}~\citep{ncbi-disease} is to recognize diseases in the sentences.
    \item The \textit{CHEMDNER}~\citep{chemdner} is to recognize chemicals in the sentences.
\end{itemize}
% For recognizing and predicting entities (e.g., diseases, chemicals) from text, we examine three datasets commonly used for biomedical NER: \textit{BC5CDR}~\citep{li2016biocreative}, \textit{NCBI-Disease}~\citep{ncbi-disease}, \textit{CHEMDNER}~\citep{chemdner}. Our evaluation compares annotated mention spans in the documents to model predictions, utilizing precision, recall, and F1-score.

% \noindent$\bullet$ \textbf{Attribute Extraction (MedAttr)}:
\item \uline{Attribute Extraction (MedAttr)}:
\begin{itemize}[leftmargin=0.2cm]
    \item The \textit{CASI} dataset~\citep{agrawal2022large,claim} aims to identify interventions including medication, dosage, route, freq, reason, duration
\end{itemize}
% For biomedical evidence extraction, we concentrate on identifying interventions from the manually re-annotated \textit{CASI} dataset~\citep{agrawal2022large,claim}, using the F1-score.
\end{itemize}
\end{itemize}

\section{Baseline Details}
\label{sec:baseline_details}

% The data augmentation (DA) models include Word Substitution~\citep{checklist}, Back Translation~\citep{uda}, Mixup~\citep{chen2020mixtext,seqmix}, Transformer~\citep{kumar2020data,melm}, LightNER~\citep{lightner}, and KGPC~\citep{chen2023few}.
% \yy{}
% Note that LightNER and KGPC are designed specifically for NER tasks. Back Translation cannot be applied to NER and RE tasks as it's non-trivial to locate the related entities in the generated sentences. 
% For LLM-based generation models, we consider ZeroGen~\citep{ye2022zerogen}, DemoGen~\citep{meng2023tuning,gpt3mix}, ProGen~\citep{ye2022progen} and S3~\citep{wang2023lets} as representative methods. 

In this section, we give a detailed introduction for all baselines used in this study.

\textbf{Data Augmentation Methods:}
\begin{itemize}[leftmargin=0.5cm]
    \item \textbf{DA-Word Sub}~\citep{checklist}: It performs word substitution for few-shot demonstrations to create new training sample. Specifically, we 
    follow Checklist~\citep{checklist} and maintain a word list to generate new examples. 
    \item \textbf{DA-Back Translation}~\citep{uda}: It employ back translation to augment the training data~\cite{uda}, including translating text from the target language to the source language and then back to the target language. 
    \item \textbf{DA-Mixup}~\citep{chen2020mixtext,seqmix}: It adds interpolation on the \emph{embedding space} of the training examples to create virtual augmented examples.
    \item \textbf{DA-Transformer (MELM)}~\citep{kumar2020data,melm}: It introduces a conditional data augmentation technique that prepends class labels to text sequences for pre-trained transformer-based models. Specifically, it leverage the sequence to sequence transformer to perform conditional text generation based on the seed examples. 
    \item  \textbf{LightNER}~\citep{lightner}: It adopts a seq2seq framework, generating the entity span sequence and entity categories under the guidance of a self-attention-based prompting module. It is designed specifically for NER tasks.
    \item  \textbf{KGPC}~\citep{chen2023few}: It injects the semantic relations of the knowledge graph to sequence to  text generation models to perform knowledge-guided instance generation for few-shot biomedical NER. It also only applies to NER tasks.
\end{itemize}

\blue{\textbf{LLM-based Generation Methods.} 
% In this study, we consider both zero-shot and few-shot learning data generation methods as baselines. 
% In zero-shot learning,
\begin{itemize}[leftmargin=0.5cm]
\item \textbf{ZeroGen}~\citep{ye2022zerogen}: It generates a dataset using simple class-conditional prompts and then trains a tiny task-specific model for zero-shot inference.  
We follow the prompting method mentioned in their original paper as implementation, which \emph{does not consider} any style information as well as domain knowledge.

\item \textbf{DemoGen} ~\citep{meng2023tuning,gpt3mix}: It leverages LLMs to synthesize novel training data by  feeding few-shot samples as demonstrations to guide the data generation process. 
Note that we focus on using the black-box LLM as the generator, thus we do not tune the LLM as \cite{meng2023tuning}.
\item \textbf{ProGen}~\citep{ye2022progen}: It first identifies the most important examples from the generated synthetic data using the influence function, then adds these examples as demonstrations to generate new training instances. To ensure fair comparison, we also add the few-shot demonstrations for data generation.
\item \textbf{S3}~\citep{wang2023lets}: It is a synthetic data generation method that iteratively extrapolates errors made by the classifier model trained on synthetic data leveraging a large language model. To adapt it in our few-shot setting, we use few-shot demonstrations $\cD$ as the validation set.  
\end{itemize}
}

% While there exist multiple approaches to clinical data generation~\citep{li2021synthetic,li-etal-2023-two,chintagunta2021medically}, it's important to highlight that these approaches predominantly adapt existing prompting techniques like ZeroGen and DemoGen to clinical downstream datasets rather than introducing novel data generation frameworks.  
% Besides, some works (e.g. \citep{yu2023large,chung-etal-2023-increasing}) rely on additional human intervention for data generation. 
% Therefore, we do not provide additional comparisons with these approaches.

% \yy{} We do not compare with \cite{tang2023does} in the main experiments as it leverages entities extracted from the entire training set and violates the true few-shot learning setting.
% The comparison is listed in Appendix~\ref{}.

\section{Prompt Format}
\label{sec:prompt_format}
\subsection{The prompts for Writing Styles Suggestion with {\ours}}
\label{sec: style_prompt}

\begin{lstlisting}[style=mystyle, caption={Prompt Format for writing styles suggestion with {\ours}.}, label=lst:prompt, escapeinside={<@}{@>}]
Suppose you need to generate a synthetic clinical text dataset on <@\textcolor{blue}{[task]}@> tasks. Here are a few examples from the original training set:
<@\textcolor{blue}{[demonstrations]}@>
Please write three potential sources, speakers or authors of the sentences.

\end{lstlisting}

\texttt{[task]}: The task names for each specific task.
\texttt{[demonstrations]}: The few-shot demonstrations from the original training set.

\subsection{The prompts for Data Generation with {\ours}}
\label{sec:generation_prompt}
In the following prompt format, \texttt{[topic]} and \texttt{[style]} are randomly sampled from the topics candidate set and styles candidate set we formulate in the knowledge extraction step, respectively.

\textbf{Named entity recognition tasks:}

\begin{lstlisting}[style=mystyle, caption={Prompt Format for NER tasks with {\ours}.}, label=lst:prompt, escapeinside={<@}{@>}]
Suppose you need to create a dataset for <@\textcolor{blue}{[domain]}@> recognition. Your task is to:
1. generate a sentence about <@\textcolor{blue}{[domain]}@>,
2. output a list of named entity about <@\textcolor{blue}{[domain]}@> only,
3. the sentence should mimic the style of <@\textcolor{blue}{[style]}@>,
4. the sentence should mention the <@\textcolor{blue}{[domain]}@> named <@\textcolor{blue}{[topic]}@>.
\end{lstlisting}

\texttt{[domain]}: ``disease" for BC5CDR-Disease and NCBI-Disease; ``chemical" for BC5CDR-Chemical and CHEMDNER.

\textbf{Medication attributes tasks:}

\begin{lstlisting}[style=mystyle, caption={Prompt Format for medication attributes tasks with {\ours}.}, label=lst:prompt, escapeinside={<@}{@>}]
Suppose you need to create a dataset for clinical attributes recognition. Your task is to:
1. generate a sentence about clinical attributes, The Clinical Attributes you need to extract include "Medication", "Dosage", "Route", "Frequency", "Reason", "Duration". For each attribute class, please return a list of attributes within the class that occurs in the Sentence.
2. the sentence should mimic the style of <@\textcolor{blue}{[style]}@>,
3. the sentence should be relevant to <@\textcolor{blue}{[topic]}@>.
\end{lstlisting}

\textbf{Text classification tasks:}

\begin{lstlisting}[style=mystyle, caption={Prompt Format for text classification tasks with {\ours}.}, label=lst:prompt, escapeinside={<@}{@>}]
 Suppose you need to create a dataset for <@\textcolor{blue}{[domain]}@>. Your task is to:
 1. generate a sentence about <@\textcolor{blue}{[domain]}@>. 
 2. the sentence should mimic the style of <@\textcolor{blue}{[style]}@>.
 3. the sentence should be relevant to the subtopic of <@\textcolor{blue}{[topic]}@> for <@\textcolor{blue}{[class\_name]}@>.
\end{lstlisting}
\texttt{[domain]}: ``COVID-19 Literature" for LitCovid and ``Cancer Document" for HOC.

\texttt{[class\_name]}: the label name for this generated sample, listed in Appendix~\ref{sec:dataset_description}.

\textbf{Relation extraction tasks:}

\begin{lstlisting}[style=mystyle, caption={Prompt Format for relation extraction tasks with {\ours}.}, label=lst:prompt, escapeinside={<@}{@>}]
Suppose you need to generate synthetic data for the biomedical <@\textcolor{blue}{[domain]}@> task. Your task is to:
1. give a sentence about <@\textcolor{blue}{[class\_name]}@> relation between <@\textcolor{blue}{[entity0]}@> and <@\textcolor{blue}{[entity1]}@>
2. the sentence should discuss the <@\textcolor{blue}{[entity0]}@>: <@\textcolor{blue}{[topic0]}@> and <@\textcolor{blue}{[entity1]}@>: <@\textcolor{blue}{[topic1]}@> with the relation <@\textcolor{blue}{[label\_desc]}@>.
3. the sentence should mimic the style of <@\textcolor{blue}{[style]}@>.
\end{lstlisting}
\texttt{[domain]}: ``Disease Gene Relation" for GAD, ``Chemical Disease Relation" for CDR, and ``Chemical Protein Relation" for ChemProt.

\texttt{[entity0]} and \texttt{[entity1]}: ``disease" and ``gene" for GAD, ``chemical" and ``disease: for CDR, and ``chemical" and ``protein" for ChemProt.

\texttt{[class\_name]}: the label name for this generated sample, listed in Appendix~\ref{sec:dataset_description}.

\texttt{[label\_desc]}: the description of the selected label. For example, the label ``upregulator" in ChemProt has a description of ``the chemical activates expression of the protein."

\textbf{Natural language inference tasks:}
\begin{lstlisting}[style=mystyle, caption={Prompt Format for generating the first sentence in NLI tasks with {\ours}.}, label=lst:prompt, escapeinside={<@}{@>}]
Suppose you need to create a set of <@\textcolor{blue}{[content]}@>. Your task is to:
1. generate one sentence for a <@\textcolor{blue}{[content]}@>.
2. the <@\textcolor{blue}{[content]}@> should be relevant to <@\textcolor{blue}{[topic]}@>,
3. The <@\textcolor{blue}{[content]}@> should mimic the style of <@\textcolor{blue}{[style]}@>.
\end{lstlisting}
\texttt{[content]}: ``health question" for MEDIQA-RQE, ``claim" for MEDIQA-NLI, MedNLI and MQP, and ``health news" for PUBHEALTH and HealthVer.

% \vspace{8ex}
\begin{lstlisting}[style=mystyle, caption={Prompt Format for generating the second sentence in NLI tasks with {\ours}.}, label=lst:prompt, escapeinside={<@}{@>}]
Suppose you need to create a pair of sentences for the <@\textcolor{blue}{[domain]}@> task with the label '<@\textcolor{blue}{[class\_name]}@>'. Given the <@\textcolor{blue}{[content]}@>: '<@\textcolor{blue}{[first\_sentence]}@>', Your task is to:
1. generate one short <@\textcolor{blue}{[content]}@> about <@\textcolor{blue}{[topic]}@> so that <@\textcolor{blue}{[label\_desc]}@>.
2. The <@\textcolor{blue}{[content]}@> should mimic the style of the first sentence.
\end{lstlisting}
\texttt{[domain]}: ``Question Entailment" for MEDIQA-RQE, ``Natural Language Entailment" for MEDIQA-NLI and MedNLI, ``Fact Verification" for PUBHEALTH and HealthVer, and ``Sentence Similarity Calculation" for MQP.

\texttt{[content]}: ``health question" for MEDIQA-RQE, ``hypothesis" for MEDIQA-NLI, MedNLI, ``evidence" for PUBHEALTH and HealthVer, and ``sentence" for MQP.

\texttt{[class\_name]}: the label name for this generated sample, listed in Appendix~\ref{sec:dataset_description}.

\texttt{[label\_desc]}: the description of the selected label. 
For "entailment", the description is "we can infer the \texttt{[content]} from the given sentence".
For "neutral", the description is "there is no clear relation between the \texttt{[content]} from the given sentence".
For "contradict", the description is "we can refute the \texttt{[content]} from the given sentence".

\texttt{[first\_sentence]}: the first sentence we generate

\subsection{Prompts for ZeroGen, DemoGen, ProGen}
\label{sec:prompt_format_bsl}
We use the same set of prompts for ZeroGen, DemoGen and ProGen, while DemoGen and ProGen have additional demonstrations augmented to the prompts. DemoGen uses the few-shot examples in the training set as demonstrations, and ProGen leverages feedbacks from previous rounds to iteratively guide the generation.

\textbf{Named entity recognition tasks:}

\begin{lstlisting}[style=mystyle, caption={Prompt Format for NER tasks with baselines.}, label=lst:prompt, escapeinside={<@}{@>}]
Suppose you need to create a dataset for <@\textcolor{blue}{[domain]}@> recognition. Your task is to generate a sentence about <@\textcolor{blue}{[domain]}@> and output a list of named entity about <@\textcolor{blue}{[domain]}@> only.
\end{lstlisting}

\texttt{[domain]}: ``disease" for BC5CDR-Disease and NCBI-Disease; ``chemical" for BC5CDR-Chemical and CHEMDNER.

\textbf{Medication attributes tasks:}

\begin{lstlisting}[style=mystyle, caption={Prompt Format for medication attributes tasks with baselines.}, label=lst:prompt, escapeinside={<@}{@>}]
Suppose you need to create a dataset for clinical attributes recognition. Your task is to generate a sentence about clinical attributes, The Clinical Attributes you need to extract include "Medication", "Dosage", "Route", "Frequency", "Reason", "Duration". For each attribute class, please return a list of attributes within the class that occurs in the Sentence.
\end{lstlisting}

\textbf{Text classification tasks:}

\begin{lstlisting}[style=mystyle, caption={Prompt Format for text classification tasks with baselines.}, label=lst:prompt, escapeinside={<@}{@>}]
 Suppose you are a writer for <@\textcolor{blue}{[domain]}@>. Your task is to give a synthetic <@\textcolor{blue}{[domain]}@> about <@\textcolor{blue}{[class\_name]}@>. 
\end{lstlisting}
\texttt{[domain]}: ``COVID-19 Literature" for LitCovid and ``Cancer Document" for HOC.

\texttt{[class\_name]}: the label name for this generated sample, listed in Appendix~\ref{sec:dataset_description}.

\textbf{Relation extraction tasks:}

\begin{lstlisting}[style=mystyle, caption={Prompt Format for relation extraction tasks with baselines.}, label=lst:prompt, escapeinside={<@}{@>}]
Suppose you need to generate synthetic data for the biomedical <@\textcolor{blue}{[domain]}@> task. Your task is to give a sentence about <@\textcolor{blue}{[class\_name]}@> relation between <@\textcolor{blue}{[entity0]}@> and <@\textcolor{blue}{[entity1]}@> so that <@\textcolor{blue}{[label\_desc]}@>.
\end{lstlisting}
\texttt{[domain]}: ``Disease Gene Relation" for GAD, ``Chemical Disease Relation" for CDR, and ``Chemical Protein Relation" for ChemProt.

\texttt{[entity0]} and \texttt{[entity1]}: ``disease" and ``gene" for GAD, ``chemical" and ``disease: for CDR, and ``chemical" and ``protein" for ChemProt.

\texttt{[class\_name]}: the label name for this generated sample, listed in Appendix~\ref{sec:dataset_description}.

\texttt{[label\_desc]}: the description of the selected label. For example, the label ``upregulator" in ChemProt has a description of ``the chemical activates expression of the protein."

\textbf{Natural language inference tasks:}
\begin{lstlisting}[style=mystyle, caption={Prompt Format for generating the first sentence in NLI tasks with baselines.}, label=lst:prompt, escapeinside={<@}{@>}]
Suppose you need to create a set of <@\textcolor{blue}{[content]}@>. Your task is to generate one sentence for a <@\textcolor{blue}{[content]}@>.
\end{lstlisting}
\texttt{[content]}: ``health question" for MEDIQA-RQE, ``claim" for MEDIQA-NLI, MedNLI and MQP, and ``health news" for PUBHEALTH and HealthVer.

% \vspace{8ex}
\begin{lstlisting}[style=mystyle, caption={Prompt Format for generating the second sentence in NLI tasks with baselines.}, label=lst:prompt, escapeinside={<@}{@>}]
Suppose you need to create a pair of sentences for the <@\textcolor{blue}{[domain]}@> task with the label '<@\textcolor{blue}{[class\_name]}@>'. Given the <@\textcolor{blue}{[content]}@>: '<@\textcolor{blue}{[first\_sentence]}@>', Your task is to generate one short <@\textcolor{blue}{[content]}@> so that <@\textcolor{blue}{[label\_desc]}@>.
\end{lstlisting}
\texttt{[domain]}: ``Question Entailment" for MEDIQA-RQE, ``Natural Language Entailment" for MEDIQA-NLI and MedNLI, ``Fact Verification" for PUBHEALTH and HealthVer, and ``Sentence Similarity Calculation" for MQP.

\texttt{[content]}: ``health question" for MEDIQA-RQE, ``hypothesis" for MEDIQA-NLI, MedNLI, ``evidence" for PUBHEALTH and HealthVer, and ``sentence" for MQP.

\texttt{[class\_name]}: the label name for this generated sample, listed in Appendix~\ref{sec:dataset_description}.

\texttt{[label\_desc]}: the description of the selected label. 
For "entailment", the description is "we can infer the \texttt{[content]} from the given sentence".
For "neutral", the description is "there is no clear relation between the \texttt{[content]} from the given sentence".
For "contradict", the description is "we can refute the \texttt{[content]} from the given sentence".

\texttt{[first\_sentence]}: the first sentence we generate. 

% % \clearpage
% \section{Example of Generated Clinical Topics and Writing Styles}
% \label{sec:generated_details}

% \ran{}

\section{Detailed Per-task Experimental Results}
\label{sec:more_experimental_results}
In this section, we present additional experimental results on every dataset in Tables~\ref{tab:single-sent}, ~\ref{tab:sent-pair}, ~\ref{tab:token-class}. \blue{We also include the experimental results combining topic from both KG and LLM, which yields a performance improvement, though not a substantial one. However, note that in practice, it is challenging to tune the ratio in the few-shot setting.
}

\begin{table*}[t]
% \floatconts
  \caption{Performance on single-sentence tasks evaluated by PubMedBERT$_{\texttt{Base}}$ and PubMedBERT$_{\texttt{Large}}$. \textbf{Bold} and \underline{underline} indicate the best and second best results for each dataset, respectively. \blue{Note that the performance of `Supervised-Full (SOTA)' is copied from the existing paper. If the value in this field is missing, this means we cannot find reported results with the same-scale model on that dataset.} (Same as below).}
  \resizebox{\linewidth}{!}{
  \begin{tabular}{lccccccccc}
  \toprule
  & \bfseries LitCovid & \bfseries HOC & \multicolumn{3}{c}{\textbf{CDR}} & \multicolumn{3}{c}{\textbf{GAD}} & \bfseries ChemProt \\
  % \midrule
  \cmidrule(lr){2-2} \cmidrule(lr){3-3} \cmidrule(lr){4-6} \cmidrule(lr){7-9} \cmidrule(lr){10-10}
  & F1 & F1 & P & R & F1 & P & R & F1 & F1\\
  \midrule
  \multicolumn{10}{l}{\textbf{PubMedBERT$_{\texttt{Base}}$}} \\
  \midrule
 \blue{Supervised-Full (SOTA)} & 73.55 & 84.35 &  67.81 & 76.60 & 71.96 & ---  & ---  & 84.39 & 77.97 \\
  Supervised-Full & 71.70 & 82.32 & 67.81 & 76.60 & 71.96 & 82.55 & 85.10 & 83.81 & 76.24 \\
  Supervised-Few & 24.08 & 13.13 & 41.62 & 52.96 & 46.61 & 57.71 & 46.54 & 51.53 & 33.54 \\
  \midrule
  DA-Word Sub & 36.49 & 44.98 & 40.50 & 46.20 & 43.16 & 51.15 & 32.10 & 39.45 & 31.82 \\
  DA-Back Trans & 39.70 & 54.78 & --- & --- & --- & --- & --- & --- & --- \\
  DA-Mixup & 40.82 & 49.35 & 41.40 & 44.80 & 43.03 & 55.44 & 48.30 & 51.62 & 35.45 \\
  DA-Transformer & 39.86 & 42.18 & 44.60 & 61.70 & 51.77 & 59.40 & 46.50 & 52.16 & 38.73 \\ \midrule
  ZeroGen & 50.50 & 67.90 & 38.82 & \textbf{91.82} & 54.57 & 84.38 & 80.68 & 82.49 & 54.46 \\
  DemoGen & 57.65 & 70.52 & 46.90 & \underline{83.3} & 60.01 & 93.14 & 80.19 & 86.18 & 56.18 \\
  ProGen & 58.06 & 72.25 & 51.35 & {71.58} & 59.80 & 90.52 & \textbf{85.14} & \underline{87.75} & 54.15 \\
  S3 & \underline{58.67} & 71.58 & 49.76 &	76.08 & 	60.17 & 94.85 &	80.19 &	86.90 & 55.75 \\
  \midrule
  % \hline
  \rowcolor{teal!10} {\ours} w/ KG & 58.01 & \underline{76.28} & \underline{56.98} & {67.38} & \underline{61.75} & \underline{93.33} & \underline{83.68} & \textbf{88.24} & \underline{57.04} \\
  \rowcolor{teal!10} {\ours} w/ LLM & \textbf{59.22} & \textbf{76.42} & \textbf{60.60} & 66.35 & \textbf{63.34} & \textbf{94.61} & 78.17 & 85.61 & \textbf{61.22} \\
  \midrule
  \rowcolor{teal!10} \blue{{\ours} w/ KG+LLM} & 56.56 & 78.02 & 57.97 & 71.09 & 63.86 & 92.57 & 88.59 & 90.54 & 58.48 \\
  \midrule
  \multicolumn{10}{l}{\textbf{PubMedBERT$_{\texttt{Large}}$}} \\
  \midrule
  \blue{Supervised-Full (SOTA)} & ---  & 84.87 & ---  & --- & --- & ---  & ---  & 84.90 & 78.77 \\
  Supervised-Full & 74.59 & 85.53 & 72.31 & 74.88 & 73.57 & 84.95 & 88.75 & 86.81 & 78.55 \\
  Supervised-Few & 22.59 & 13.13 & 42.27 & 67.51 & 51.99 & 57.58 & 90.07 & 70.25 & 35.80 \\
  \midrule
  DA-Word Sub & 37.20 & 50.78 & 47.70 & 43.50 & 45.50 & 63.40 & 42.00 & 50.53 & 37.01 \\
  DA-Back Trans & 40.50 & 61.46 & --- & --- & --- & --- & --- & --- & --- \\
  DA-Mixup & 40.03 & 53.45 & 43.34 & 73.50 & 54.53 & 62.20 & 59.93 & 60.52 & 37.87 \\
  DA-Transformer & 38.95 & 49.86 & 50.70 & 31.60 & 38.93 & 59.80 & 57.76 & 58.76 & 40.66 \\
  \midrule
  ZeroGen & 52.86 & 70.16 & 42.95 & \textbf{80.67} & 56.06 & 92.26 & 76.73 & 83.78 & 55.71 \\
  DemoGen & \underline{56.29} & 73.65 & 50.86 & 74.30 & 60.39 & \textbf{96.85} & 76.83 & 85.69 & 59.88 \\
  ProGen & 54.71 & 75.31 & 50.36 & \underline{76.08} & 60.60 & 91.11 & 85.63 & 88.29 & 58.79 \\
  S3 & 53.56	&75.11 & 51.51 &	78.30 &	62.14 &	92.12 &	83.80 &	87.76 & 	59.05 \\
  \midrule
  \rowcolor{teal!10} {\ours} w/ KG & 55.81 & \underline{77.71} & \underline{60.45} & 65.04 & \underline{62.66} & 94.30 & \textbf{89.08} & \textbf{91.62} & \underline{60.12} \\
  \rowcolor{teal!10} {\ours} w/ LLM & \textbf{57.07} & \textbf{78.14} & \textbf{67.13} & 62.98 & \textbf{64.99} & \underline{95.08} & \underline{86.14} & \underline{90.39} & \textbf{63.05} \\
  \midrule
  \rowcolor{teal!10} \blue{{\ours} w/ KG+LLM} & 56.80 & 79.07 & 64.19 & 67.70 & 65.90 & 92.41 & 92.07 & 92.24 & 59.95 \\
  \bottomrule
  \end{tabular}
  }
  \label{tab:single-sent}
\end{table*}
\clearpage
\begin{table*}[t]
% \floatconts
  \caption{Performance on sentence-pair tasks evaluated by PubMedBERT$_{\texttt{Base}}$ and PubMedBERT$_{\texttt{Large}}$.}
  \resizebox{\linewidth}{!}{
  \begin{tabular}{lcccccccccc}
  \toprule
  & \bfseries MEDIQA-RQE & \bfseries MEDIQA-NLI & \bfseries MedNLI & \multicolumn{2}{c}{\textbf{PUBHEALTH}} & \multicolumn{2}{c}{\textbf{HealthVer}} & \bfseries MQP & \bfseries PubmedQA & \bfseries BioASQ \\
  % \midrule
  \cmidrule(lr){2-2} \cmidrule(lr){3-3} \cmidrule(lr){4-4} \cmidrule(lr){5-6} \cmidrule(lr){7-8} \cmidrule(lr){9-9}
  \cmidrule(lr){10-10}
  \cmidrule(lr){11-11}
  & ACC & ACC & ACC & ACC & F1 & ACC & F1 & ACC& ACC & ACC \\
  \midrule
  \multicolumn{9}{l}{\textbf{PubMedBERT$_{\texttt{Base}}$}} \\
  \midrule
  \blue{Supervised-Full (SOTA)} & --- & --- & 86.60 & 70.52 & 69.73 & 73.54 & 74.82 & 79.20 & 70.20 & 91.43\\
  Supervised-Full & 77.15 & 79.01 & 81.43 & 65.16 & 62.96 & 70.00 & 68.02 & 75.70 & 61.84 & 87.56 \\
  Supervised-Few & 57.51 & 40.00 & 36.40 & 28.30 & 23.70 & 30.55 & 30.49 & 55.70 & 55.90 & 53.57  \\
  \midrule
  DA-Word Sub & 58.60 & 50.24 & 56.40 & 23.67 & 17.64 & 34.05 & 34.02 & 54.40 & 52.88 & 54.28\\
  DA-Back Trans & 59.16 & 49.92 & 53.82 & 30.70 & 23.32 & 33.60 & 32.76 & 55.80 & 53.70 & 52.86\\
  DA-Mixup & 57.71 & 49.38 & 53.47 & 31.45 & 24.45 & 34.11 & 33.78 & 58.20 & 51.68 & 52.14 \\
DA-Transformer & 62.25 &	51.19 &	 53.70 & 34.81	& 27.75	 &35.83	 & 35.78 & 58.80 & 54.14 & 58.57\\
  \midrule
  ZeroGen & 63.28 & 52.89 & 57.71 & 35.80 & 31.50 & 34.80 & 33.50 & 68.35 & 55.20 & 	68.57 \\
  DemoGen & 66.56 & 56.29 & 58.56 & 42.60 & 35.40 & 38.00 & 36.50 & 70.85 &57.60	&66.42 \\
  ProGen & 65.94 & 57.28 & 59.49 & 38.70 & 33.10 & 36.72 & 35.97 & 69.30  &57.90	 &63.57 \\
  S3 & 66.02 & 58.30 &	59.75 & 42.40 &	34.90 &	37.94 &	37.97 & 70.20 & 58.60	& 68.57\\
  \midrule
  \rowcolor{teal!10} {\ours} w/ KG & \textbf{74.85} & \underline{58.03} & \underline{61.80} & \underline{44.60} & \underline{36.80} & \underline{43.05} & \underline{42.06} & \underline{72.20} & \bf 65.80	& \underline{77.14}\\
  \rowcolor{teal!10} {\ours} w/ LLM & \underline{72.40} & \textbf{64.44} & \textbf{64.89} & \textbf{48.50} & \textbf{40.60} & \textbf{44.50} & \textbf{42.32} & \textbf{73.30} & \underline{61.30} &\bf 77.85 \\
  \midrule
  \rowcolor{teal!10} \blue{{\ours} w/ KG+LLM} & 75.10 & 64.12 & 65.81 & 50.57 & 40.65 & 40.60 & 39.59 & 68.30 &66.70 &	77.85 \\
  \midrule
  \multicolumn{9}{l}{\textbf{PubMedBERT$_{\texttt{Large}}$}} \\
  \midrule
  \blue{Supervised-Full (SOTA)} & --- & --- & 86.57 & --- & --- & --- & --- & 81.00 & 72.18 & 94.82 \\
  Supervised-Full & 81.10 & 82.89 & 83.96 & 70.21 & 63.45 & 75.72 & 75.01 & 78.80 & 67.38 & 93.36 \\
  Supervised-Few & 63.79 & 47.40 & 38.80 & 46.20 & 27.20 & 35.60 & 33.80 & 59.73 & 60.44 & 58.57 \\
  \midrule
  DA-Word Sub & 64.26 & 51.20 & 57.53 & 35.60 & 31.60 & 35.41 & 32.29 & 55.30 & 55.72& 61.42\\
  DA-Back Trans & 65.52 & 51.43 & 58.21 & 34.45 & 30.50 & 33.78 & 32.21 & 56.40 & 54.38	& 60.00\\
  DA-Mixup & 64.10 & 50.91 & 57.03 & 34.23 & 30.78 & 33.79 & 31.42 & 58.50 & 54.80 & 	58.57\\
  DA-Transformer & 68.97	& 51.05	 & 56.79 & 
  38.46 &	31.40 &	31.72 &	30.50 & 58.10 & 58.60 &	60.00
  \\
  \midrule
  ZeroGen & 67.26 & 60.74 & 62.42 & 42.50 & 33.30 & 39.74 & 38.90 & 72.69 & 57.75 & 	74.28\\
  DemoGen & 69.22 & 62.97 & 64.55 & 44.50 & 36.80 & 40.72 & 40.57 & 74.37 &  61.50	& 68.57\\
  ProGen & 67.82 & 60.98 & 63.15 & 44.15 & 36.37 & 41.42 & 40.89 & 74.90 & 59.40	& 67.14\\
  S3 & 67.98&	63.15&	64.10& 43.72 &	35.67 &	39.80 &	39.78&	 73.20 & 61.20 & 	71.42\\
  \midrule
  \rowcolor{teal!10} {\ours} w/ KG & \textbf{79.92} & \underline{63.59} & \underline{69.19} & \underline{50.20} & \underline{41.26} & \textbf{47.03} & \underline{43.64} & \underline{75.40} & \bf 68.60 &	\underline{79.28}\\
  \rowcolor{teal!10} {\ours} w/ LLM & \underline{77.36} & \textbf{64.69} & \textbf{69.46} & \textbf{52.96} & \textbf{43.31} & \underline{46.05} & \textbf{44.12} & \textbf{76.20} & \underline{66.80} &\bf	80.00 \\
  \midrule
  \rowcolor{teal!10} \blue{{\ours} w/ KG+LLM} & 80.77 & 63.30 & 70.56 & 51.98 & 41.61 & 47.44 & 44.25 & 71.90 & 67.40 &	79.28\\
  \bottomrule
  \end{tabular}
  }
  \label{tab:sent-pair}
\end{table*}
\clearpage
\begin{table*}[t]
% \floatconts
  \caption{Performance on token-classification tasks evaluated by PubMedBERT$_{\texttt{Base}}$ and PubMedBERT$_{\texttt{Large}}$.}
  \resizebox{\linewidth}{!}{
  \begin{tabular}{lccccccccccccccc}
  \toprule
  & \multicolumn{3}{c}{\textbf{BC5CDR-Disease}} & \multicolumn{3}{c}{\textbf{BC5CDR-Chemical}} & \multicolumn{3}{c}{\textbf{NCBI-Disease}} & \multicolumn{3}{c}{\textbf{CHEMDNER}} & \multicolumn{3}{c}{\textbf{CASI}} \\
  % \midrule
  \cmidrule(lr){2-4} \cmidrule(lr){5-7} \cmidrule(lr){8-10} \cmidrule(lr){11-13} \cmidrule(lr){14-16}
  & P & R & F1 & P & R & F1 & P & R & F1 & P & R & F1 & P & R & F1\\
  \midrule
  \multicolumn{16}{l}{\textbf{PubMedBERT$_{\texttt{Base}}$}} \\
  \midrule
  \blue{Supervised-Full (SOTA)} & --- & --- & 86.10 & --- & --- &  93.33 & --- & --- & 88.76 & --- & ---  &92.35 & ---  & --- & ---  \\
  Supervised-Full & 83.84 & 87.92 & 85.83 & 92.22 & 91.74 & 91.98 & 87.54 & 89.92 & 88.71 & 91.84 & 92.45 & 92.14 & --- & --- & --- \\
  Supervised-Few & 24.86 & 39.47 & 30.51 & 63.73 & 46.07 & 53.48 & 36.16 & 39.47 & 37.74 & 48.00 & 28.70 & 35.92 & 38.11 & 43.82 & 40.77 \\
  \midrule
  DA-Word Sub & 35.34 & 39.54 & 37.32 & 63.13 & 52.52 & 57.34 & 53.40 & 36.70 & 43.50 & 47.45 & 33.15 & 39.03 & 40.25 & 47.65 & 43.64 \\
  DA-Mixup & 36.13 & 42.90 & 39.23 & 66.43 & 50.54 & 57.41 & 56.57 & 26.48 & 36.07 & 52.40 & 27.53 & 36.10 & 42.37 & 48.96 & 45.43 \\
  LightNER & 39.80 & 33.20 & 36.20 & --- & ---  & --- & 43.70 & 41.90 & 42.78 & --- & --- & --- & --- & --- & --- \\						
  DA-MELM & 34.20 & 41.30 & 37.42 & 47.23 & 72.81 & 57.29 & 36.90 & 48.50 & 41.91 & 39.33 & 45.95 & 42.38 & 37.82 & 44.28 & 40.80 \\
  KGPC & 50.80 & 51.30 & 51.05 & --- & --- & --- & 52.20 & 52.10 & 52.15 & --- & --- & --- & --- & --- & --- \\
  \midrule
  ZeroGen & 55.60 & 39.10 & 45.91 & 73.20 & 82.85 & 77.73 & 56.25 & 45.98 & 50.60 & \textbf{54.34} & 52.93 & 53.63 & 52.80 & 49.53 & 51.11 \\
  DemoGen & \underline{63.10} & 48.44 & 54.81 & 76.40 & 81.65 & 78.94 & 57.65 & 49.08 & 53.02 & \underline{54.00} & 53.77 & 53.88 & 58.15 & 56.84 & 57.49 \\
  ProGen & 61.60 & 50.50 & 55.50 & \underline{77.10} & 82.02 & 79.48 & 56.01 & \underline{53.50} & 54.73 & 51.55 & 53.00 & 52.26 & 57.76 & 58.57 & 58.16 \\
  S3 & 58.26	&55.96	&57.08&	77.28	&80.80&	79.00&56.39	&49.34&	52.62& 48.53&	57.79	&52.75 &56.21&	63.60&	59.68 \\
  \midrule
  \rowcolor{teal!10} {\ours} w/ KG & 58.64 & \textbf{63.02} & \underline{60.75} & 74.96 & \textbf{85.45} & \underline{79.86} & \textbf{62.62} & \textbf{56.62} & \textbf{59.47} & 48.33 & \textbf{69.28} & \textbf{56.94} & \textbf{71.75} & \underline{65.20} & \textbf{68.32} \\
  \rowcolor{teal!10} {\ours} w/ LLM & \textbf{63.41} & \underline{58.83} & \textbf{61.03} & \textbf{77.68} & \underline{84.33} & \textbf{80.87} & \underline{62.58} & 50.59 & \underline{55.95} & 51.40 & \underline{58.77} & \underline{54.84} & \underline{68.19} & \textbf{66.79} & \underline{67.48} \\
  \midrule
  \rowcolor{teal!10} \blue{{\ours} w/ KG+LLM} & 60.57 & 66.21 & 63.26 & 73.66 & 87.30 & 79.90 & 58.01 & 65.37 & 59.17 & 52.07 & 63.62 & 57.27 & 72.57 & 70.48 & 71.51 \\
  \midrule
  \multicolumn{16}{l}{\textbf{PubMedBERT$_{\texttt{Large}}$}} \\
  \midrule
\blue{Supervised-Full (SOTA)} & --- & --- & 86.39 & --- & --- &  94.04 & --- & --- & 89.18 & --- & ---  &92.72 & ---  & --- & ---  \\

  Supervised-Full & 86.77 & 85.92 & 86.34 & 92.80 & 92.94 & 92.87 & 87.97 & 90.09 & 89.02 & 92.23 & 92.48 & 92.35 & --- & --- & --- \\
  Supervised-Few & 25.52 & 45.85 & 32.79 & 61.40 & 54.41 & 57.69 & 44.86 & 40.12 & 42.35 & 43.40 & 34.60 & 38.50 & 41.30 & 45.02 & 43.08 \\
  \midrule
  DA-Word Sub & 38.54 & 38.85 & 38.69 & 64.85 & 53.96 & 58.91 & 52.59 & 45.35 & 48.70 & 44.85 & 36.69 & 40.36 & 46.77 & 43.52 & 45.09 \\
  DA-Mixup  & 36.27 & 46.67 & 40.82 & 67.63 & 54.15 & 60.14 & 55.64 & 38.06 & 45.20 & 45.51 & 36.66 & 40.61 & 41.25 & 52.09 & 46.04 \\
  LightNER & --- & --- & --- & --- & --- & --- & --- & --- & --- & --- & --- & --- & --- & --- & --- \\
  DA-MELM  & 33.40 & 41.61 & 37.06 & 53.80 & 66.71 & 59.56 & 44.20 & 57.40 & 49.94 & 36.40 & 47.41 & 41.18 & 43.36 & 45.78 & 44.54 \\
  KGPC & --- & --- & --- & --- & --- & --- & --- & --- & --- & --- & --- & --- & --- & --- & --- \\
  \midrule
  ZeroGen & 57.40 & 39.21 & 46.59 & 78.08 & 80.97 & 79.49 & 54.52 & 49.00 & 51.61 & 48.56 & 59.44 & 53.45 & 54.04 & 51.40 & 52.69 \\
  DemoGen & 57.34 & 49.48 & 53.12 & \underline{78.27} & 83.90 & 80.99 & 59.43 & 56.83 & 58.10 & 48.03 & 60.39 & 53.51 & 62.67 & 61.02 & 61.83 \\
  ProGen & \underline{60.34} & 54.13 & 57.07 & \textbf{78.42} & 82.94 & 80.62 & 60.02 & 55.28 & 57.55 & \underline{50.40} & 59.64 & 54.63 & 57.21 & 63.70 & 60.28 \\
  S3 & 65.46 &	51.86 &	57.87 & 77.89 &	84.31 &	80.97 &	56.00 &	53.49 &	54.72	 & 54.80 &	53.88 & 54.33 & 63.07 &	62.72 &	62.89 \\
  \midrule
  \rowcolor{teal!10} {\ours} w/ KG & 54.28 & \textbf{70.14} & \underline{61.21} & 77.88 & \underline{86.32} & \underline{81.88} & \textbf{62.46} & \textbf{64.08} & \textbf{63.26} & 47.03 & \textbf{67.86} & \textbf{55.56} & \underline{70.96} & \textbf{69.66} & \textbf{70.30} \\
  \rowcolor{teal!10} {\ours} w/ LLM & \textbf{61.05} & \underline{65.40} & \textbf{63.15} & 78.08 & \textbf{86.98} & \textbf{82.29} & \underline{61.12} & \underline{60.16} & \underline{60.64} & \textbf{50.92} & \underline{60.67} & \underline{55.37} & \textbf{71.61} & \underline{66.86} & \underline{69.15} \\
  \midrule
  \rowcolor{teal!10} \blue{{\ours} w/ KG+LLM} & 65.67 & 66.22 & 65.94 & 75.89 & 87.61 & 81.33 & 65.70 & 59.22 & 62.31 & 52.49 & 65.07 & 58.11 & 73.21 & 69.30 & 71.20\\
  \bottomrule
  \end{tabular}
  }
  \label{tab:token-class}
\end{table*}

\begin{figure*}[t!]
    \centering
    \begin{minipage}{0.48\textwidth}
        \centering
        \subfigure[CDR]{
            \includegraphics[width=0.48\textwidth]{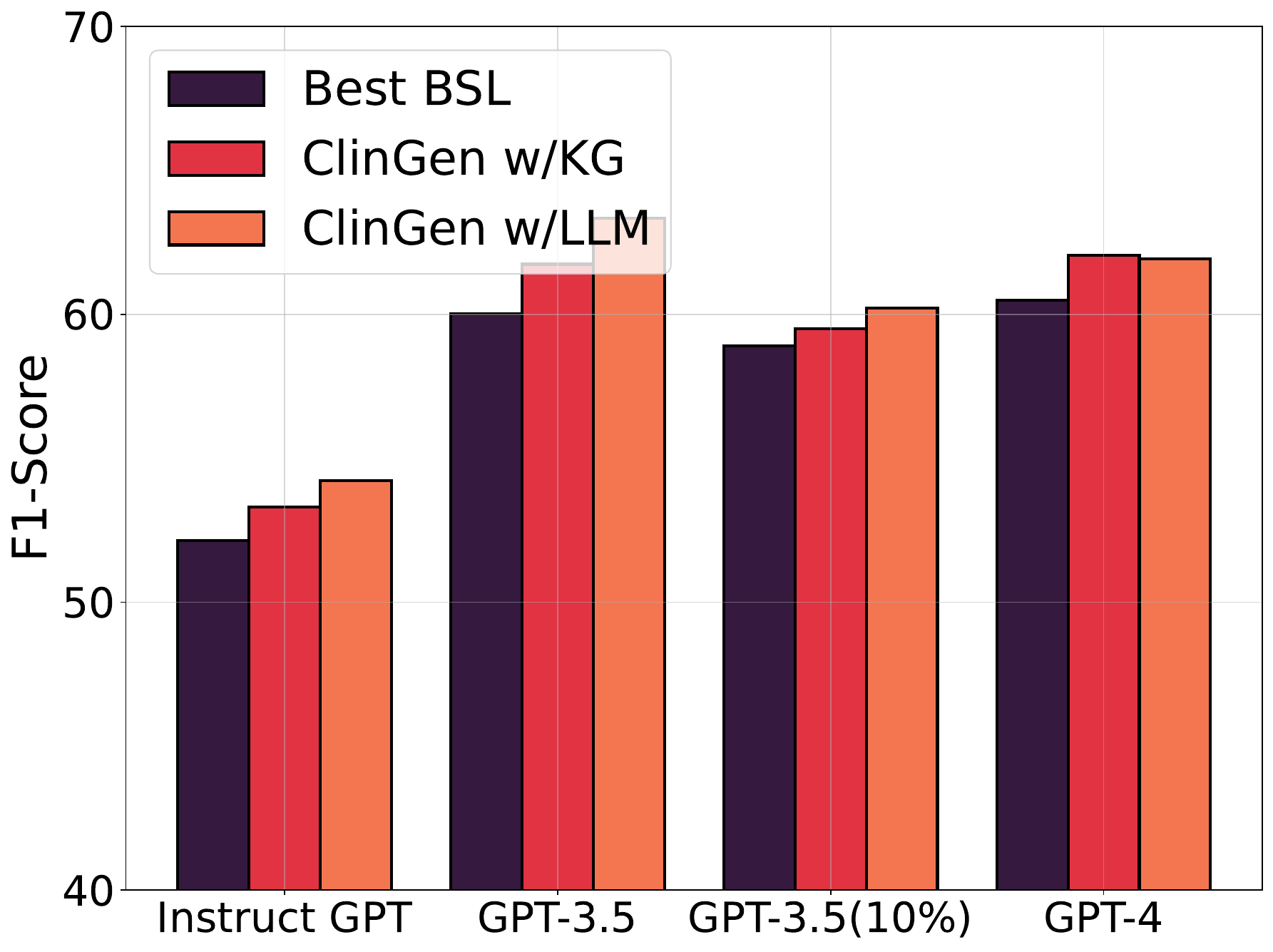}
            \label{fig:generator-CDR}
        } \hspace{-3mm}
        \subfigure[NCBI-Disease]{
            \includegraphics[width=0.48\textwidth]{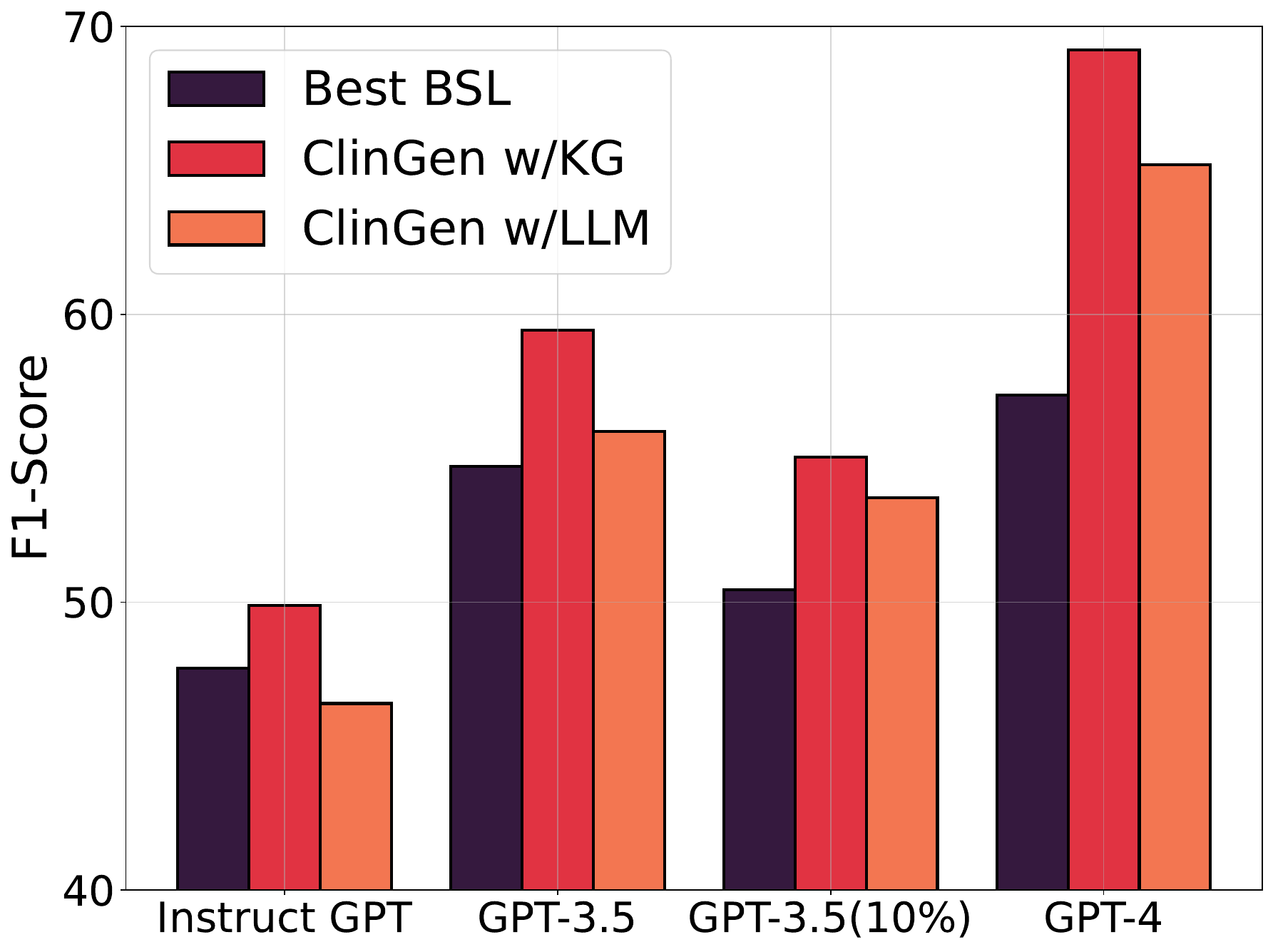}
            \label{fig:generator-NCBI}
        }
        \vspace{-2ex}
        \RawCaption{\caption{Different generators at \texttt{Base}.}\label{fig:generator-add}}
    \end{minipage}%
    \begin{minipage}{0.48\textwidth}
        \centering
        \subfigure[CDR]{
            \includegraphics[width=0.5\textwidth]{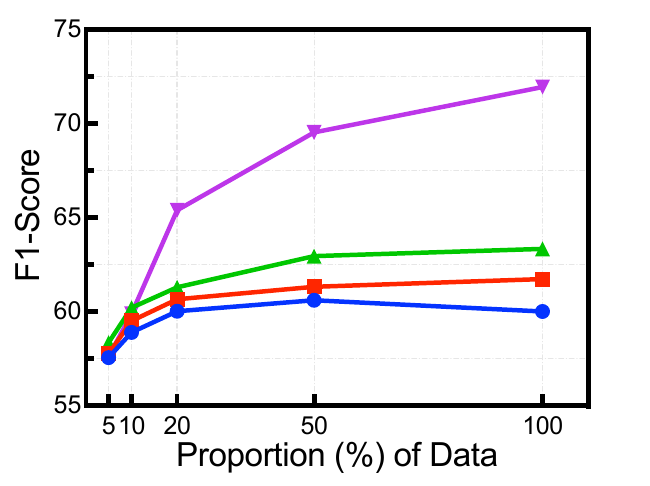}
            \label{fig:size-CDR}
        } \hspace{-6mm}
        \subfigure[NCBI-Disease]{
            \includegraphics[width=0.5\textwidth]{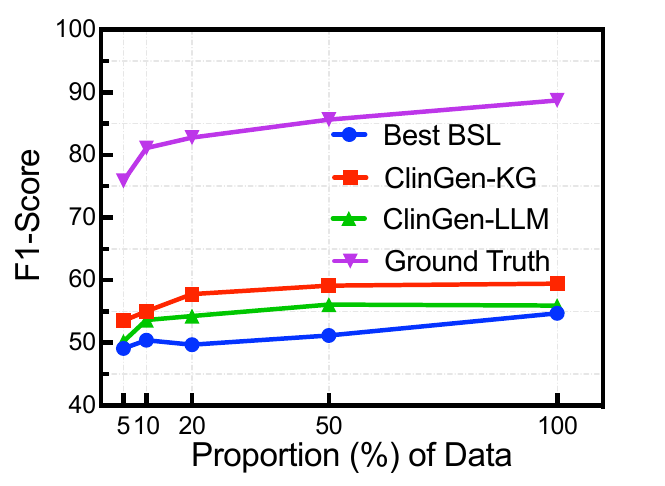}
            \label{fig:size-NCBI}
        }
        \vspace{-2ex}
        \RawCaption{\caption{Different proportion of data at \texttt{Base}.}\label{fig:size-synthetic-add}}
    \end{minipage}%
    \vspace{-0.5ex}
\end{figure*}
% \clearpage
\begin{table*}[t]
% \floatconts
% \vspace{-1ex}
  \caption{Performance with Different Random Seeds using PubMedBERT$_{\texttt{Base}}$.}
  \resizebox{\linewidth}{!}{
  \begin{tabular}{l ccc | ccc | ccc| ccc}
  \toprule
  & \multicolumn{3}{c}{\textbf{HOC}} & \multicolumn{3}{c}{\textbf{CDR}} & \multicolumn{3}{c}{\textbf{MEDIQA-RQE}} & \multicolumn{3}{c}{\textbf{NCBI-Disease}}\\
  % \midrule
  \cmidrule(lr){2-4} \cmidrule(lr){5-7} \cmidrule(lr){8-10} \cmidrule(lr){11-13}
  & Best Baseline & {\ours}-KG& {\ours}-LLM & Best Baseline & {\ours}-KG& {\ours}-LLM& Best Baseline & {\ours}-KG& {\ours}-LLM & Best Baseline & {\ours}-KG& {\ours}-LLM\\
  \midrule
  1 &70.04 &	74.30 &	77.30	 &61.52 &	61.66 &	63.34	 & 68.30	 &76.85 &	74.50 &	56.12	 &60.22	 &54.51 \\
2 & 75.30	& 79.73& 	73.63	& 60.69& 	63.77	& 64.66	& 64.20	& 71.80	& 71.19	& 54.19& 	60.64& 	57.81 \\
3 & 71.41& 	74.81	& 78.33& 	57.82& 	59.79	& 62.02& 	67.18	& 75.90	& 71.51	& 53.85& 	57.52& 	55.50\\
  \bottomrule
  \end{tabular}
  }
  \label{tab:random_seed}
  \vspace{-1ex}
\end{table*}
\clearpage

 \begin{figure*}[!t]
	\centering
	% \vspace{-2ex}
	\subfigure[LitCovid]{
		\includegraphics[width=0.31\linewidth]{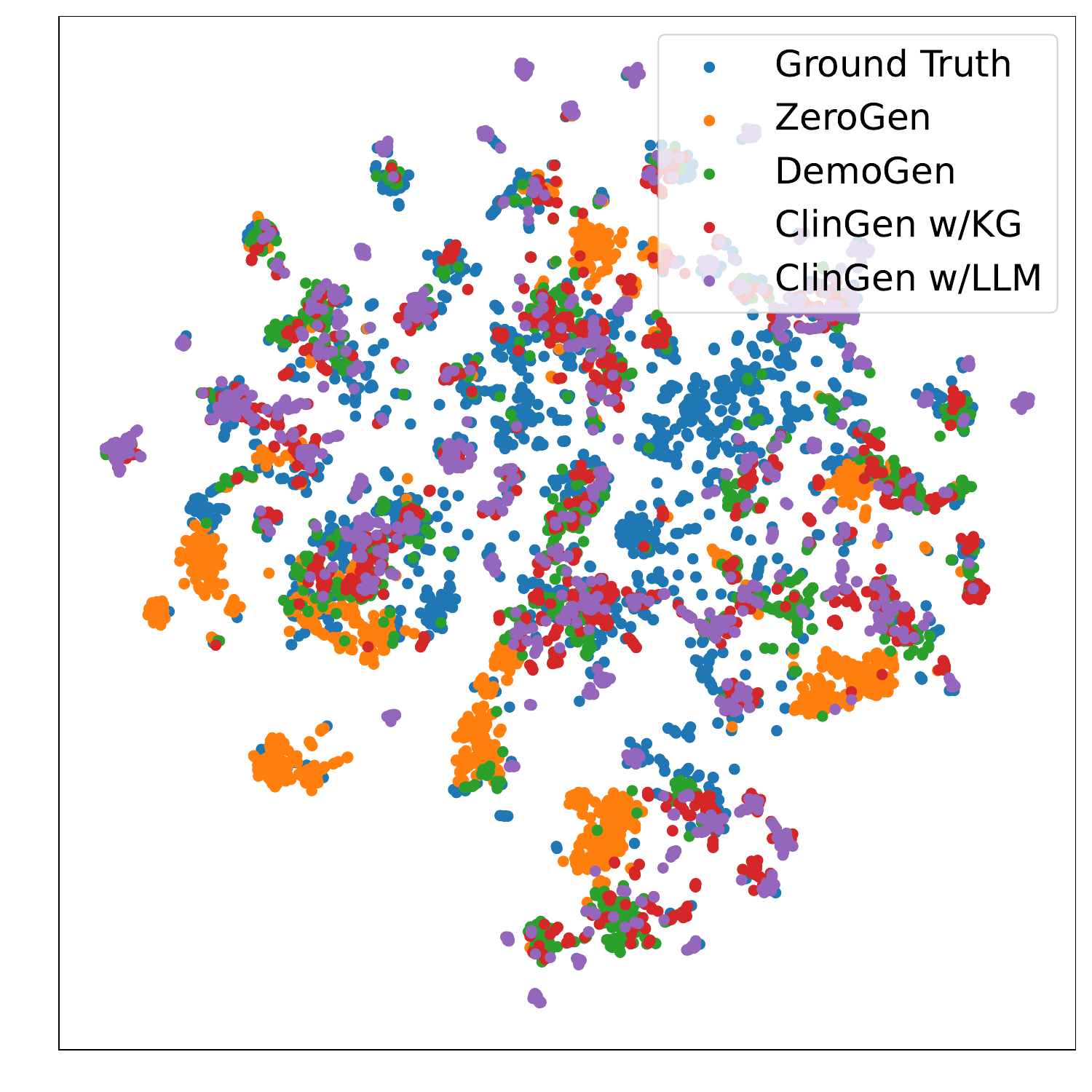}
	} %\hfill
         % \hspace{-1.5ex}
     \subfigure[GAD]{
		\includegraphics[width=0.31\linewidth]{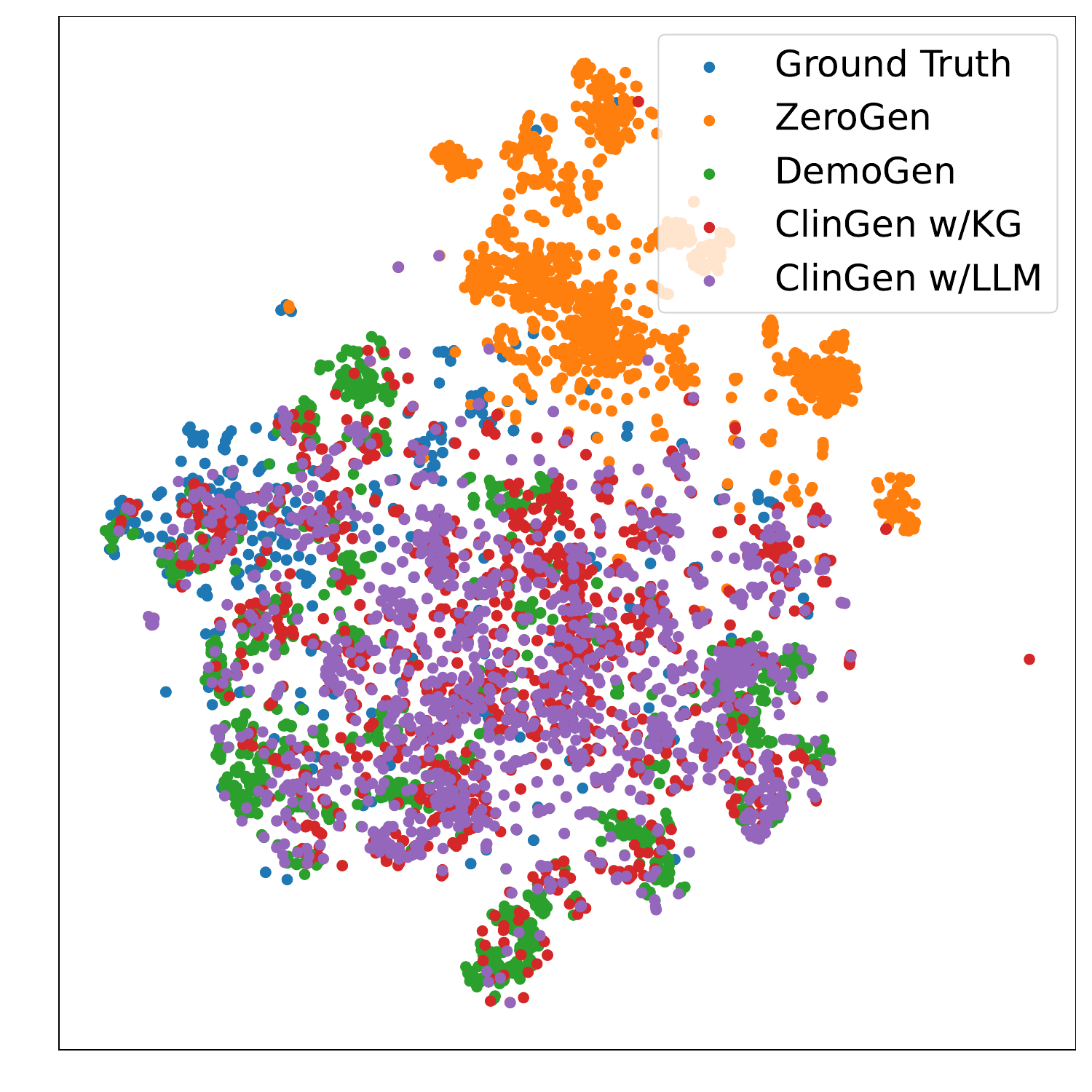}
	}
        % \hspace{-1.5ex}
      \subfigure[CDR]{
		\includegraphics[width=0.31\linewidth]{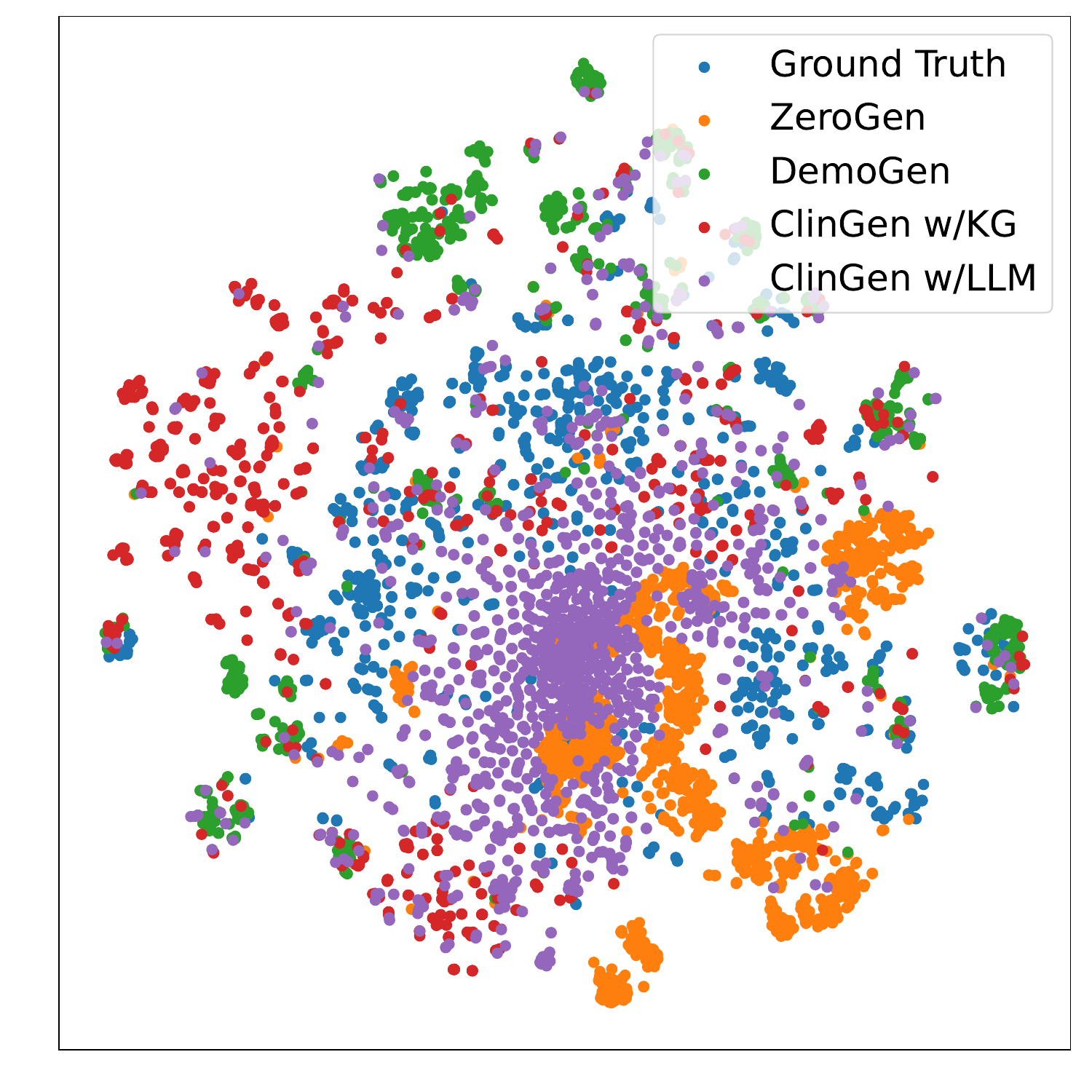}
	}
 	\subfigure[MEDIQA-RQE]{
		\includegraphics[width=0.31\linewidth]{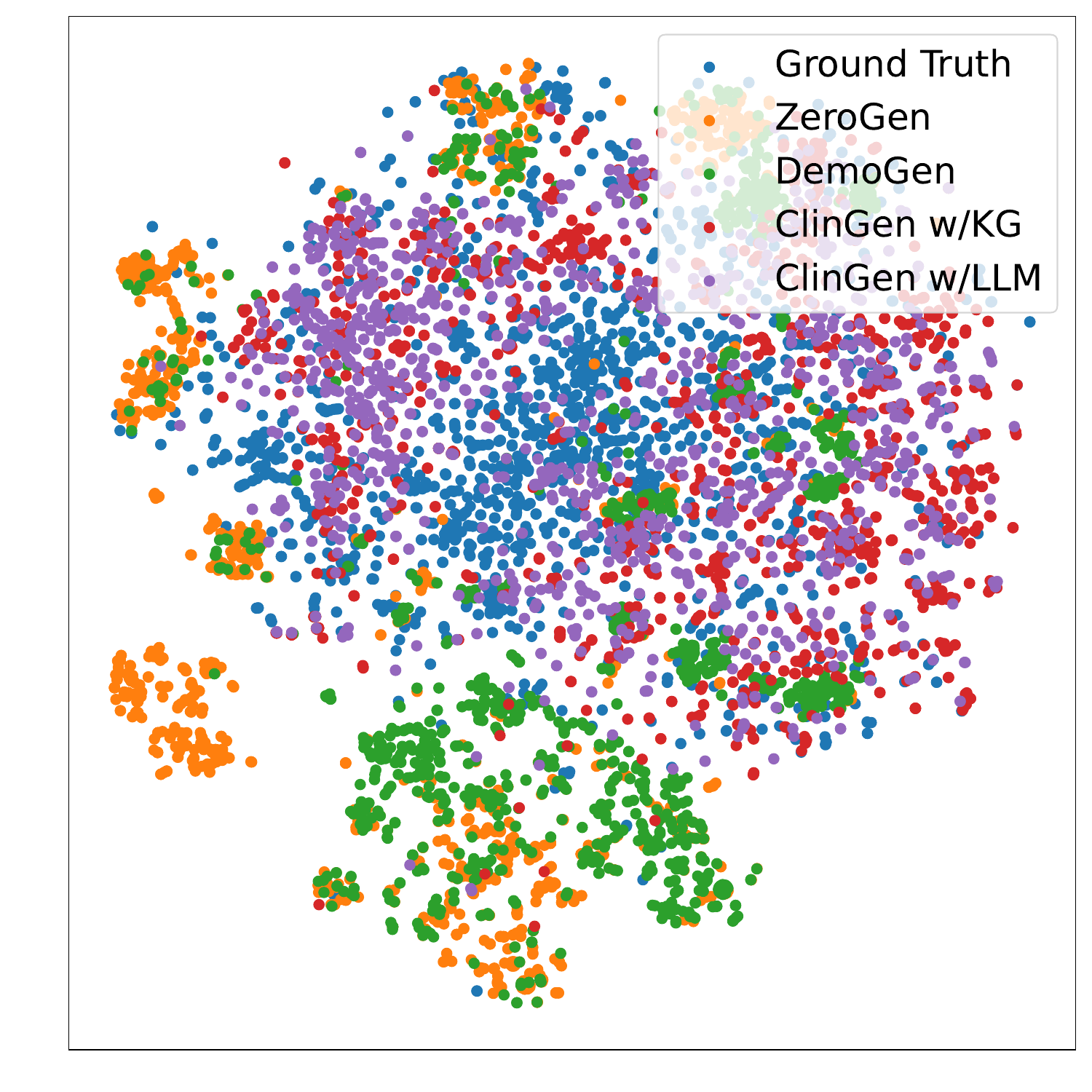}
	} %\hfill
         % \hspace{-1.5ex}
     \subfigure[MQP]{
		\includegraphics[width=0.31\linewidth]{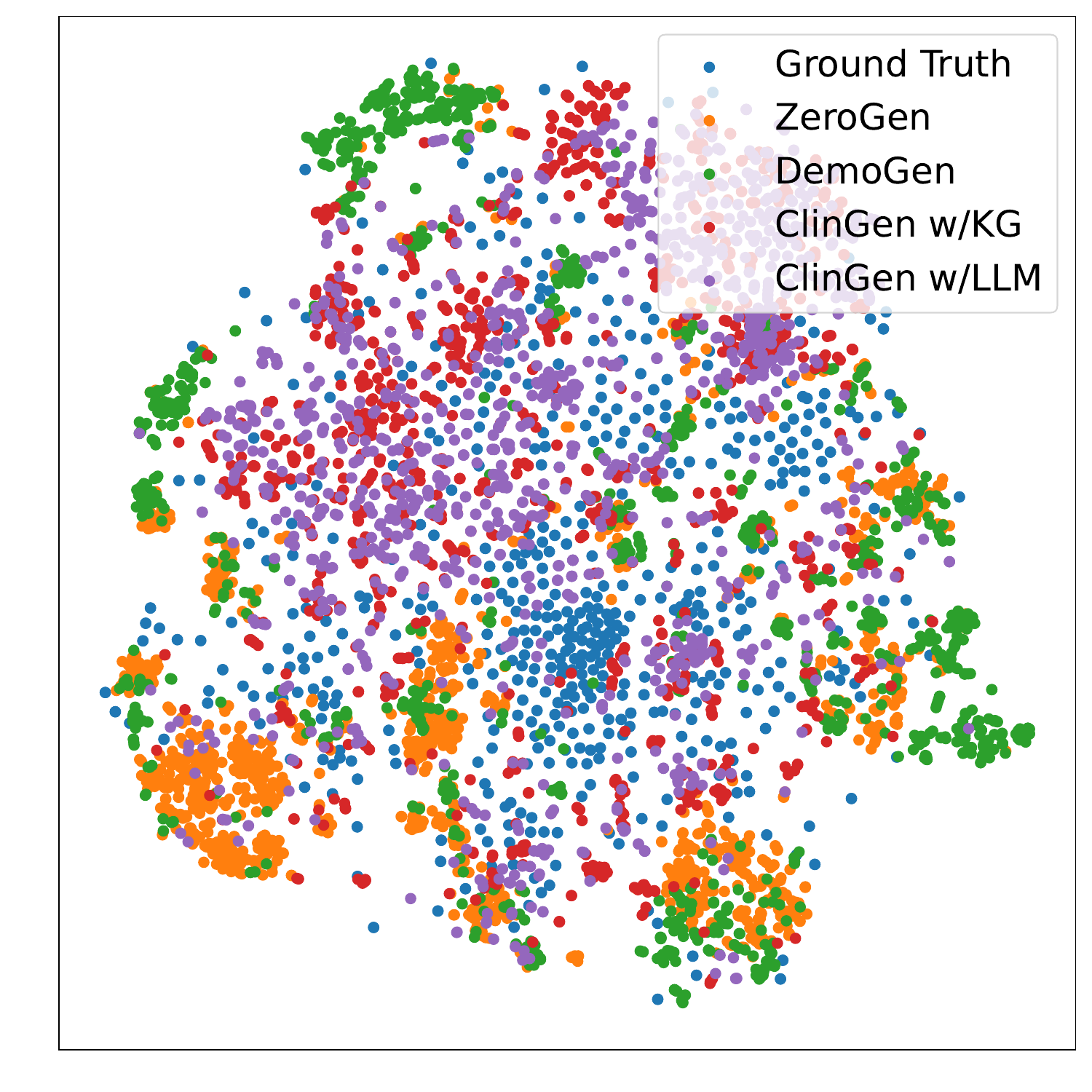}
	}
        % \hspace{-1.5ex}
      \subfigure[CHEMDNER]{
		\includegraphics[width=0.31\linewidth]{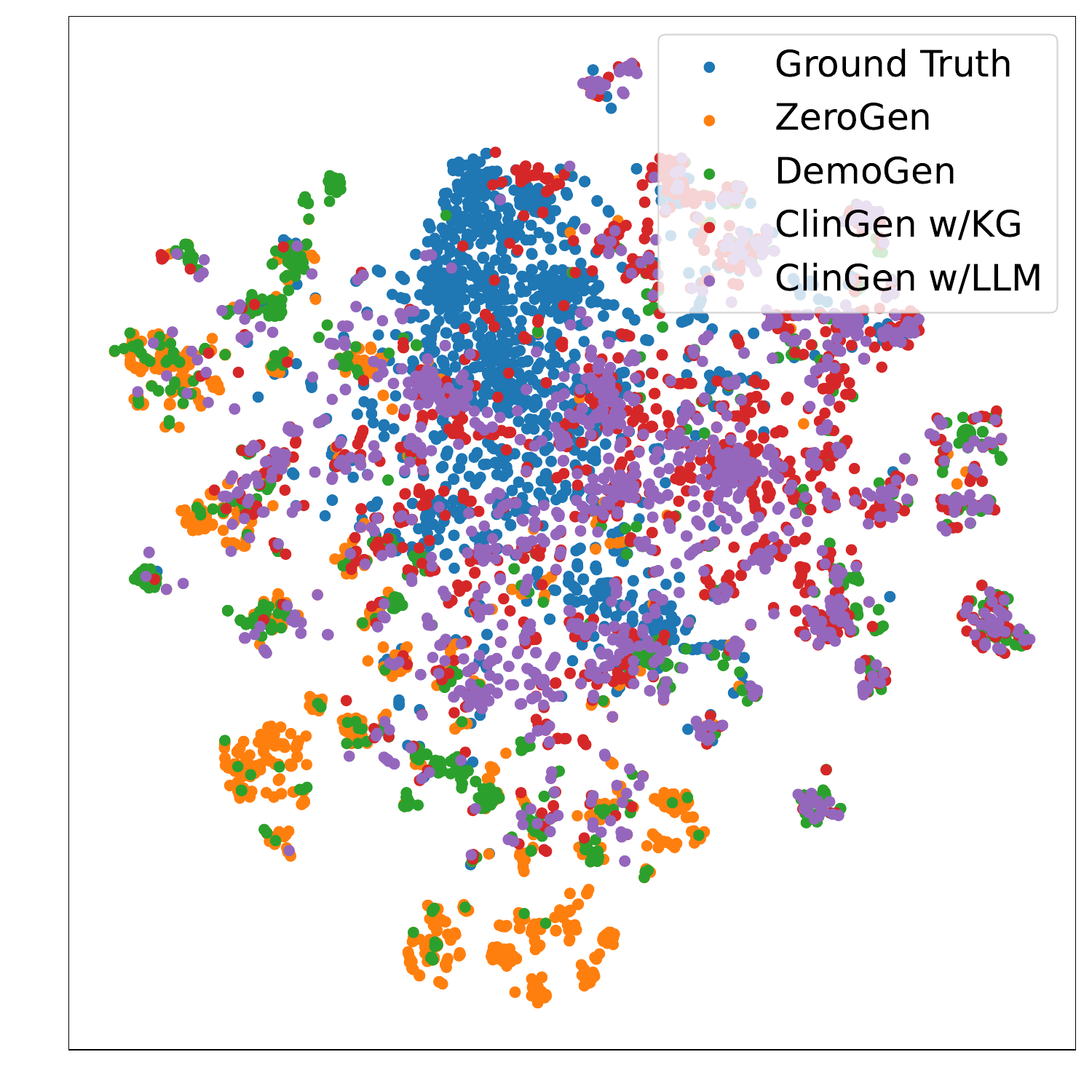}
	}
	\caption{The t-SNE plots of datasets generated by {\ours}, ZeroGen and DemoGen compared with the ground truth.}
\label{fig:add_quality_tsne}
\end{figure*}

 \begin{figure*}[!t]
	\centering
	\vspace{-2ex}
	\subfigure[LitCovid]{
		\includegraphics[width=0.31\linewidth]{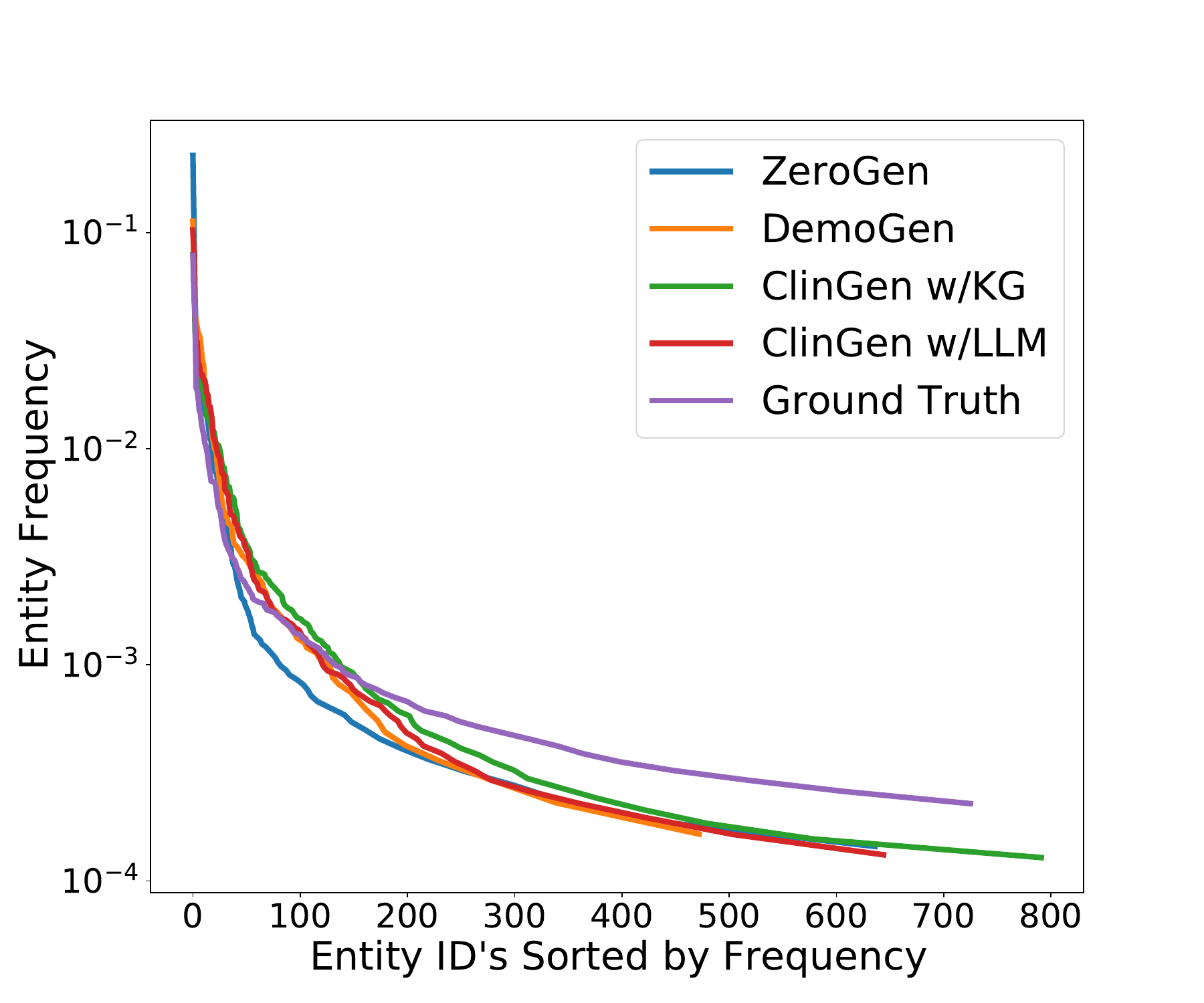}
	} %\hfill
         % \hspace{-1.5ex}
     \subfigure[GAD]{
		\includegraphics[width=0.31\linewidth]{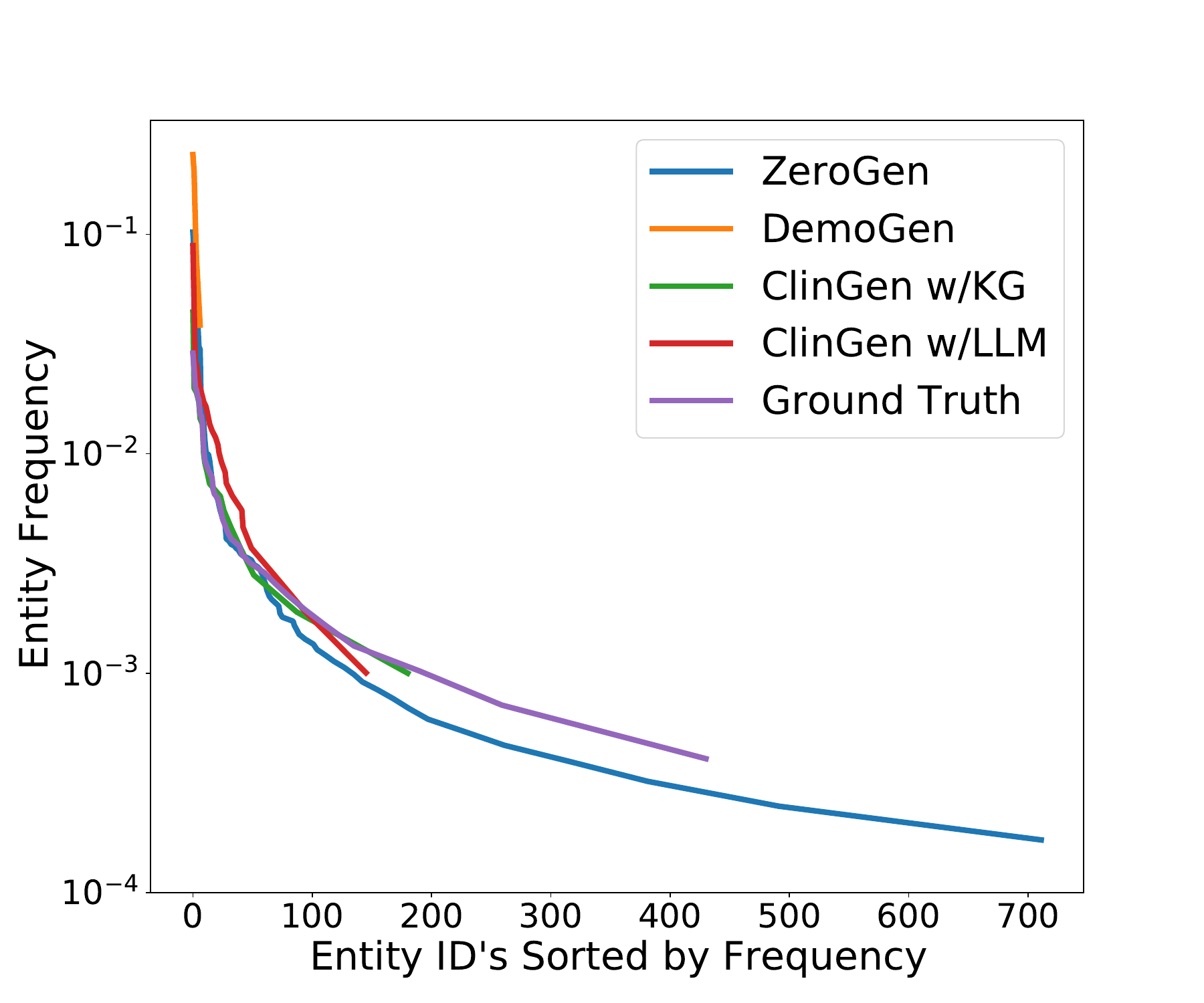}
	}
        % \hspace{-1.5ex}
      \subfigure[CDR]{
		\includegraphics[width=0.31\linewidth]{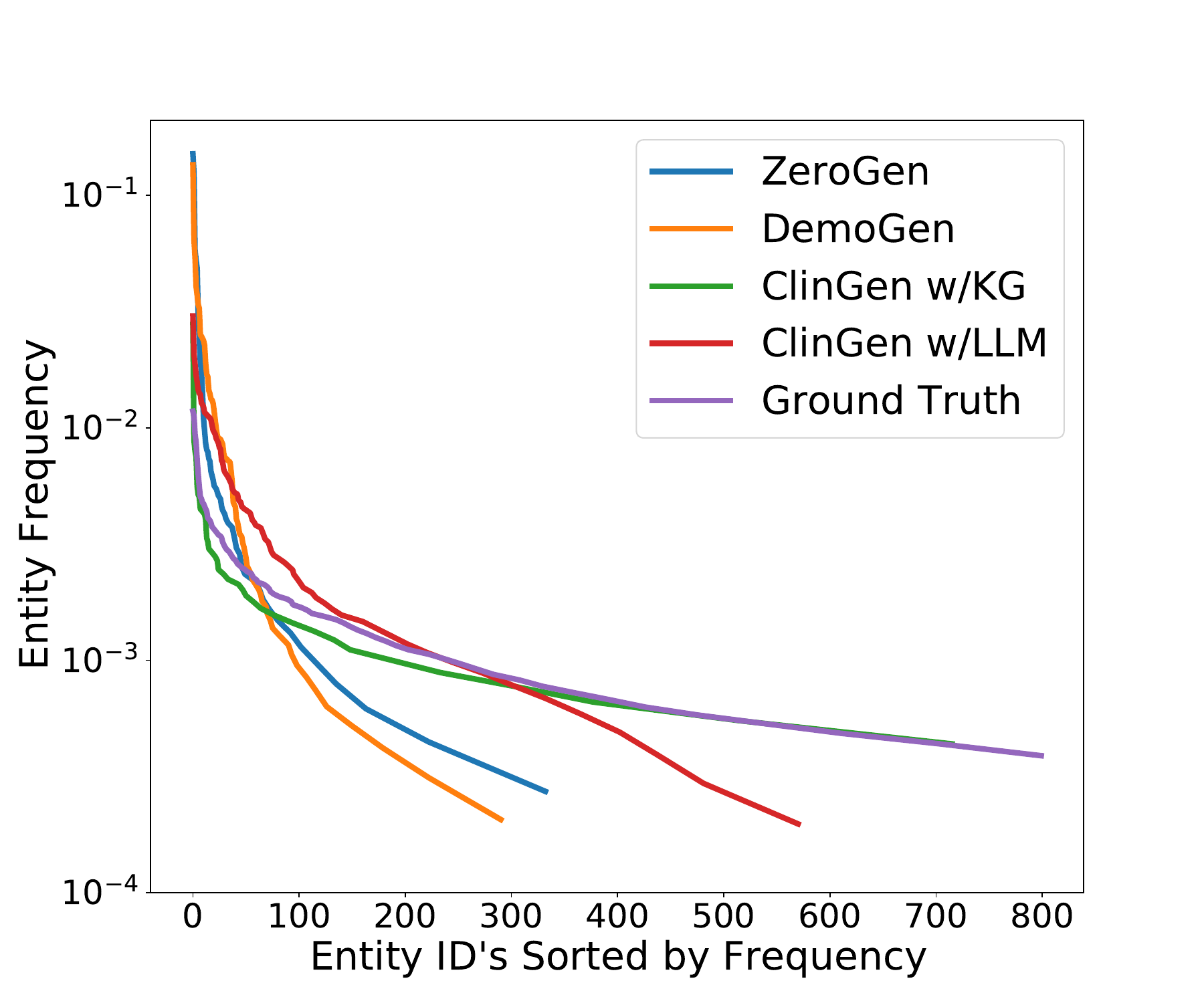}
	}
 	\subfigure[MEDIQA-RQE]{
		\includegraphics[width=0.31\linewidth]{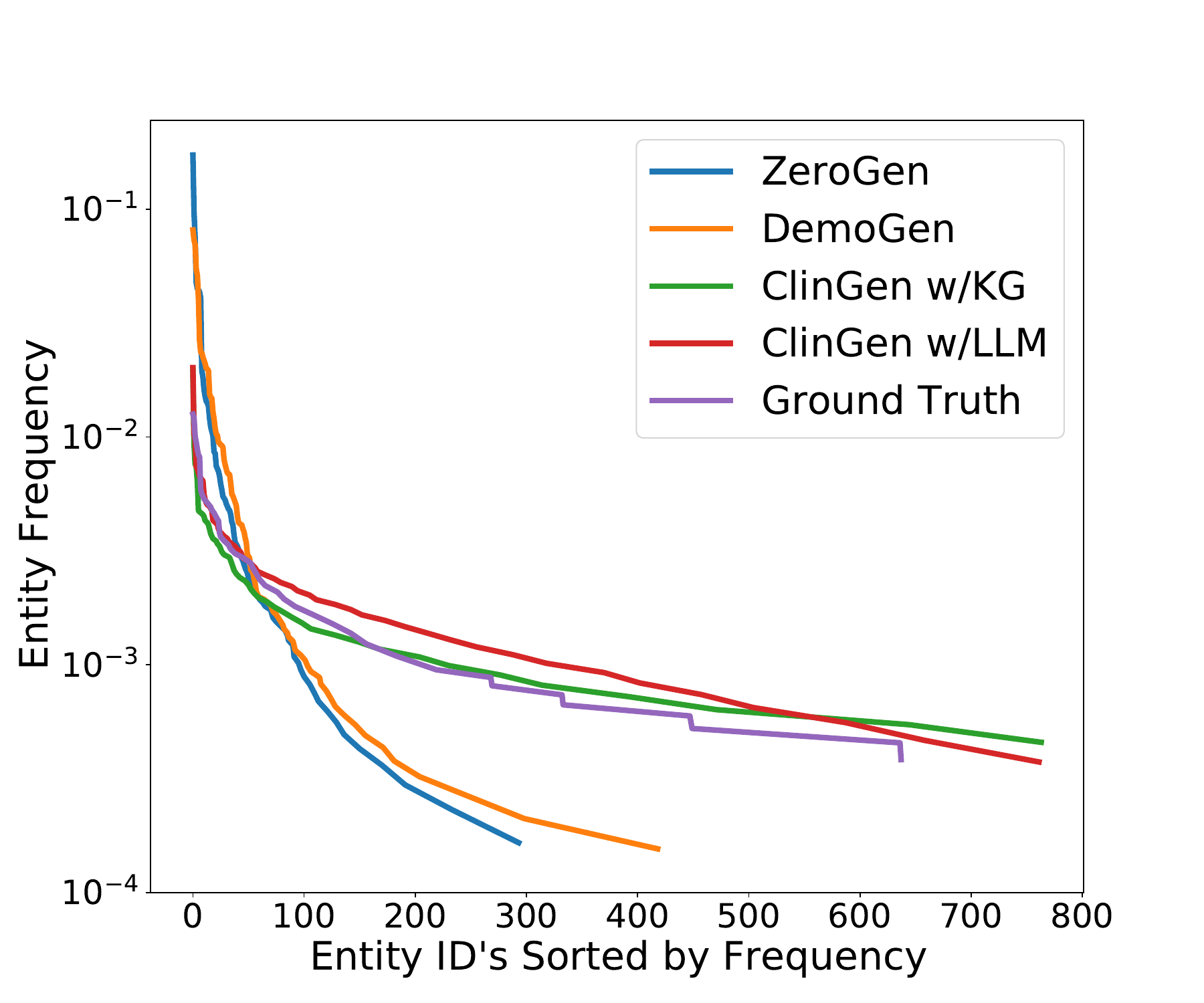}
	} %\hfill
         % \hspace{-1.5ex}
     \subfigure[MQP]{
		\includegraphics[width=0.31\linewidth]{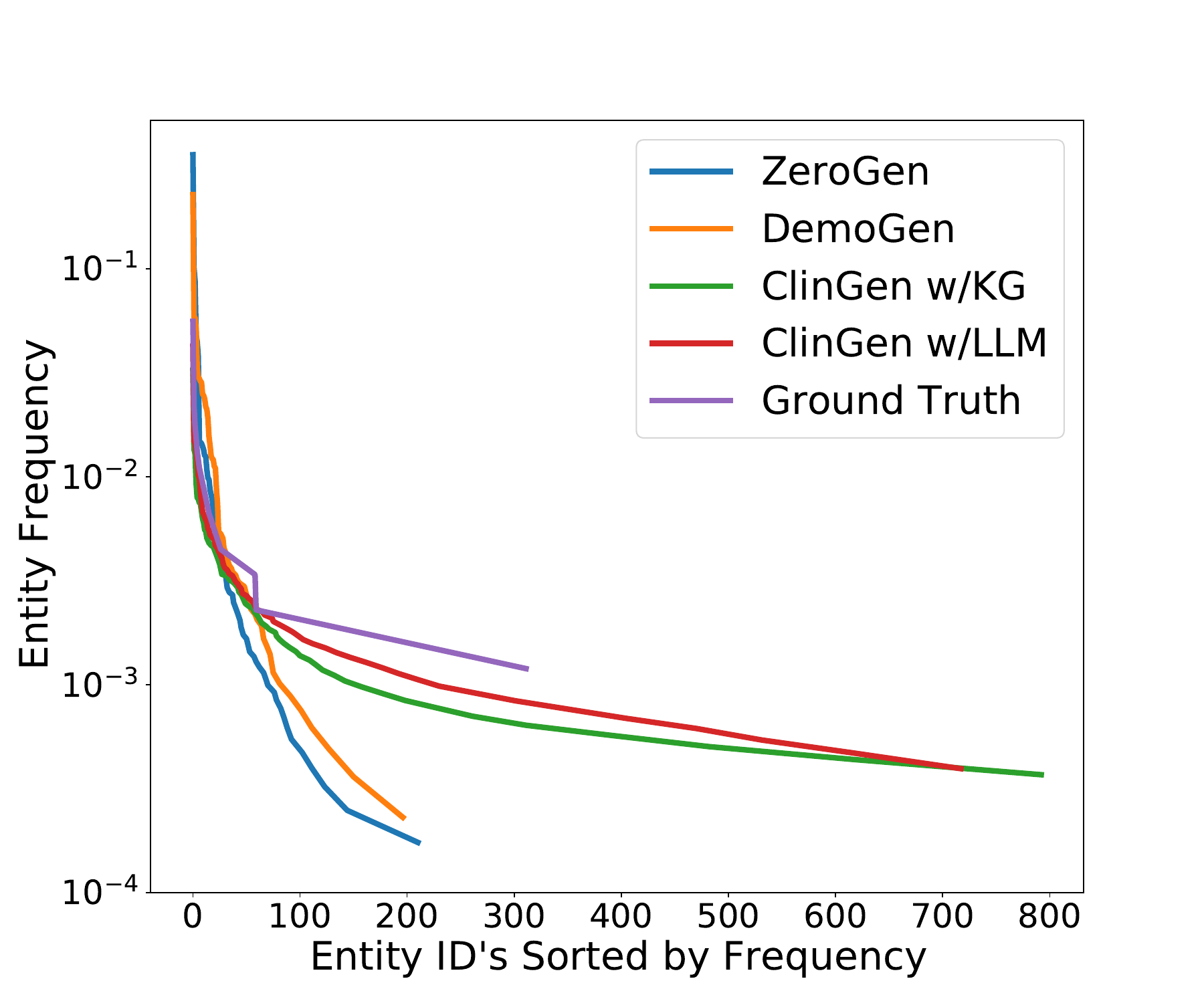}
	}
        % \hspace{-1.5ex}
      \subfigure[CHEMDNER]{
		\includegraphics[width=0.31\linewidth]{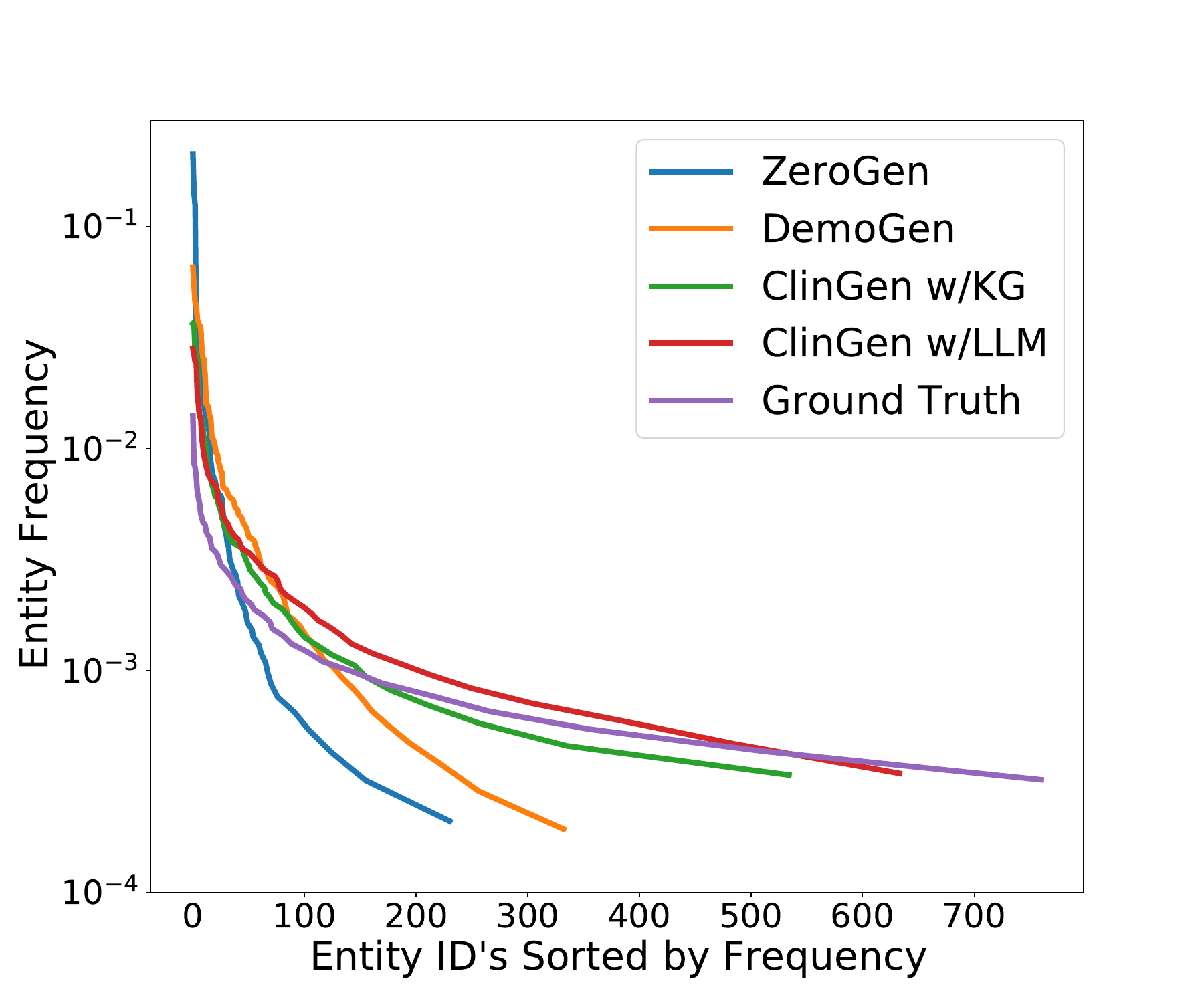}
	}
	\caption{The regularized entity frequencies of datasets generated by {\ours}, ZeroGen and DemoGen compared with the ground truth in log scale.}
\label{fig:add_quality_freq}
\end{figure*}
\clearpage

\section{Additional Ablation and Parameter Studies}
\label{sec:add_ablation_para}

Figure~\ref{fig:generator-add} and \ref{fig:size-synthetic-add} show the effect of different generators and the effect of the proportion of data on two additional datasets, respectively. Overall, our method generally outperform the best baseline. One interesting finding for the NCBI-Disease dataset is that {\ours} performs worse than the best on one variant. We hypothesize that it is because this task involves more complex input and output, potentially posing a challenge for moderate-size LLMs to follow the instructions. 

Besides, as few-shot sample selection is important for the final performance, we show the performance of different 3 random seeds in Table~\ref{tab:random_seed} (with different seed examples/training process), and observe that our method {\ours} generally outperforms the baselines with non-negligible margins, which indicates the robustness of {\ours} as it does not rely on a specific subset of few-shot training examples to perform well.

\section{Additional Quality Analysis}
We present additional quality analysis of the synthetic dataset with t-SNE plots in Figure~\ref{fig:add_quality_tsne} and the regularized entity frequencies in Figure~\ref{fig:add_quality_freq}.

\section{\blue{Comparison with different prompt designs}}
\label{sec:diff_prompt_design}
\subsection{\blue{Model Performance}}
We carry out an additional analysis with two recent and representative prompt optimization techniques, namely Reframe~\citep{mishra-etal-2022-reframing}, APE~\citep{zhou2023large} and PromptAgent~\citep{wang2023promptagent}. 
% It is worth noting that some other prompt optimization methods, such as OPRO~\citep{yang2023large} and ProTeGi~\citep{pryzant-etal-2023-automatic} are \textbf{not suitable} for this task, as they mainly target the reasoning tasks and prompt LLMs to generate the solution, since they need to use the correct/error examples from the previous iterations to refine the prompts, while we work on a different setup (i.e. synthetic data generation) and it is hard to define ``correct'' or ``error'' instances.
 
In our setting, Reframe incorporates several principles (e.g. using low-level patterns, itemizing instructions, etc.) to produce high-quality prompts to enhance text generation, whereas APE and PromptAgent leverage the LLM to optimize the prompts based on the target task information. We demonstrate their performance on various clinical tasks in Table~\ref{tab:diff_prompt_design}.  
The results indicate that our proposed {\ours} consistently outperforms both baselines. This performance gain is attributed to the fact that the prompts generated by these baselines do not \textit{adequately address the unique challenges for the clinical data generation}, i.e. distribution shift and lack of diversity. 
As a result, although they tend to include some generic task-specific information for guiding LLMs to generate training data, the performance gains brought by these advanced techniques are limited.
One important avenue of future work is to design effective approach to combine these automatic prompt optimization approaches with our extracted clinical-related concepts.

\begin{table*}[t]
% \floatconts
% \vspace{-1ex}
  \caption{Comparison between existing prompting optimization methods and {\ours}.}
  \resizebox{\linewidth}{!}{
  \begin{tabular}{lccccccc}
  \toprule
  & \bfseries LitCovid & \bfseries CDR & \bfseries MEDIQA-RQE & \bfseries MQP & \bfseries CHEMDNER & \bfseries BC5CDR-Disease & \bfseries Average\\
  % \midrule
  \cmidrule(lr){2-2} \cmidrule(lr){3-3} \cmidrule(lr){4-4} \cmidrule(lr){5-5} \cmidrule(lr){6-6} \cmidrule(lr){7-7} \cmidrule(lr){8-8}
  & F1 & F1 & ACC & ACC & F1 & F1 & -- \\
  \midrule
  \multicolumn{8}{l}{\textbf{PubMedBERT$_{\texttt{Base}}$}}\\
  \midrule
  Reframe~\scriptsize{\citep{mishra-etal-2022-reframing}} & 56.74 & 57.27 & 61.92 & 67.60 & 54.61 & 59.17 & 59.55 \\
  APE~\scriptsize{\citep{zhou2023large}} & 56.24 & 61.12 & 66.55 & 68.00 & 52.10 & 58.79 & 60.47 \\
  PromptAgent~\scriptsize{\citep{wang2023promptagent}} & 56.62 & 48.44 & 63.64 & 61.00 & 54.47 & 59.98 & 57.36
  \\
  \rowcolor{teal!10} {\ours} w/ KG & 58.01 & 61.75 & \textbf{74.85} & 72.20 & \textbf{56.94} & 60.75 & \textbf{64.08} \\
  \rowcolor{teal!10} {\ours} w/ LLM & \textbf{59.22} & \textbf{63.34} & 72.40 & \textbf{73.30} & 54.84 & \textbf{61.03} & 64.02 \\
  \midrule
  \multicolumn{8}{l}{\textbf{PubMedBERT$_{\texttt{Large}}$}}\\
  \midrule
  Reframe~\scriptsize{\citep{mishra-etal-2022-reframing}} & 54.06 & 58.78 & 66.57 & 71.30 & 55.05 & 60.41 & 61.03 \\
  APE~\scriptsize{\citep{zhou2023large}} & 53.54 & 61.65 & 69.20 & 71.00 & 53.03 & 59.87 & 61.38 \\
  PromptAgent~\scriptsize{\citep{wang2023promptagent}} & 54.54 &50.10  & 65.56 & 64.20 &\textbf{55.91} & 62.17 & 58.75
  \\
  \rowcolor{teal!10} {\ours} w/ KG & 55.81 & 62.66 & \textbf{79.92} & 75.40 & 55.56 & 61.21 & 65.16 \\
  \rowcolor{teal!10} {\ours} w/ LLM & \textbf{57.07} & \textbf{64.99} & 77.36 & \textbf{76.20} & 55.37 & \textbf{63.15} & \textbf{65.69} \\
  \bottomrule
  \end{tabular}
  }
  \label{tab:diff_prompt_design}
\end{table*}

\subsection{\blue{Prompt Templates}}
\blue{We provide the detailed prompt templates we use for Reframe~\citep{mishra-etal-2022-reframing}, APE~\citep{zhou2023large} and PromptAgent~\citep{wang2023promptagent} in the followings.
}

\textbf{Natural Language Inference tasks:}
\begin{lstlisting}[style=mystyle, caption={Prompt Format for generating sentences in NLI tasks with Reframe.}, label=lst:prompt, escapeinside={<@}{@>}]
Generate a pair of sentences for the <@\textcolor{blue}{[domain]}@> task. Follow these guidelines:
1. Formulate a medical premise in the first sentence, such as a clinical observation or a patient's medical history.
2. Craft a medical hypothesis or claim related to the premise in the second sentence.
3. Ensure that the hypothesis logically follows from the premise.
4. Avoid introducing any unrelated or contradictory information in either sentence.
5. The length should be in 50 words.
\end{lstlisting}

\begin{lstlisting}[style=mystyle, caption={Prompt Format for generating sentences in NLI tasks with APE.}, label=lst:prompt, escapeinside={<@}{@>}]
Generate a pair of sentences for the <@\textcolor{blue}{[domain]}@> task. The first sentence should be a medical premise, such as a clinical observation or a patient's medical history. The second sentence should be a medical hypothesis or claim, related to the premise. The goal is to determine whether the hypothesis logically follows from the premise, and you can use various medical scenarios, conditions, or treatments for creating these sentence pairs.
\end{lstlisting}

\begin{lstlisting}[style=mystyle, caption={Prompt Format for generating sentences in NLI tasks with PromptAgent.}, label=lst:prompt, escapeinside={<@}{@>}]
You've been assigned the task of creating a dataset for determining the <@\textcolor{blue}{[domain]}@> in medical text pairs. Ensure that you do not include any irrelevant information. Keep in mind that the content may involve medical conditions, treatments, and observations in various formats. Your goal is to accurately label the relationships for each medical text pair based on their logical connections.
\end{lstlisting}

\texttt{[domain]}: ``Question Entailment" for MEDIQA-RQE.

\textbf{Sentence similarity tasks:}
\begin{lstlisting}[style=mystyle, caption={Prompt Format for generating sentences in sentence similarity tasks with Reframe.}, label=lst:prompt, escapeinside={<@}{@>}]
Suppose you need to generate two sentences for the <@\textcolor{blue}{[domain]}@> task. Your task is to give a pair of sentences with the following instructions:
(1) Generate two sentences that exhibit a clear similarity or dissimilarity in meaning without using complex or specialized terms.
(2) express attributes affirmatively.
(3) Ensure that both sentences have a common attribute for comparison.
(4) The length should be in 50 words.
\end{lstlisting}

\begin{lstlisting}[style=mystyle, caption={Prompt Format for generating sentences in sentence similarity tasks with APE.}, label=lst:prompt, escapeinside={<@}{@>}]
Suppose you need to generate two sentences for the <@\textcolor{blue}{[domain]}@> task. The goal is to assess how close or similar the meaning of two sentences is, including 'equivalent' or 'not equivalent'.
\end{lstlisting}

\begin{lstlisting}[style=mystyle, caption={Prompt Format for generating sentences in sentence similarity tasks with PromptAgent.}, label=lst:prompt, escapeinside={<@}{@>}]
You've been assigned the job of creating a dataset for <@\textcolor{blue}{[domain]}@>. Make sure not to include any extraneous details. Keep in mind that sentences can vary in structure and wording while conveying similar meanings. Your task is to calculate the similarity score accurately for each sentence pair.\end{lstlisting}
\texttt{[domain]}: ``Sentence Similarity Calculation" for MQP.

\textbf{Text classification tasks:}

\begin{lstlisting}[style=mystyle, caption={Prompt Format for generating sentences in text classification tasks with Reframe.}, label=lst:prompt, escapeinside={<@}{@>}]
Suppose you are a writer for <@\textcolor{blue}{[domain]}@>. Your task is to give a synthetic <@\textcolor{blue}{[domain]}@> about <@\textcolor{blue}{[class\_name]}@> with the following instructions:
(1) Illustrate points with everyday scenarios related to the <@\textcolor{blue}{[class\_name]}@>.
(2) about 50 - 100 words.
\end{lstlisting}

\begin{lstlisting}[style=mystyle, caption={Prompt Format for generating sentences in text classification tasks with APE.}, label=lst:prompt, escapeinside={<@}{@>}]
Suppose you are a writer for <@\textcolor{blue}{[domain]}@>. Generate a clinical article discussing the latest advancements in <@\textcolor{blue}{[domain]}@> with a focus on <@\textcolor{blue}{[class\_name]}@>. Please include information on recent clinical trials, emerging research findings, and potential implications for healthcare practitioners and patients.
\end{lstlisting}

\begin{lstlisting}[style=mystyle, caption={Prompt Format for generating sentences in text classification tasks with PromptAgent.}, label=lst:prompt, escapeinside={<@}{@>}]
You've been assigned the responsibility of creating a dataset for classifying text related to <@\textcolor{blue}{[domain]}@>. Ensure that you do not include any irrelevant information. Keep in mind that references to COVID-19 may appear in various forms, including abbreviations and synonyms. Your objective is to accurately identify and classify text that is relevant to <@\textcolor{blue}{[domain]}@>.\end{lstlisting}

% You've been assigned the responsibility of creating a dataset for classifying text related to COVID-19 from the provided sentences. Ensure that you do not include any irrelevant information. Keep in mind that references to COVID-19 may appear in various forms, including abbreviations and synonyms. Your objective is to accurately identify and classify text that is relevant to COVID-19.
\texttt{[domain]}: ``COVID-19 Literature" for LitCovid.

\texttt{[class\_name]}: the label name for this generated sample.

\textbf{Relation extraction tasks:}

\begin{lstlisting}[style=mystyle, caption={Prompt Format for generating sentences in relation extraction tasks with Reframe.}, label=lst:prompt, escapeinside={<@}{@>}]
Suppose you need to generate a dataset for the biomedical <@\textcolor{blue}{[domain]}@> task where the relationships between entities in biomedical texts need to be identified. Your task is to give a synthetic example about <@\textcolor{blue}{[class\_name]}@> relation with the following instructions:
(1) Provide the sentence or text snippet where the relationship is mentioned.
(2) The length should be in 50 words.
\end{lstlisting}

\begin{lstlisting}[style=mystyle, caption={Prompt Format for relation extraction tasks with APE.}, label=lst:prompt, escapeinside={<@}{@>}]
Generate a sentence that describes a <@\textcolor{blue}{[class\_name]}@> <@\textcolor{blue}{[domain]}@> between <@\textcolor{blue}{[entity0]}@> and <@\textcolor{blue}{[entity1]}@>. The sentence should provide information about how these terms are related, such as its potential therapeutic use, side effects, or any relevant research findings.
\end{lstlisting}

\begin{lstlisting}[style=mystyle, caption={Prompt Format for relation extraction tasks with PromptAgent.}, label=lst:prompt, escapeinside={<@}{@>}]
You've been assigned the task of creating a <@\textcolor{blue}{[class\_name]}@> <@\textcolor{blue}{[domain]}@> dataset for identifying relationships between <@\textcolor{blue}{[entity0]}@> and <@\textcolor{blue}{[entity1]}@> from the provided text. Be sure to exclude any extraneous information. Keep in mind that chemicals and diseases may be referred to using various names, abbreviations, or synonyms. Your goal is to recognize and extract these associations accurately.
\end{lstlisting}
\texttt{[domain]}: ``Chemical Disease Relation" for CDR.

\texttt{[entity0]} and \texttt{[entity1]}: ``chemical" and ``disease: for CDR.

\texttt{[class\_name]}: the label name for this generated sample.

\textbf{Named entity recognition tasks:}

\begin{lstlisting}[style=mystyle, caption={Prompt Format for generating sentences in NER tasks with Reframe.}, label=lst:prompt, escapeinside={<@}{@>}]
Suppose you need to create a dataset for <@\textcolor{blue}{[domain]}@> recognition. Your task is to generate a sentence about <@\textcolor{blue}{[domain]}@> and also output the <@\textcolor{blue}{[domain]}@> name with the following instructions:
(1) Generate a sentence that contains a named entity. The named entity should be a recognizable entity type within the sentence. 
(2) The named entity must be contextually relevant and correctly labeled with its type. 
(3) The length should be in 50 words.
\end{lstlisting}

\begin{lstlisting}[style=mystyle, caption={Prompt Format for NER tasks with APE.}, label=lst:prompt, escapeinside={<@}{@>}]
Suppose you need to create a dataset for <@\textcolor{blue}{[domain]}@> recognition. Generate a sentence or short text passage where you mention a <@\textcolor{blue}{[domain]}@> entity within a context. The named entity should be clearly identifiable within the text.
\end{lstlisting}

\begin{lstlisting}[style=mystyle, caption={Prompt Format for NER tasks with PromptAgent.}, label=lst:prompt, escapeinside={<@}{@>}]
You're tasked with generating a dataset for recognizing <@\textcolor{blue}{[domain]}@> from the given sentence. Remember to avoid incorporating any
associated elements. Consider both specific diseases and broader categories, and remember diseases and conditions can also appear as common abbreviations
or variations. 
\end{lstlisting}

\texttt{[domain]}: ``disease" for BC5CDR-Disease; ``chemical" for CHEMDNER.

\section{Using Medical LLMs as Data Generator}
\label{sec:med_llm_gen}
In this work, we mainly evaluate {\ours} using GPT-family models as the LLM.
However, we are aware that many LLMs have been fine-tuned on additional clinical contexts as well as instructions and achieved superior performance on clinical NLP benchmarks.
We select MedAlpaca-13b~\citep{han2023medalpaca} as one representative clinical LLM and study the effect of {\ours} using a medical LLM as the data generator. Many other medical LLMs, such as Med-PALM\footnote{\url{https://sites.research.google/med-palm/}}, are not open-sourced, thus we cannot run them in our experiments. 

From the results shown in Table~\ref{tab:med_llm_generator}, we observe that using medical LLM as the clinical text data generator exhibits lower downstream performance. This could be attributed to the medical LLMs having fewer parameters than ChatGPT, which results in limited instruction-following capabilities.
% An interesting future work could be how to leverage these Clinical LLMs as Data Generators to further boost the performance. Besides, it can be beneficial to inject more fine-grained clinical knowledge beyond entities and relations to further benefit data generation pipelines.

\begin{table}[h]
% \floatconts
  \caption{The performance of {\ours} with the medical LLM MedAlpaca as data generator.}
  \resizebox{0.9\linewidth}{!}{
  \begin{tabular}{lcc}
  \toprule
  & \bfseries LitCovid & \bf CHEMDNER \\
  \midrule
  \multicolumn{3}{l}{\textbf{PubMedBERT$_{\texttt{Base}}$}}\\
  \midrule
  \rowcolor{teal!10} {\ours} w/ KG & 58.01 & \textbf{56.94} \\
  \rowcolor{teal!10} {\ours} w/ LLM (ChatGPT) & \textbf{59.22} & 54.84 \\
  {\ours} w/ LLM (MedAlpaca) & 55.45 & 52.15  \\ 
  \midrule
  \multicolumn{3}{l}{\textbf{PubMedBERT$_{\texttt{Large}}$}}\\
  \midrule
  \rowcolor{teal!10} {\ours} w/ KG & 55.81 & \textbf{55.56} \\
  \rowcolor{teal!10} {\ours} w/ LLM (ChatGPT) & \textbf{57.07} & 55.37 \\
  {\ours} w/ LLM (MedAlpaca) & 53.90 & 52.67  \\ 
  \bottomrule
  \end{tabular}
  }
  \label{tab:med_llm_generator}
  \vspace{-1ex}
\end{table}

\section{Effect of Data Mixing Ratio}
\label{sec:data_mix}

In this work, we present KGs and LLMs as two alternative and complementary sources for obtaining topics. However, we also consider combining topics from KGs and LLMs as a potential approach to enhance performance. Thus, we conduct experiments to demonstrate the impact of combining topics from KGs and LLMs at various ratios. Note that we still keep a total of 5000 generated synthetic samples to maintain a fair comparison.
The experimental results in Table~\ref{tab:diff_mixing_ratio} indicate that combining knowledge from KGs and LLMs can yield a performance improvement, though not a substantial one. However, note that in practice, it is challenging to tune the ratio in the few-shot setting due to the limited volume of validation labels~\citep{perez2021true}, and thus we only include the 1:1 results in Tables~\ref{tab:single-sent}, ~\ref{tab:sent-pair}, ~\ref{tab:token-class} in Appendix~\ref{sec:more_experimental_results} for all the datasets.

\begin{table}[h]
% \floatconts
% \vspace{-1ex}
  \caption{Effect of mixing topics generated from KG and LLM in different ratio.}
  \resizebox{\linewidth}{!}{
  \begin{tabular}{lccccc}
  \toprule
  \multirow{2.5}{*}{\bfseries KG : LLM} & \bfseries LitCovid & \bfseries CDR & \bfseries MEDIQA-RQE & \bfseries BC5CDR-Disease & \bfseries Average\\
  % \midrule
  \cmidrule(lr){2-2} \cmidrule(lr){3-3} \cmidrule(lr){4-4} \cmidrule(lr){5-5} \cmidrule(lr){6-6}
  & F1 & F1 & ACC & F1 & -- \\
  \midrule
  \multicolumn{6}{l}{\textbf{PubMedBERT$_{\texttt{Base}}$}}\\
  \midrule
  1:0 & 58.01 & 61.75 & 74.85 & 60.75 & 63.84 \\
  2:1 & 56.18 & 62.89 & 73.50 & 60.53 & 63.28 \\
  1:1 & 56.76 & 63.86 & 74.01 & 63.26 & 64.47 \\
  1:2 & 55.49 & 64.33 & 75.10 & 61.62 & 64.14 \\
  0:1 & 59.22 & 63.34 & 72.40 & 61.03 & 64.00 \\
  \midrule
  \multicolumn{6}{l}{\textbf{PubMedBERT$_{\texttt{Large}}$}}\\
  \midrule
  1:0 & 55.81 & 62.66 & 79.92 & 61.21 & 64.90 \\
  2:1 & 54.21 & 64.22 & 76.15 & 62.40 & 64.25 \\
  1:1 & 56.80 & 65.90 & 79.12 & 65.94 & 66.94 \\
  1:2 & 54.41 & 64.68 & 80.77 & 64.55 & 66.10 \\
  0:1 & 57.07 & 64.99 & 77.36 & 63.15 & 65.64 \\
  \bottomrule
  \end{tabular}
  }
  \label{tab:diff_mixing_ratio}
\end{table}

\end{document}